\documentclass[nohyperref]{article}

\usepackage{microtype}
\usepackage{graphicx}
\usepackage{subfigure}
\usepackage{booktabs} % for professional tables

\usepackage{hyperref}

\usepackage[accepted]{icml2022}

\usepackage{amsmath}
\usepackage{amssymb}
\usepackage{mathtools}
\usepackage{amsthm}

\usepackage[capitalize,noabbrev]{cleveref}
\usepackage{empheq,etoolbox}
\usepackage[font=small,skip=0pt]{caption}

\makeatletter
\newtheorem*{rep@theorem}{\rep@title}
\newcommand{\newreptheorem}[2]{%
\newenvironment{rep#1}[1]{%
\def\rep@title{#2 \ref{##1}}%
\begin{rep@theorem}}%
{\end{rep@theorem}}}
\makeatother

\newreptheorem{lemma}{Lemma}
\newreptheorem{proposition}{Proposition}
\newreptheorem{theorem}{Theorem}
\newreptheorem{corollary}{Corollary}

\newcommand{\R}{\mathbb{R}} % all real numbers.
\newcommand{\diag}{\text{diag}}%  
\theoremstyle{plain}
\newtheorem{theorem}{Theorem}[section]
\newtheorem{proposition}[theorem]{Proposition}
\newtheorem{lemma}[theorem]{Lemma}
\newtheorem{corollary}[theorem]{Corollary}
\theoremstyle{definition}
\newtheorem{definition}[theorem]{Definition}

\newtheorem{example}[theorem]{Example}
\theoremstyle{remark}
\newtheorem{remark}[theorem]{Remark}

\usepackage[textsize=tiny]{todonotes}

\icmltitlerunning{Discrete Probabilistic Inverse Optimal Transport}

\begin{document}

\twocolumn[
\icmltitle{Discrete Probabilistic Inverse Optimal Transport}



\icmlsetsymbol{equal}{*}

\begin{icmlauthorlist}
\icmlauthor{Wei-Ting Chiu}{equal,yyy}
\icmlauthor{Pei Wang}{equal,yyy}
\icmlauthor{Patrick Shafto}{yyy,comp}
\end{icmlauthorlist}

\icmlaffiliation{yyy}{Department of Mathematics and Computer Science, Rutgers University Newark, NJ}
\icmlaffiliation{comp}{School of Mathematics, Institute for Advanced Study (IAS), Princeton NJ}

\icmlcorrespondingauthor{Patrick Shafto}{patrick.shafto@gmail.com}
\icmlcorrespondingauthor{Wei-Ting Chiu}{wc570@rutgers.edu}
\icmlcorrespondingauthor{Pei Wang}{peiwang425@gmail.com}

\icmlkeywords{Machine Learning, ICML}

\vskip 0.3in
]



\printAffiliationsAndNotice{\icmlEqualContribution} % otherwise use the standard text.

\begin{abstract}
Inverse Optimal Transport (IOT) studies the problem of inferring the underlying cost that gives rise to an observation on coupling two probability measures.
Couplings appear as the outcome of matching sets (e.g. dating) and moving distributions (e.g. transportation).
Compared to Optimal transport (OT), the mathematical theory of IOT is undeveloped.
We formalize and systematically analyze the properties of IOT using tools from the study of entropy-regularized OT.
Theoretical contributions include characterization of the manifold of cross-ratio equivalent costs, the implications of model priors,
and derivation of an MCMC sampler. Empirical contributions include visualizations of cross-ratio equivalent effect on basic examples, simulations validating theoretical results and experiments on real world data.
% Inverse Optimal Transport (IOT) studies the problem of 
% inferring the latent \pw{or underlying?} cost that underlies \pw{or gavels?} given observations on coupling two probability measures.
% %   Optimal transport (OT) formalizes the problem of finding an optimal coupling between probability measures given a cost matrix. 
% %   The inverse problem of inferring the cost given a coupling is Inverse Optimal Transport (IOT).  
% Such couplings appear commonly when one set is matched with a second set \pw{put an concrete example like the underlying cost given transportation patterns?}.
% However, comparing to Optimal transport (OT), IOT is less well understood. 
% In this paper, we formalize and systematically analyze the properties of IOT using tools from the study of entropy-regularized OT. 
% Theoretical contributions include characterization of the manifold of cross-ratio equivalent costs, the implications of model priors, 
% and derivation of an MCMC sampler. Empirical contributions include visualizations of cross-ratio equivalent effect on basic examples, simulations validating theoretical results and 
% experiments on real world data. %We conclude with limitations and implications.
%   %and validating predictions on real-world NBA performance.  
%   %assessing compositional behavior of costs, inference of censored values, and inference of costs underlying team performance in the NBA. 
\end{abstract}

%A common goal in machine learning is to infer, given some observations, estimate the costs that underlie the observations. 
%Given a coupling of two discrete probability distributions, can we infer the underlying costs that explain the observed joint density? Such couplings appear commonly when one set is matched with a second set. Examples include inferring the underlying distances given transportation patterns, affinity between matched pairs \cite{}, and  \pat{FIXME}. The forward problem of computing stochastic couplings is typically formulated as Entropy regularized optimal transport. Here we investigate the problem of inferring underlying costs via Inverse entropy regularized Optimal Transport.  

%%%%%%%%%%%%%%%%%%%%%%% intro %%%%%%%%%%%%%%%%%%%%%%%
\section{Introduction}

Inverse Optimal Transport (IOT)  involves inferring latent costs from stochastic couplings, a problem that arises across many domains where matching or movement of distributions arises \citep{ma2020learning,stuart2020inverse,li2019learning,paty2020regularized}. 
%has received less systematic attention than (entropy-regularized) Optimal Transport (OT) \citep{villani2008optimal,cuturi2013sinkhorn}. 
%Although researchers have studied IOT in migration \citep{stuart2020inverse} and matching problems \citep{li2019learning}, and through adversarial robustness \citep{paty2020regularized} formulation and investigation of the general problem of probabilistic inference of latent costs from stochastic couplings given priors has not been systematically undertaken. 
%Formulation and investigation of the general problem of probabilistic inference of latent costs from stochastic couplings given priors has not been systematically undertaken.
Unlike (Entropy-regularized) Optimal Transport \citep{villani2008optimal,cuturi2013sinkhorn}, mathematical foundations and computational implications have not been systematically investigated. 
IOT inherits interesting geometric structure from Sinkhorn scaling \citep{sinkhorn1964relationship}, which gives rise to natural questions about the nature of posterior distributions, the existence of samplers, and feasibility of inferences from observed couplings to underlying costs. 

IOT problems has been applied in many areas. Practical examples include: (1) inferring the cost criterion of international migration \citep{stuart2020inverse}; (2) estimating the marriage matching affinity \citep{dupuy2019estimating,ma2020learning}; (3) learning interaction cost function from noisy and incomplete matching matrix and make prediction on new matching \citep{li2019learning}. These works provide methodologies to solve \textit{specific} IOT problems with extra hard constraints on latent costs. In this paper we provide a framework to analyze discrete probabilistic inverse optimal transport (PIOT) for  \textit{general} problems, and to understand the geometric properties of the space of latent costs.
%while in this paper we provide a framework for analyzing the general problem to understand the geometric properties of probabilistic inverse optimal transport (PIOT).

Our contributions include: (1) formulating the PIOT problem and identify the support manifold for latent costs; 
(2) characterizing the geometry of the support manifold along with the implications of prior distributions on posterior inference; 
(3) analyzing the support manifold for noisy and incomplete observations.
(4) developing MCMC algorithms for posterior inference which derives directly from geometric properties
of Sinkhorn scaling; 
(5) demonstrating PIOT through simulations on general examples 
including visualizations of latent cost manifold and  missing/noisy value predictions for migration data. 
%\pw{need some help on item (5).}
%We investigate properties of PIOT with several contributions. We formalize the general problem and proving basic results; characterize the manifold of support of the posterior distribution along with the implications of prior distributions on posterior inference; and present MCMC algorithms for posterior inference which derives directly from properties of Sinkhorn scaling, and show by simulation that samplers are effective on synthetic problems.  
%Our third contribution is to explore the feasibility of inference in this model by simulation and by demonstration on a real-world performance of NBA basketball teams. We conclude by discussing implications of the approach and interesting directions for extending the work. 

\textbf{Notation:} $\mathbb{R}^+$ denotes positive reals, $\mathbb{R}^*$ denotes non-negative reals.
$\overline{\mathbb{R}^*}$ denotes the extended non-negative reals $ \mathbb{R}^* \cup \{+\infty\}$. 
$\Delta_k$ is the $k$-dimensional simplex.
For a space $X$, $\mathcal{P}(X)$ denotes the set of distributions over $X$. 
%All matrices are assumed to be non-negative, and have no zero rows or zero columns unless otherwise stated. 
Matrices are in uppercase and their elements are in the corresponding lowercase.
For a matrix $A_{m\times n}$, $\mathbf{a}_1, \dots,\mathbf{a}_{n}$ denotes columns of $A$. %$\text{Col}(M)$ denotes column normalization of $M$.
Column normalizing a matrix $A_{m\times n}$
with respect to a $n$-dim row vector $\nu$ is defined as 
$\text{Col}(A, \nu) = A \, \text{diag} (\frac{\nu}{\mathbf{1}_{m} A})$, 
where $\mathbf{1}_{m}$ is a $m$-dim row vector with each element equals to~$1$. 
`with respect to $\nu$' is omitted when $\nu = \mathbf{1}_{n}$. 
Row normalization is defined in the same fashion.

%%%%%%%%%%%%%%%%%%%%%%% theory %%%%%%%%%%%%%%%%%%%%%%%

\section{Background and Related Work}
%Consider the situation where data from a collection of models were observed and the inference needs to be made on the underlying interactions between those models.
%In this section we formalize such inference as probabilistic inverse optimal transport. 
%\pw{discuss, notations, name for H,  T and C, $t_{ij}, c_{ij}$}
%\pat{pull out all details about column normalization to make it fully general to dirichlet prior on cost}

Given two spaces $X = \{x_1, \dots, x_m\}$ and $Y = \{y_1, \dots, y_n\}$, 
let $T = (t_{ij})_{m\times n} \in (\mathbb{R}^*)^{m\times n}$ 
be a \textit{coupling} that records the co-occurrence distribution between $X$ and $Y$.
To avoid trivial cases, we assume $m,n \geq 2$.
$t_{ij}$ is allowed to be zero here.
%Each column of $P$ is indexed by an $h$ and each row of $P$ is indexed by a $d$.
Each element $t_{ij}$ represents the probability of observing $x_i$ and $y_j$ simultaneously. 
The interaction between $X$ and $Y$ can be captured by a \textit{cost matrix} 
$C = (c_{ij})_{m\times n} \in (\overline{\mathbb{R}^*})^{m\times n}$, 
where $c_{ij}$ measures the underlying cost of coupling $x_i$ and $y_j$. 
Hence, the probabilistic inference on the underlying structure over $X$ and $Y$ is then to obtain $P(C|T)$, 
the distribution over all possible $C$ given an observed coupling $T$.

%The total cost of a plan $T$ is defined to be $\langle C, T\rangle = \sum_{i\in \D, j\in \hp} c_{ij}t_{ij}$, the Frobenius inner product.
%Let the prior of $C$ be $P_0(C)$, a distribution over $(\mathbb{R}^*)^{m\times n}$. %($\mathbb{R}^*$ is the set of non-negative real numbers). 
%Then the inference on $C$ 
%can be concluded using Bayes Rule: $P_t(C|T) = \frac{P(T|C) P_0(C)}{P(T)}$.

% \pw{How do we justify the assumption that $T$ is optimal, by assuming the underlying model is cooperative?}
% \pw{directly go for EOT or OT}

%Assuming the $T$ is obtained by minimizing the total cost, we may 
\textbf{Entropy regularized optimal transport} is an fundamental building block in modelling $P(C|T)$. 
Given a cost matrix $C$ and distributions $\mu,\nu$ over $X$ and $Y$ respectively, 
\textit{Entropy regularized optimal transport} (EOT) \citep{cuturi2013sinkhorn,peyre2019computational} solves the optimal coupling $T^{\lambda}$ that 
minimizes the entropy regularized cost of transfer $X$ with distribution $\mu$ into $Y$ with distribution $\nu$. 
Thus for a parameter $\lambda >0$: 
\begin{equation}\label{eq:eot}
T^{\lambda} = \text{argmin}_{T\in U(\mu, \nu)} \{\langle C, T\rangle - \frac{1}{\lambda} H(T)\},
\end{equation}
\noindent where $U(\mu, \nu)$ is the set of all couplings between 
$\mu$ and $\nu$ (i.e. joint distributions with marginals $\mu$ and $\nu$),
$\langle C, T\rangle = \sum_{i\in X, j\in Y} c_{ij}t_{ij}$ is 
the inner product between $C$ and $T$, and $H(T) := - \sum_{i=1}^{m}\sum_{j=1}^{n} t_{ij} \log t_{ij}$ 
is the \textit{entropy} of $T$.
$T^{\lambda}$ is called an \textit{optimal coupling} with parameter $\lambda$. 

\textbf{Sinkhorn scaling} is used to efficiently compute optimal couplings.
$(\mu,\nu)$-\textit{Sinkhorn Scaling (SK)} \citep{sinkhorn1964relationship} of a matrix M 
is the iterated alternation between row normalization of $M$ with respect to $\mu$ and 
column normalization of $M$ with respect to $\nu$. \citep{cuturi2013sinkhorn} proved that: 
given a cost matrix $C$, the optimal coupling $T^{\lambda}$ between $\mu$ and $\nu$ 
can be solved by applying $(\mu, \nu)$-Sinkhorn scaling on 
the \textit{negative exponential cost matrix} $K^{\lambda}$,
where $K^{\lambda} = e^{-\lambda C} = (e^{-\lambda\cdot c_{ij}})_{m\times n}$ (See Example~\ref{eg: sk}).
% \pw{include an example of Sinkhorn?}
% \pat{yeah. i think this will help when we need to define the inverse.}

% \begin{example}\label{eg: sk}
% Let $\mu =\nu=(\frac{3}{8},\frac{5}{8})$, 
% $\tiny C = \begin{pmatrix} \ln 1 & \ln 2 \\
%  \ln 4 & \ln 1 \end{pmatrix}$. 
% For $\lambda=1$, we may obtain $T$ by applying SK scaling on
% $\tiny K^{\lambda} = e^{-C} = \begin{pmatrix} 1 & 1/2\\ 1/4 & 1 \end{pmatrix}$:
% first, row normalize $K^{\lambda} $ with respect to $\mu$ results:
% $\tiny K_1^{\lambda} =  \begin{pmatrix} 1/4 & 1/8\\ 1/8 & 1/2 \end{pmatrix}$. 
% Then column normalization of $ K_1^{\lambda}$ with respect to $\nu$ outputs 
% $\tiny K_2^{\lambda} = \begin{pmatrix} 1/4 & 1/8\\ 1/8 & 1/2 \end{pmatrix}$. 
% As $K_1^{\lambda}= K_2^{\lambda}$, the SK scaling has converged with $T = K_1^{\lambda}$.
% In general, multiple iterations may be required to reach the limit. \pw{may move to apdx}
% \end{example} 

Given a coupling $T$, the inference on the cost matrix $C$
can be formulated as an inverse entropy regularized optimal transport problem. 
%can be interpreted as map from an optimal coupling $T^{\lambda}$ to the set of all possible triples of cost and marginals $\{(\widehat{C}, \widehat{\mu},\widehat{\nu})\}$. 
Given $\lambda$, EOT is the map $\Phi$:$(\overline{\mathbb{R}^*})^{m\times n} \times \mathcal{P} (\mathbb{R}^{m}) \times 
\mathcal{P} (\mathbb{R}^{n}) \to \mathcal{P} ((\mathbb{R}^*)^{m\times n})$ with $\Phi(C, \mu, \nu) = T^{\lambda}$.
Therefore \textbf{inverse entropy regularized optimal transport (IOT)} can be defined as $\Phi^{-1}: T^{\lambda} \to \{(C, \mu,\nu)\}$,
where $\{(C, \mu,\nu)\}$ represents the preimage set of $T^{\lambda}$ under $\Phi$.

\begin{remark}
In IOT, when $T^{\lambda}$ is observed without any noise,
$T^{\lambda}$ completely determines $\mu$ and $\nu$. 
Indeed $(C, \mu,\nu) \in \Phi^{-1}(T^{\lambda})$ implies that
$T^{\lambda} \in U(\mu,\nu)$. Thus $\mu$ and $\nu$ must equal the row and column marginals of $T^{\lambda}$.
Hence we may reduce IOT into $\Phi^{-1}: T^{\lambda} \to \{C\}$. 

Further, note that $\Phi^{-1}: T^{\lambda} \to \{C\}$ 
is not well-defined as a function, 
since $T^{\lambda}$ does not uniquely determine cost $C$. 
For instance, 
any two cost matrices differ by an additive constant 
output the same optimal coupling.
% For instance, any matrix $C'$ in form of 
% $\tiny C' = \begin{pmatrix} \ln \alpha \cdot 1 & \ln \alpha \cdot 2 \\
%  \ln \beta \cdot 4 & \ln \beta \cdot 1 \end{pmatrix}$, for $\alpha, \beta>0$
% outputs the same optimal coupling $T$ as $C$ in Example~\ref{eg: sk} \pw{need to modify if example is moved.}
%\pat{insert example here too? build off the SK one}
\end{remark}

\begin{remark}
The inverse problem is constructed to entropy regularized OT rather than OT because:
OT plan is general sparse (deterministic), and zero elements do not carry sufficient information to apply any
meaningful inference on the cost. In particular, zero in the
plan does not necessarily imply that the corresponding cost
is infinity. Moreover, EOT takes into account of the uncertainty
and incompleteness of observed data, and is faster \cite{courty2016optimal}. 
In addition, when $\lambda$ goes to $0$, EOT recovers OT.
\end{remark}
%\pw{we probably may move the next two paragraphs into intro.}

\textbf{Related Work.} 
The forward OT problem has a long history and is an active field of research 
\citep{kantorovich1942translocation, villani2008optimal,peyre2019computational,genvay2016stochastic}.
Applications in machine learning include: 
supervised learning \citep{frogner2015learning}, Bayesian inference \citep{el2012bayesian}, generative models \citep{salimans2018improving},
Transfer learning \citep{courty2016optimal}, and NLP \citep{alaux2018unsupervised}. 

However, the inverse OT problem is much less studied. The few existing inverse OT papers \citep{dupuy2019estimating, li2019learning, stuart2020inverse,ma2020learning} 
to our best knowledge, all 
focused on obtaining a unique approximation of the `ground truth' $C^*$, even though $\Phi^{-1}(T)$ contains a set of cost matrices.
To output an estimate of $C^*$, existing approaches place constraints on the cost matrix. 
For example in \citet{stuart2020inverse}, $C$ has to be either a Toeplitz matrix or
determined by a given graph structure.
In \citet{li2019learning}, $C$ must be constructed from a metric over a space where both $X$ and $Y$ are able to embedded in, plus $X$ and $Y$ need to have the same dimensionality. In \citet{dupuy2019estimating}, the affinity matrix has a low-rank constraint. In \citet{ma2020learning}, a regularization term on the cost is added to the objective to express the constraints such as symmetric costs.
However, these assumptions do not hold in general.

Rather than adding another constraint, 
we aim to analyze the intrinsic properties and 
the underlying geometric structure of the entire set $ \{C\} = \Phi^{-1} (T^{\lambda})$ 
for the discrete case in the following sections.
%
% Other related approaches, largely the lovely work of Curturi and colleagues, formulate the problem as ground metric learning. 
% \citep{cuturi2014ground} proposed algorithms for Ground Metric Learning (GML) which learns a distance from labeled normalized histograms given a criterion.
% \citep{heitz2021ground} extended the work of GML to learn the ground metric where only a geodesic distance on a graph is considered.
% \citep{paty2020regularized} re-interpreted regularized OT problems as ground cost adversarial and solved for robust solutions under different regularization. 
%
To simplify the notation, we will omit the superscript $\lambda$,
and fix $\lambda = 1$ unless otherwise stated.
% Some results will be described in term of negative exponential cost matrix $K$. 
% Corresponding statements in terms of $C$ can be obtained by $K = e^{-C}$.
% % Prior $P_0(K)$ is set to be the distribution where each column is follows symmetric Dirichlet distribution with hyperparameter $\alpha$ independently.
% Since for any non-negative cost matrix $C$, elements in $K$ are in the range of $(0,1]$, which can be cross-ratio equivalently 
% converted into a $K$ in the prior by column normalization.
% Observed $T$ is assumed to be a positive matrix, the case when $T$ is non-negative can be derived using  machinery of \textit{maximal partial pattern} 
% developed in \citep{pei2019generalizing}.

\section{Probabilistic Inverse Optimal Transport} \label{sec:piot}

%\pat{remind people of our goal?}-\pw{added}
%In this section, we formalize the general IOT problem through Bayesian inference, 
%analyze key properties of $\Phi^{-1}(T)$ and reveal its underlying geometric structure. 

\begin{definition}\label{def: PIOT}
Let $T$ be a noisy observation of the optimal coupling~$T^*$.
\textbf{Probabilistic inverse optimal transport} is defined as: 
\begin{equation} \label{eq: PIOT}
 P(C|T) = \int _{T^* \in (\mathbb{R}^*)^{m\times n}} P(C|T^*) P(T^*|T) d T^*,    
\end{equation}
where $\displaystyle P(C|T^*) $ is obtained through Bayes' Rule:
\begin{equation} \label{eq: bayes}
 P(C|T^*)  = \frac{P(T^*|C) P_0(C)}{P(T^*)},   
\end{equation}
$P_0(C)$ is the prior of cost $C$; 
%over $(\overline{\mathbb{R}^*})^{m\times n}$; 
Likelihood $P(T|C) = 1$ if $C\in \Phi^{-1}(T)$, otherwise $P(T|C) = 0$;
$P(T^*) = \int P(T^*|C) P_0(C) \text{d} C$ is the normalizing constant;
$P(K|T)$ is used to denote the induced posterior on $K = e^{-C}$.
\end{definition}

\subsection{No observation noise}
In this section, we study the case when $T^* = T$.
As a direct application of Definition~\ref{def: PIOT}, we have:
\begin{proposition}\label{prop:piot}\footnote{Proofs for all results are included in Appendix \ref{apdx:proofs}.}
$ P(C|T)$ is supported on the intersection between $\Phi^{-1} (T)$ and the domain of $ P_0(C)$. Moreover, we have that 
$\displaystyle P(C|T) = \frac{P_0(C)}{\int_{\Phi^{-1} (T)} P_0(C) \text{d} C}$.
\end{proposition}

Therefore, characterization of $\Phi^{-1} (T)$ is essential to understand
the manifold on which the posterior distribution $P(C|T)$ is supported.

Notice that for any $C \in \Phi^{-1} (T)$, 
$T$ is obtained from a Sinkhorn scaling of $K = e^{-C}$, and 
each step of Sinkhorn Scaling is a normalization which equivalent to 
multiplication of a diagonal matrix. We show that: 

\begin{proposition}\label{prop:DAD}
Let $T$ be a non-negative optimal coupling of dimension $m\times n$. $C \in \Phi^{-1} (T)$ if and only if for every $\epsilon > 0$,
there exist two positive diagonal matrices $D^r = \diag\{d^r_1, \dots, d^r_m\}$ and $D^c =\diag \{d^c_1,\dots, d^c_n\}$ such that: 
$\displaystyle |D^r K D^c - T| < \epsilon$, where $K = e^{-C}$ and $|\cdot|$ is the $L^1$ norm. 
In particular, if $T$ is a positive matrix, then $C \in \Phi^{-1} (T)$ 
if and only if there exist positive diagonal matrices $D^r,D^c$ such that $D^r K D^c = T$, i.e.
\begin{equation}\label{eq:diag_equiv}
\Phi^{-1} (T) = \{C| \exists  D^{r} \text{ and } D^{c} \text{ s.t. } K = D^rTD^c\}.   
\end{equation}

\end{proposition}

Proposition~\ref{prop:DAD} identifies a key feature of matrices in $\Phi^{-1} (T)$. 
However, it is not easy to
verify whether an arbitrary matrix belongs to $\Phi^{-1} (T)$ nor reveal any underlying 
geometric structure of $\Phi^{-1} (T)$.

Towards these goals, we now introduce an equivalent condition: \textit{cross ratio equivalence} between \textit{positive} matrices, 
and show that the IOT set $\Phi^{-1} (T)$ can be completely characterized by $T$'s \textit{cross ratios}. 

\begin{definition}
Let $A, B$ be positive $m\times n$ matrices. $A$ is \textbf{cross ratio equivalent} to $B$, 
denoted by $A\overset{c.r.}{\sim} B$,
if all cross ratios of $A$ and $B$ are the same, i.e.
$$r_{ijkl}(A) := \frac{a_{ik} a_{jl}}{a_{il} a_{jk}} = \frac{b_{ik} b_{jl}}{b_{il} b_{jk}} := r_{ijkl}(B) $$
holds for any $i, j \in \{1, \dots, m\}$ and $k, l \in \{1, \dots, n\}$. 

\end{definition}

\begin{lemma}\label{lemma:cr_dr}
For two positive matrices $A, B$, 
$A \overset{c.r.}{\sim}B$ if and only if 
there exist positive diagonal matrices $D^r$ and $D^c$ such that $A = D^r B D^c$.
\end{lemma}

\begin{theorem}\label{thm:IOT}
Let $T$ be an observed positive optimal coupling of dimension $m\times n$. 
Then $\Phi^{-1} (T)$ is a hyperplane of dimension 
$m+n-1$ embedded in $ (\mathbb{R}^*)^{m\times n}$, which consists all the cost matrices 
that of the form:
\begin{equation}\label{eq:cr_equiv}
 \Phi^{-1} (T) = \{C \in (\mathbb{R}^*)^{m\times n} | K= e^{-C} \overset{c.r.}{\sim} T\}.    
\end{equation}

\end{theorem}

Cross ratio equivalence provides a strong connection between IOT and algebraic geometry.
Algebraically, cross ratios generate the set of algebraic invariants of matrix scaling.
The set of all scale reachable matrices from a coupling $T$, 
i.e. $\Phi^{-1}_K(T) \coloneqq \{K| D^{r}TD^{c}\} = \{K = e^{-C}| C\in \Phi^{-1}(T)\}$ forms an algebraic variety defined by the cross-ratios of $T$.
% For instance: the matrices $\tiny K = \begin{pmatrix} a & b \\ c & d \end{pmatrix} \in \mathcal{K}(T) $ for
% $\tiny T = \begin{pmatrix} 0.1 & 0.2 \\ 0.3 & 0.4 \end{pmatrix}$ (i.e. the set of possible negative exponential cost matrix) 
% are generic solutions to the equation $3ad-2bc=0$. 
Geometrically, the set $\Phi^{-1}_K(T)$ forms a special manifold \citep{fienberg1968geometry}.
$T$ is the unique intersection between $\Phi^{-1}_K(T)$ 
and the hyperplane determined by the linear marginal conditions.

\begin{remark}
Since a matrix $T$ and its normalization $T/\sum_{ij}(t_{ij})$ 
have the same cross ratios (hence have the same image under IOT),
we may relax the distribution constraint on the observed coupling $T$.
In this case, instead of probability, 
$t_{ij}$ represents the frequency of observing $x_i$ and $y_i$ simultaneously.
\end{remark}

\begin{example}\label{eg:hyperplane}
Let $\tiny T = \begin{pmatrix} 1 & 2 & 3\\ 2 & 3 & 1\end{pmatrix}$. 
$T$ has three cross ratios $r_{1212} (T) = \frac{3}{4}, r_{1213}(T) =\frac{1}{6} , r_{1223}=\frac{2}{9}$.
Only two are independent as $r_{1213}(T) = r_{1212}(T) * r_{1223} (T)$. Hence $\Phi^{-1}_K(T)$ 
is the solution set of the algebraic equation:
\begin{equation*}
\begin{aligned}[c]
\footnotesize
    \begin{cases}
      k_{11} k_{22}/ k_{12}k_{21} = 3/4\\
      k_{12} k_{23}/k_{13}k_{22} = 2/9
    \end{cases}
\end{aligned}
\Longleftrightarrow \hspace{0.08in}
\begin{aligned}[c]
\footnotesize
    \begin{cases}
      4 k_{11} k_{22} - 3 k_{12}k_{21} = 0\\
      9 k_{12} k_{23} - 2 k_{13}k_{22} = 0.
    \end{cases}
\end{aligned}
\end{equation*}
Further as $k_{ij} = e^{-c_{ij}}$, we obtain that $\Phi^{-1}(T)$ is the 4-dim hyperplane defined by the linear equations:
% \begin{subequations} \label{eq:T_1}
% \small
%   \begin{align}
%     & c_{12}+c_{21} - c_{11}-c_{22} = \small{\ln 3/4} \label{eq:T_1a}\\
%     & c_{13}+ c_{21} - c_{11} - c_{23} = \ln 1/6 \label{eq:T_1b}
%   \end{align}
% \end{subequations}
\begin{subequations} \label{eq:T_1}
\footnotesize
  \begin{empheq}[left=\empheqlbrace]{align}
    c_{12}+c_{21} - c_{11}-c_{22} &= \ln (3/4) \label{eq:T_1a}\\
    c_{13}+ c_{22} - c_{12} - c_{23} &= \ln (2/9) \label{eq:T_1b}.
  \end{empheq}
\end{subequations}

% \begin{equation}\label{eq:T_1}
% \small
%     \begin{cases}
%       c_{12}+c_{21} - c_{11}-c_{22} = \ln\frac{3}{4}\\
%       c_{13}+ c_{21} - c_{11} - c_{23} = \ln \frac{1}{6}
%     \end{cases}
% \end{equation}
\end{example}

\begin{remark}\label{rmk:basis}
For a $m\times n$ matrix $A$, 
although there are $\binom{m}{2} \binom{n}{2}$ ways to choose $i,j,k,l$ to form cross ratios of $A$, 
not all are independent as seen in Example~\ref{eg:hyperplane}. 
A set of independent cross ratios that generates the entire collection of cross ratios is called a \textbf{basis}.
For example, it is easy to check that $\mathcal{B} = \{r_{1j1k}(A)| j = 2, \dots, m, k=2, \dots, n\}$ forms a basis for $A$.
\end{remark}

% \pw{The following paragraph may delete or simplified.}
% Another special case is when $T$ is a square matrix.
% Two square matrices are said to be \textit{\textbf{diagonal} ratio equivalent} 
% if the ratio between any two positive diagonals of the two matrices are the same (See definition and proofs in supplemental materials.).
% \citep{pei2019generalizing} showed that when $T$ is a square matrix, any $C \in \Phi^{-1} (T)$ is diagonal ratio equivalent to $T$.
% In the supplementary material, we prove that for positive square matrices, \textbf{diagonal} ratio equivalent coincides with cross ratio equivalent.
Observed $T$ is assumed to be a positive matrix for the rest of the paper,
the case when $T$ is non-negative can be derived using  machinery of \textit{maximal partial pattern} 
developed in \citep{pei2019generalizing}.

\textbf{Support manifold.} 
%Now we are ready to study the support manifold 
%of posterior distribution $P(C|T)$.
Since $\text{Supp}[P(C|T)] = \Phi^{-1} (T) \cap \text{Domain}[P_0(C)]$, 
different priors would add different constraints on $C$, which will result different submanifolds embedded in 
the hyperplane formed by $\Phi^{-1} (T)$.

For instance, suppose $P_0(C)$ follows a Dirichlet distribution over the entire matrix ~($\mathbf{P_1}$). 
Then 
$$\small \text{supp}[P(C|T)]  =\{C \in (\mathbb{R}^*)^{m\times n} | e^{-C} \overset{c.r}{\sim} T, \sum_{ij} c_{ij} = 1\}.$$
Notice that $K$ encodes the cross ratios of $T$, and to avoid the estimation of $\lambda$,
it is handy to directly put prior over $K=e^{-C}$. 
For example, assume that $P_0(K)$ is the 
distribution where each column follows an independent Dirichlet distribution~($\mathbf{P_2}$) with hyperparameter $\alpha$.
% According to Proposition~\ref{prop:DAD}, we have:
% \[\Phi^{-1} (T) = \{C| \exists  D^{r} \text{ and } D^{c} \text{ s.t. } K = D^rTD^c\}\]
Then the domain of $P_0(K)$ is $\mathcal{P}(\mathbb{R}^{n})^{m}$. 
Therefore, based on Eq.~\eqref{eq:diag_equiv}, we have the intersection is of the form:
\begin{equation} \label{eq:supp_posterior}
    \text{supp}[P(C|T)] = \{ C | \exists D^{r} \text{ s.t. } K = \text{Col}(D^{r}T) \}.
\end{equation}

% Since for any non-negative cost matrix $C$, elements in $K$ are in the range of $(0,1]$, which can be cross-ratio equivalently 
% converted into a $K$ in the prior by column normalization.
%\centerline{$\d
% %isplaystyle \text{supp}[P(C|T)] = \{ \widehat{C} \in (\mathbb{R}^{+})^{m\times n} | \exists D^{r} \text{ s.t. } \widehat{C} \text{ is column normalization of } D^{r}T \}$ }
\begin{remark}\label{rmk: one_free_column}
Eq.~\eqref{eq:supp_posterior} indicates that: \textbf{(i)} a known column of $K$ uniquely determines all the other columns. 
Indeed, denote the $j$-th column of $K, T$ by $\mathbf{k}_j, \mathbf{t}_j$. 
For a known column $\mathbf{k}_j$, let $d_i = k_{ij}/t_{ij} $ and $D^{r} = \diag\{d_1, \dots, d_m\}$. 
Then the $l$-th column $\mathbf{k}_l$ equals to the column normalization of $D^{r} \mathbf{t}_l$.
\textbf{(ii)} for any $\mathbf{v} \in \mathcal{P}((\R^+)^m)$, there exists a $K \in  \text{supp}[P(K|T)]$ such that $\mathbf{k}_j = \mathbf{v} $.
Therefore, we have:

\end{remark}

\begin{corollary}\label{cor:proj_single_column}
Under prior $\mathbf{P_2}$,
the projection of $\text{supp}[P(K|T)]$ onto each column is a $(m-1)$-dimensional manifold that is homeomorphic to the simplex $\Delta_{m-1}$. 
\end{corollary}
\textbf{Subspace of $\text{Supp}[P(C|T)]$.} \label{para:subspace}
Let $C_s$ be the submatrix of the cost $C$ 
that corresponds to $X_s\times Y_s$, where $X_s = \{x_1, \dots, x_{s_1}\} \subset X$, $Y_s = \{y_1, \dots, y_{s_2}\}\subset Y$.
In this section, we characterize the support of $P(C_s|T)$ as a subspace of $\text{Supp}[P(C|T)]$.

Denote the sub-coupling of $T$ corresponding to $C_s$ by $T_s$.
According to Eq.~\eqref{eq:diag_equiv} and Eq.~\eqref{eq:cr_equiv}, 
the projection of $\Phi^{-1}(T)$ onto the cost over $X_s\times Y_s$
is in form of:
\begin{align*}
 \Phi^{-1}_s(T) &:= \{C_s | C_{s} \text{ is submatrix of a } C \in \Phi^{-1}(T)\} \\
 & = \{C_s| \exists D^r_s, D^c_s \text{ s.t. } K_s = D^r_s T_s D^c_s \}\\
 & = \{C_s|  K_s=e^{-C_s},  K_s \overset{r.c.}{\sim} T_s\}.
\end{align*}
$\text{Supp}[P(C_s|T)]$ is determined by  $\Phi^{-1}_s(T)$ and $P_0(C)$.
Intuitively, $\text{supp}[P(C_s|T)]$ contains all the matrices $C_s \in  \Phi^{-1}_s(T) $ such that there exists a proper extension of 
$D^r_s, D^c_s$ to $D^r, D^c$ such that $D^rTD^c$ is in the domain of $P_0(C)$. 

Take prior $\mathbf{P_2}$ as an example. Since the domain for a column of $K$ is a copy of $\Delta_{m-1}$ as shown in Corollary~\ref{cor:proj_single_column}, 
the domain for a column of $K_s = e^{-C_s}$ is then
a copy of $\Sigma_{s_1} = \{\mathbf{k} = (k_1, \dots, k_{s_1}) \in (\mathbb{R}^+ )^{s_1}| k_{1}+\dots +k_{s_1} < 1\}$. 
Hence, $\text{supp}[P(C_s|T)] $ should be contained in the set :
\begin{align*}
\mathcal{W}&  = \{C^s |\exists \text{ diagonal matrices } D_s^r, D_s^c \text{ s.t. }  \\
& K_s = D_s^rT_sD_s^c  \text{ and } \mathbf{1}_{s_1} K_s = (\alpha_1, \dots, \alpha_{s_2}), \alpha_i < 1 \}.   
\end{align*}

It is important to note that $\text{supp}[P(C_s|T)]$ is a proper subset of $\mathcal{W}$. 
As $K_s$ is a submatrix of $K$, the choice of $D_s^c$ must form a proper column normalizing constant of a $D^rT$. 
Thus, the choice of $D_s^c = \diag \{d^c_1, \dots, d^c_{s_2}\}$ must guarantee that there exist an extension of $D_s^r$, 
denoted by $D'^{r} = \diag\{D_s^r,d^r_{s_1 +1}, \dots, d^r_{m}\}$, with $d^r_{s_1 +1}, \dots, d^r_{m}>0$, 
such that the $j$-th column sum of $D'^{r} T$ is $1/d^c_j$.
Let $T_{m-s}$ be the submatrix of $T$ corresponding to $\{x_{s_1+1}, \dots, x_{m}\} \times Y^{s}$.  %We show that:
\begin{proposition}\label{prop:submanifold}
$C_s \in \text{supp}[P(C_s|T)]$ if and only if there exists positive diagonal matrices $D_s^c, D_s^r$ such that $K_s = D_s^c T_s D_s^r$ and
the system of equations 
$(x_1, \dots, x_{m-s_1}) T_{m-s}  = (1/d^c_1, \dots, 1/d^c_{s_2}) - \mathbf{1}_s D_s^rT_s$ have a positive solution for $\{x_1, \dots, x_{m-s_1}\}$.
\end{proposition}

\textbf{Incomplete Observation.}
\label{para:missing_element}
%We characterize the support of $P(C|T)$, when $T$ has a missing element.
Say $T$ is observed with $t_{m1}$ missing. Then generically, we have: 

\centerline{$\text{supp}[P(C|T)] = \cup_{\widehat{T} \in U(T)} \text{supp}[P(C|\widehat{T})]$,}
where $U(T)$ is the set of matrices with $T$ completed by a choice of $t_{m1}>0$.
Given a prior, say $\textbf{P1}$, we have: 

\vspace{0.1in}
\centerline{$\text{supp}[P(C|T)]  =\{C | \exists t_{m1} >0 \text{ s.t. } K \overset{c.r}{\sim} \widehat{T}, \sum_{ij} c_{ij} = 1\}$,}

where $\widehat{T}$ is $T$ with $t_{m1}$ as its $m1$-th element. With prior $\textbf{P2}$, according to \eqref{eq:supp_posterior} we have:
$$\text{supp}[P(C|T)] = \{ C | \exists D^{r}, t_{m1} >0, \text{ s.t. } K = \text{Col}(D^{r}\widehat{T}) \}.$$

Moreover, according to Remark~\ref{rmk: one_free_column}, 
a known column $\mathbf{k}_l$ of $K$ completely determines 
other fully observed columns. Whereas, for a column with a missing element, say the first column, 
$\mathbf{k}_l$ determines an 1-dim set $\mathcal{K}_1$ for $\mathbf{k}_1$.
% forms a line segment in $\Delta_{m-1}$.
In particular, let $D^{r} = \diag \{d_1, \dots, d_m\}$, where $d_i = k_{il}/t_{il}$.
%We show:
%$D^{r} = \diag \{k_{1l}/t_{1l}, \dots, k_{ml}/t_{ml}\}$
\begin{corollary}\label{cor:missing_element}
Under prior $\mathbf{P_2}$, 
$\mathcal{K}_1$ is a line segment in $\Delta_{m-1}$ that can be parameterized as:
\centerline{$\mathcal{K}_1 = \{ (d_1t_{11}, \dots, d_m t_{m1})/\sum_{i = 1}^{m} d_it_{i1}| t_{m1} \in (0,\infty) \}$.}
\end{corollary}

% Then the line segment can be parameterized as 
% $\mathcal{K}_1 = \{ (d_1t_{11}, \dots, d_m t_{m1})/\sum_{i = 1}^{m} d_it_{i1}| t_{m1} = \widehat{t}_{m1} \in (0,\infty) \}$.

Note that for different choice of priors, $\mathcal{K}_1$ could be a curve instead of a line segment. 
Denote the prior imposed constraint on $\mathbf{k}_1$ by $f_1(\mathbf{k}_1) = 0$,
the ratio imposed linear constraint on $\mathbf{k}_1$ by $f_i(\mathbf{k}_1) = 0$, where $f_i(\mathbf{k}_1) = k_{11}/k_{i1} - d_1*t_{11}/(d_i*t_{i1})$,
for $i = 2, \dots, m-1$.
Then $\mathcal{K}_1$ is the solution set for system of equations $\{f_1, f_2, \dots, f_{m-1}\}$.
With Dirichlet prior, $f_1$ is also linear, so $\mathcal{K}_1$ is a segment.

\subsection{With observation noise}
%In this section, we study the general case when there are observation noises. 

%First, f
For any two observed couplings with the same dimension, we introduce a natural distance between 
their images under IOT.

\begin{corollary}\label{cor:parallel}
Let $T_1, T_2$ be two positive matrices of dimension $m \times n$.
The hyperplanes $\Phi^{-1}(T_1)$ and $\Phi^{-1}(T_2)$, have the same normal direction.
In particular, if $T_1 \overset{c.r.}{\sim}T_2$ then $\Phi^{-1}(T_1) = \Phi^{-1}(T_2)$.
Otherwise $\Phi^{-1}(T_1)$ is parallel to $\Phi^{-1}(T_2) $.
\end{corollary}

Therefore, $\Phi^{-1}(T_1) , \Phi^{-1}(T_2)$ are the same or parallel hyperplanes
embedded in $(\mathbb{R}^{+})^{m\times n}$. \textbf{The distance between IOT of $T_1$ and $T_2$}, 
denoted by $d(\Phi^{-1}(T_1) , \Phi^{-1}(T_2))$ 
is then well-defined to be the Euclidean distance between $\Phi^{-1}(T_1) $ and $\Phi^{-1}(T_2)$.

Moreover, $\Phi^{-1}(T) $ is completely determined by $T$'s cross ratios.
So $d(\Phi^{-1}(T_1) , \Phi^{-1}(T_2))$ can be expressed in term of cross ratios of $T_1$ and $T_2$.
We now illustrate how to obtain $d(\Phi^{-1}(T_1) , \Phi^{-1}(T_2))$.
\begin{example}
Let $T_1 = T$ in Example~\ref{eg:hyperplane}, 
$\tiny T_2 = \begin{pmatrix} 1 & 2 & 3\\ 3 & 2 & 1\end{pmatrix}$.
Then $\Phi^{-1}(T_1)$ is the hyperplane show in~Eq.~\eqref{eq:T_1}.
Taking $r_{1212}, r_{1213}$ as $T$'s cross ratio basis, 
then we have $\Phi^{-1}(T_2)$ is: 
\begin{subequations} \label{eq: T_2}
\small
  \begin{empheq}[left=\empheqlbrace]{align}
    c_{12}+c_{21} - c_{11}-c_{22} &= \ln (1/3) \label{eq:T_2a}\\
    c_{13}+ c_{22} - c_{12} - c_{23} &= \ln (1/3) \label{eq:T_2b}.
  \end{empheq}
\end{subequations}
%Eq~\eqref{eq:T_1} and Eq~\eqref{eq:T_2} have the same coefficients, 
Both $\Phi^{-1}(T_1)$ and $\Phi^{-1}(T_2)$ are intersections of two 5-dim hydroplanes 
(i.e.\eqref{eq:T_1a} $\cap$ \eqref{eq:T_1b}, \eqref{eq:T_2a} $\cap$ \eqref{eq:T_2b})
with the same coefficient, so they are are parallel. 
Let the Dihedral angle between \eqref{eq:T_1a} and \eqref{eq:T_1b} be $\theta$.
It is easy to check $\cos \theta = 1/2$. 
So %by law of cosines, 
we have \scalebox{0.9}{$d(\Phi^{-1}(T_1), \Phi^{-1}(T_1) ) = \sqrt{\ln^2 \frac{4}{9} + \ln^2 \frac{3}{2} -\ln \frac{3}{2} \ln \frac{4}{9}}$}.
%$d(\Phi^{-1}(T_1), \Phi^{-1}(T_1) ) = \sqrt{(\ln 1/3 - \ln3/4)^2 + (\ln 1/6 - \ln1/9)^2 - 2(\ln 1/3 - \ln3/4)(\ln 1/6 - \ln1/9) \cos \theta } $
\end{example}

\begin{remark}
The hyperparameter $\lambda$ in Eq.\eqref{eq:eot} can be viewed as a greedy data selection. 
For a forward EOT problem, $\lambda$ raises the cross ratios of $K=e^{-C}$ to a power of $\lambda$.
Hence, $\lambda$ either exaggerates or suppresses the cross ratios of $T$, 
depending on whether $\lambda$ is greater or less than $1$ \citep{wang2020mathematical}.
Conversely, given a backward IOT problem, for a fixed coupling $T$, the larger the $\lambda$ is assumed, 
the smaller the cross ratios of $K$ are. 
Moreover, for a pair of observed couplings $T_1$ and $T_2$, 
the equations for hyperplanes corresponding $\Phi^{-1} (T_1), \Phi^{-1} (T_2)$ with different choices of $\lambda$, only differ 
on the constants by a scalar, i.e. $d_{\lambda}(\Phi^{-1} (T_1), \Phi^{-1} (T_2)) = d(\Phi^{-1} (T_1), \Phi^{-1} (T_2))/\lambda$.
Therefore, the larger $\lambda$ is assumed, the smaller the distance between $\Phi^{-1} (T_1)$ and $\Phi^{-1} (T_2)$ are.
\end{remark}

Assuming uniform prior $P_0(C)$ over $(\mathbb{R}^*)^{m\times n}$, 
we now investigate $P(C|T)$ under two common noise types. 
Results for other specific priors can be obtained by restricting the generic results to the prior's domain.
To simplify the notation, without loss, 
we will assume noise only occurs on one element of $T$, say $t_{11}$, 
as the general case may be treated as compositions of such.

\textbf{Bounded noise.} 
% Consider the case where elements of the noisy observation $T$ 
% is within a bounded distance away from the optimal coupling $T^*$.
Suppose that $t_{11}$ is perturbed 
by a uniform noise from $t^*_{11}$, i.e. $t^*_{11} = t_{11} + \epsilon$, 
where $\epsilon$ is a random variable with uniform distribution over $[-a, a]$, for $a>0$.
%Then according to Eq~\ref{eq: PIOT}, assuming uniform prior $P_{0}(C)$,
% $P(C|T)$ is a uniform distribution over it support.

Theorem~\ref{thm:IOT} and Corollary~\ref{cor:parallel} imply that 
$P(C|T)$ is supported on a collection of parallel hyperplanes 
within bounded distance from $\Phi^{-1}(T)$. 
%More precisely, we show:

\begin{proposition} \label{prop: bounded_noise}
For a coupling $T$,  assume uniform observation noise on $t_{11}$ with bounded size $a$, then
$$\text{supp}[P(C|T)] = \cup_{T'\in \mathbb{B}_{a}(T)} \Phi^{-1}(T'),$$
where $\mathbb{B}_{a}(T)$ is the set of matrices $T'$ of the same dimension as $T$ with the property that:
$t'_{11}>0$, $|t'_{11} - t_{11}| \leq a$ and $t'_{ij} = t_{ij}$ for other $i, j$.
Moreover, $\Phi^{-1}(T')$ can be expressed as intersection of two hyperplanes (may be in different dimensions):
one with equation: $c_{11} + c_{22} - c_{21} - c_{12} = - \ln\frac{t'_{11} t_{22}}{t_{21}t_{12}}$, 
and the other equation does not depend on the value of $t'_{11}$. Assume the angle between these two hyperplanes is $\theta$.
Then $d(\Phi^{-1}(T_1'), \Phi^{-1}(T_2')) \leq \ln \frac{t_{11} + a}{t_{11} - a} /\sin \theta $, for $T_1', T_2' \in \mathbb{B}_{a}(T)$.
\end{proposition}

\begin{example}\label{eg:bounded_noise}
Let $T$ be the same as Example~\ref{eg:hyperplane}.
Suppose there is a bounded noise $\epsilon$ on $t_{11}$ of size $a>0$.
Hence, $\text{supp}[P(C|T)]$ is the union of a collection of hyperplanes determined by $\mathbb{B}_{a}(T)$ 
of the following form, where $\epsilon \in [-a, a]$.
\begin{subequations} \label{eq:T_3}
\small
  \begin{empheq}[left=\empheqlbrace]{align}
    c_{12}+c_{21} - c_{11}-c_{22} &= \ln [(3+3\epsilon)/4] \label{eq:T_3a}\\
    c_{13}+ c_{22} - c_{12} - c_{23} &= \ln (2/9) \label{eq:T_3b}
  \end{empheq}
\end{subequations}
Let the angle between \eqref{eq:T_3a} and \eqref{eq:T_3b} be $\theta$.
Computation shows that $\sin \theta = \sqrt{3}/2$, Hence for $T_1', T_2' \in \mathbb{B}_{a}(T)$, 
$d(\Phi^{-1}(T_1'), \Phi^{-1}(T_2')) \leq \sqrt{3}\ln (1+a)$.
\end{example}

\textbf{Gaussian noise.} Suppose that $t_{11}$ is perturbed 
by a Gaussian noise from the ground truth $t^*_{11}$, i.e. $t^*_{11} = t_{11} + \epsilon$, 
where $\epsilon \sim \mathcal{N}(0, \sigma^2)$ with standard deviation $\sigma >0$.

Gaussian distribution is defined over the entire real line, 
% $P(T^*|T)$ is supported on the set $\mathcal{L}(T)$ consists matrices $T'$ of the same dimension of $T$ 
% with the propriety that:
% $t'_{ij} = t_{ij}$ for $(i, j) \neq (1,1)$. 
so $\text{supp}[P(C|T)] = \cup_{T'\in \mathbb{L}_{a}(T)} \Phi^{-1}(T')$,
where $\mathbb{L}(T)$ is the set of matrices $T'$ of the same dimension as $T$ with the property that:
$t_{11}>0$ and $t'_{ij} = t_{ij}$ for $(i, j) \neq (1,1)$.
Therefore, cross ratios of $T'$ that depend on $t_{11}$ can be any arbitrary number.
So similar to the case of missing elements in $T$, 
$P(C|T)$ is supported on a hyperplane that is one dimensional higher than $P(C|T^*)$.
%In particular, we show:

\begin{proposition} \label{prop: gaussian_noise}
Let $T$ be an observed coupling of dimension $m\times n$ with Gaussian noise on $t_{11}$.
Further, let $\mathcal{B}$ be a basis for cross ratios of $m\times n$ matrices, that contains only one cross ratio depending on $t_{11}$.
Eliminate the cross ratio depending on $t_{11}$ in $\mathcal{B}$, denote the new set by $ \mathcal{B}^{-}$. Then: 

\centerline{$\text{supp}[P(C|T)] = \{C|r(K) = r(T) \text{ for } r \in \mathcal{B}^{-}\}$}
In particular, $P(C|T)$ is supported on a hyperplane that is one dimensional higher than $P(C|T^*)$.
\end{proposition}

\begin{example}\label{eg:Gaussian_noise}
Let $T$ be the same as Example~\ref{eg:hyperplane}.
Suppose there is a Gaussian noise $\epsilon \sim \mathcal{N}(0, \sigma^2) $ on $t_{11}$.
Hence, $\text{supp}[P(C|T)]$ is the union of a collection of hyperplanes determined by $\mathbb{L}(T)$,
which are in the same form as Eq~\ref{eq:T_3}, with constraint on $\epsilon$ being:
% \begin{subequations} \label{eq:T_4}
% \small
%   \begin{empheq}[left=\empheqlbrace]{align}
%     c_{12}+c_{21} - c_{11}-c_{22} &= \ln [(3+3\epsilon)/4] \label{eq:T_4a}\\
%     c_{13}+ c_{22} - c_{12} - c_{23} &= \ln (2/9) \label{eq:T_4b}
%   \end{empheq}
% \end{subequations}
$3+3\epsilon >0$ and $\epsilon \sim \mathcal{N}(0, \sigma^2) $. Thus,
$\epsilon \in (-1, \infty)$, which further suggests no restriction is put on 
$c_{12}+c_{21} - c_{11}-c_{22}$. Hence:
\vspace{0.1in}
\centerline{$\text{supp}[P(C|T)] = \{C| c_{13}+ c_{22} - c_{12} - c_{23} = \ln \frac{2}{9} \}.$}
\end{example}

\begin{remark}
In this case, $P(C|T)$ is not uniform over its support. 
According to Eq~\eqref{eq: PIOT}, $P(C|T) \propto P(T_C|T)$ where $T_C \in \mathbb{L}(T) $ and $K = e^{-C} \overset{c.r.}{\sim}T_C$.
In particular, $P(C|T)$ has the highest probability on the hyperplane $\Phi^{-1}(T)$. 
If further constraints are added to the cost matrix $C$ such that 
there is a unique $C \in \Phi^{-1}(T)$ that meets such constraints, 
then one may obtain a maximum a posteriori estimation of $C$.
This explains why the simulations in \citep{stuart2020inverse}, failed to identify a unique cost matrix
for the general case:
 $P(C|T)$ is supported on an entire hyperplane. 
%In contrast, m
Whereas, maximum a posteriori estimations of $C$ were successfully obtained in both the 
graph and Toeplitz cases.% \citep{stuart2020inverse}. %\wt{Do they use MAP or just sampling the posterior?}
\end{remark}
% \pw{pictures within each subsection or all pictures in a simulation section?}
% \wt{Noise on marginals? Eg. when we do PIOT on NBA data, can we add noises on the original row marginals to make predictions? After we obtain the posterior distributions we samples sets of noisy row marginals, and then use the posterior as cost and noisy marginals to find OT plans. }

%%%%%%%%%%%%%%%%%%%%%%% experiment %%%%%%%%%%%%%%%%%%%%%%%

\section{Algorithm}

\textbf{Approximate inference via MCMC.}
Given a perfectly observed $T$ and $P_0(C)$, we propose a Markov Chain Monte Carlo (MCMC) method 
% based  on the Metropolis algorithm\citep{metropolis1953equation} 
for sampling $P(C|T)$ (see Algorithm 1).
%from the posterior distribution. 
%The cross-ratio of the generated cost matrices are the same as $T$. 
%
%The algorithm works as follows. %\footnote{We suppress the $\lambda$ superscript to simplify notation of iterations.} 
At iteration $i$, denote the current cost $C$ by $C^{(i)}$, $K^{(i)} = e^{-C^{(i)}}$.
We generate two vectors $D^r$ and $D^c$ with each element sampled from an exponential of a Gaussian distribution $\mathcal{N}(0,\sigma^2)$. 
$K^{(i+1)}$ is then obtained by $K^{(i+1)} = \diag{D^r}K^{(i)}\diag{D^c}$.
%and scale rows in $K^{(i)}$ by left multiplying $D^r$ and columns by right multiplying $D^n$ to give $K^{(i+1)}$. 
%We compute the probability of the cost matrix, $P($log$(K^{(i)} | \mu)$. 
The acceptance ratio, $r = \frac{P(-\ln (K^{(i+1)}) | T)}{P(-\ln(K^{(i)}|T)} \overset{(*)}{=} \frac{P_0(C^{i+1}, \beta)}{P_0(C^{i}, \beta)}$
%$r = \frac{P_0(C^{i+1})}{P_0(C^{i})} = \frac{P(\text{log}(K^{(i+1)}) | \theta)}{P(\text{log}(K^{(i)}) | \theta)}$, 
where $\beta$ is the hyperparameter of prior $P_0(C)$. %represents the parameters of the model. %\pw{check with wei-ting.}
Equality~$(*)$ holds because our proposal preserves the cross-ratios, so we are guaranteed to stay on the manifold of support. 
Therefore, $P(T|C)$ is always $1$ and is omitted (see Proposition~\ref{prop:piot}). 
We accept or reject the move by comparing the acceptance ratio to an uniform random variable, $u$, 
which concludes a single step of the MCMC algorithm. 
We refer this method as MetroMC.
%\pw{the algorithm should be on cost $C$, not $K$}
\begin{algorithm}[tb]
\caption{ MetroMC }\label{algo:MetroMC}
\begin{algorithmic}[1]
\INPUT{: coupling $T$, variance $\sigma$, prior parameters $\beta$, uniform random variable $u$ $\in$ [0, 1)}
\SET{: $C^{(0)}=-\ln(K^{(0)})$; $K^{(0)}=T_{m\times n}$}
\FOR{ i = 1 to ITERMAX} 
    \STATE $D^{r} =$ exp$( N(0, \sigma * \mathbb{I}_{m}) )$
    \STATE $D^{c} =$ exp$( N(0, \sigma * \mathbb{I}_{n}) ) $
    \STATE $K^{(i+1)}\leftarrow$ diag($D^r$) $K^{(i)}$ diag($D^c$)
    \STATE $C^{(i+1)} = -\ln (K^{(i+1)})$
    \STATE $a$ $\leftarrow P(C^{(i+1)}| T) / P(C^{(i)}|T)$
    \IF{ a $>$ u}
        \STATE Accept $C^{(i+1)}$
    \ELSE
        \STATE Reject $C^{(i+1)}$
        \STATE $C^{(i+1)} \leftarrow C^{(i)}$
    \ENDIF
\ENDFOR
\OUTPUT{: $\mathbf{C}$}
\end{algorithmic}
\end{algorithm}

\textbf{Metropolis-Hastings Monte Carlo method.}
We can avoid dealing with the parameter $\lambda$ by putting a prior on $K$. However, 
the MetroMC has long autocorrelation time for such priors. 
%does not work well for such priors. The inverse procedure involves scaling 
%the rows and columns of the observed coupling with exponential factors, and experiments show that symmetric proposal distribution used in MetroMC is not efficient (has very long autocorrelation time) 
%for $\mathbf{P}_2$. \pat{we should explain why this is so. it will not at all be obvious to the reader.} 
%\wt{Explain:} 
%The reason is that i
In each iteration, $K^{(i+1)}$ is obtained by \textit{multiplying} random factors on $K^{(i)}$, 
whereas $C^{(i+1)}$ is obtained by \textit{adding} random factors on $C^{(i)}$.
In more detail, rows of $K^{(i)}$ are scaled by exponential of Gaussian variables.
A lot of time only a few of rows are at the same scale. After the columns are normalized, 
the rows that have large scale will have most of the weight in a column, which means we will sample a lot on the boundaries of the support.
In order to improve the efficiency, we utilize the Metropolis-Hastings \citep{hastings1970monte} sampling method in below.

We propose the following algorithm to sample from $P({K|T})$ with a prior on $K$.
The sufficient condition for the ergodicity of the MCMC method is detailed balance, which is defined as
% \begin{equation}
% \pi(s)P(s'|s)=\pi(s')P(s|s'),
% \label{eq:detailed_balance}
% \end{equation}
% where $\pi$ is the equilibrium distribution and $P(s'|s)$ is the transition probability to any state s' given a state s. At the end of a MC simulation, we will have the target distribution $\pi(s)$. Eq.~\eqref{eq:detailed_balance} can be re-written as 
$\pi(K)Q(K'|K)A(K', K)=\pi(K')Q(K|K')A(K, K')$,
% \begin{equation}
% \pi(s)Q(s'|s)A(s', s)=\pi(s')Q(s|s')A(s, s'),
% \label{eq:detailed_balance_2}
% \end{equation}
where $\pi(K)$ is the stationary distribution, $Q(K'|K)$ is the proposal transition probability from a state $K$ to $K'$ and $A(K',K)$ is the acceptance probability which we choose
\begin{equation}
A(K', K)=\text{min} \left(1, \frac{P(K')}{P(K)}\frac{Q(K|K'))}{Q(K'|K)} \right).
\label{eq:acceptance_prob}
\end{equation}
We replace the acceptance probability $a$ in Alg.~\ref{algo:MetroMC} with Eq.~(\ref{eq:acceptance_prob}). We refer this algorithm as the MHMC.

The proposal transition probability $Q(K'|K)$ is defined as follows.
In each iteration of MHMC, we draw a random number $\epsilon_i$ from a Gaussian distribution $\mathcal{N}(0, \sigma^2)$, where $\sigma$ depends on the row sums of $K$.
More precisely, $\sigma=\sigma_0*s_i^{\gamma}+\delta$, where $\sigma_0$, $\gamma$, and $\delta$ are hyper-parameters for the model and $s_i = \sum_j K_{ij}$. 
$K' = \text{diag}(D_i^r) K$, where $D_i^r = e^{[0,..,0,\epsilon_i,0,...,0]}$.

We %use the Gibbs sampling method to 
scale a row at once, so in each step the acceptance ratio is reduced to $A(s_i'|s_i)$. To compute it we need 
$Q(s_i'|s_i)=F_N(x=\ln(s_i'/s_i), \mu, \sigma) $ and 
$Q(s_i|s_i')=F_N(x=\ln(s_i/s_i'), \mu', \sigma')$,
% \begin{align*}
% Q(s'|s)=&F_N(x=\ln\frac{s'}{s}, \mu=0, \sigma) \\
% Q(s|s')=&F_N(x=\ln\frac{s}{s'}, \mu'=0, \sigma'),
% \end{align*}
where $F_N(x, \mu, \sigma)$ is the Gaussian probability density function.% $\mathcal{N}(\mu, \sigma^2)$.
%Gaussian probability density function with mean = $\mu$ and standard deviation = $\sigma$ at $x$. Namely, 
% \begin{equation}
% F_N(x, \mu, \sigma) = \frac{1}{\sqrt{2\pi}}e^{-\frac{1}{2}\frac{(x-\mu)^2}{\sigma^2}}
% \label{eq:Gaussian_pdf}
% \end{equation}

% In our simulations, we sample the next state $s'$ based on the current state $s$. We take s to be the sum of the row/column being scaled, $\mu=\mu'=0$, the standard deviation $\sigma=\sigma_0*s^k+\delta$ where $\sigma_0$, $k$, and $\delta$ are hyper-parameters.
%

% Move to supplements
% \subsection{Auto-correlation function.}

% To monitor the efficiency of our MCMC method, we compute the auto-correlation function during the burn-in phase and choose the lags accordingly.

% The auto-correlation function is defined as
% \begin{equation}
% R(t)=\frac{1}{(N-t)\sigma^2}\sum_{l=1}^{N-t}\sum_{i,j}(K_{i,j}^{(l)}-\bar{K}_{i,j})\cdot(K^{(l+t)}_{i,j}-\bar{K}_{i,j}),
% \label{eq:autocorrelation}
% \end{equation}
% where $N$ is the total number of samples generated, (i,j) runs over all indices of the matrices $K$, $\bar{K}$ is the mean averaging over all the samples, and $\sigma_K^2=\sum_{i,j}\sigma_{K,i,j}^2$ is the variance. 

\section{Experiments}\label{sec:simulations}

The experiments are performed on computing clusters with nodes equipped with 40 Intel Xeon 2.10 GHz CPUs and 192 GB memory. Note that the PIOT inference has a linear computational complexity of $\mathcal{O}(mn)$. (See appendix for further details.) 

%\subsection{Effects of matrix size on manifold behavior.}

\begin{figure}[h!]
    \centering
    \begin{tabular}{c c c c}
    \hspace{-7pt}
        \begin{picture}(70,70)% width and height of the picture
            \put(0,0){\includegraphics[width=0.16\textwidth]{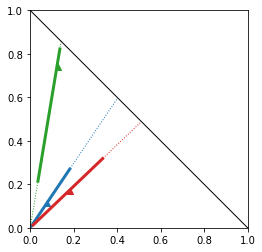}}
            \put(60,60){\textbf{a.}}
        \end{picture}
        %\textbf{a.}\includegraphics[width=0.2\textwidth]{figs/submat_1.png}
        &\hspace{-7pt}
        \begin{picture}(70,70)% width and height of the picture
            \put(0,0){\includegraphics[width=0.16\textwidth]{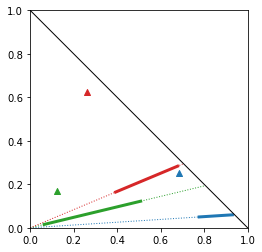}}
            \put(60,60){\textbf{b.}}
        \end{picture}
    %\end{tabular}
    %\begin{tabular}{c c}
        &\hspace{-7pt}
        \begin{picture}(70,70)% width and height of the picture
            \put(0,0){\includegraphics[width=0.16\textwidth]{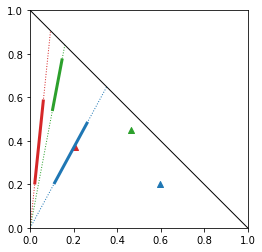}}
            \put(60,60){\textbf{c.}}
        \end{picture}
    %     &\hspace{-7pt}
    %     \begin{picture}(50,50)% width and height of the picture
    %         \put(0,0){\includegraphics[width=0.12\textwidth]{figs/submat_4.png}}
    %         \put(40,40){\textbf{d.}}
    %     \end{picture}
    \end{tabular}
    \caption{
    Visualization of $2\times 2$ subspaces of $\text{supp}[P(K|T)]$. 
    In each plot, a $T$ is sampled from a Dirichlet distribution. 
    Each colored mark represents a uniformly sampled $\mathbf{k}_1 = (k_{11}, k_{21})^T$.
    Solid segment with the same color plots the corresponding set $\mathcal{K}_2$ for the second column, and
    the dashed ray plots $\mathcal{W}_2$. 
     }%Randomly generates the elements of first column of $K$ with $\sum_{i} K_{i,j} = 1$. Three sets of points are sampled for each plot and represented in different colors. Triangular points are ($K_{1,1}$, $K_{2,1}$) of first column; Lines are sets of points ($K_{2,1}$, $K_{2,2}$) of column 2.}
    \label{fig:2x2_subspace}
    \label{fig:subspace}
\end{figure}
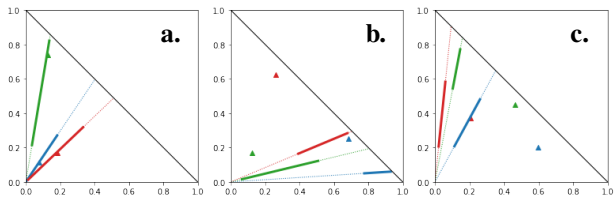

\textbf{Visualizing subspaces of $\text{supp}[P(K|T)]$.} Fig~\ref{fig:2x2_subspace}(a-c) illustrates 
support of $2\times 2$ subspaces $\text{supp}[P(K^s|T)]$, for a $4\times 4$ coupling $T$ under prior $\mathbf{P}_2$.
In each plot, a $T$ is sampled from a Dirichlet distribution \footnote{All coupling $T$s in this section are reported in Appendix~\ref{apd:sec:matrices}.}.
Consider submatrix %$\tiny K^s=\begin{pmatrix} k_{11}& k_{12} \\ k_{21} & k_{22}\end{pmatrix}$ 
$K_s = [[k_{11}, k_{12}],[k_{21}, k_{22}]]$ of $K$. 
According to section~\ref{para:subspace}, both columns $\mathbf{k}_1, \mathbf{k}_2$ of $K_s$ lie in a copy of $\Sigma_2$. 
Each colored triangle mark represents a uniformly sampled $\mathbf{k}_1$ from $\Sigma_2$. 
A sampled $\mathbf{k}_1$ determines a $D^r_s = \diag\{k_{11}/t_{11}, k_{21}/t_{21}\}$, 
which further determines a set $\mathcal{K}_2 $ for $\mathbf{k}_2$  shown by solid segment with the same color.
% the corresponding $\mathcal{K}^s_2$, the set of $\mathbf{k}^s_2$, is shown by solid segment with the same color,
Projection of $\mathcal{W}$ into the simplex corresponding to $\mathbf{k}_2$, denoted by $\mathcal{W}_2$, is shown by the dashed ray.
Here, $\mathcal{K}_2$ is obtained by: fix $\mathbf{k}_1 = (k_{11}, k_{21})^T$, 
further sample $k_{31}$ and $k_{41}$ to form the first column of $K$.
Then compute and plot the uniquely determined $\mathbf{k}_2$. 
As predicted in section~\ref{para:subspace}, $\mathcal{K}_2$ is only a subset of $\mathcal{W}_2$. 
The sizes and locations of $\mathcal{K}_2$ in $\mathcal{W}_2$ vary according to cross ratios of $T$.

\begin{figure}[h!]
    \centering
    \begin{tabular}{c | c c c}
        \includegraphics[width=0.07\textwidth]{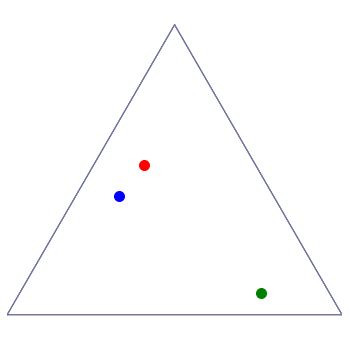} &
        \includegraphics[width=0.07\textwidth]{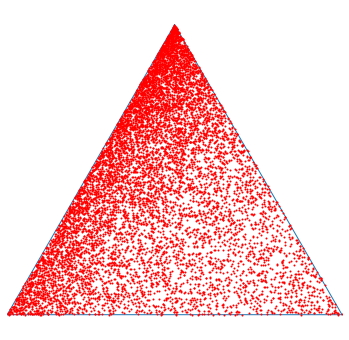} &
        \includegraphics[width=0.07\textwidth]{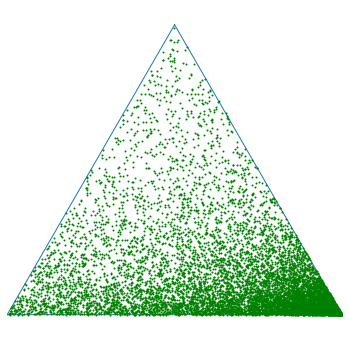} &
        \includegraphics[width=0.07\textwidth]{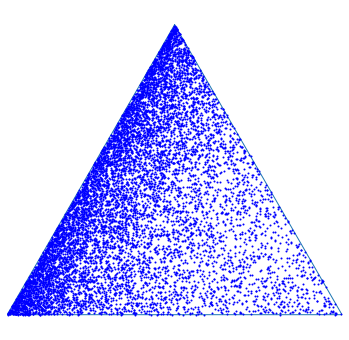}
    \end{tabular}
    \begin{tabular}{c | c c c}
        \includegraphics[width=0.07\textwidth]{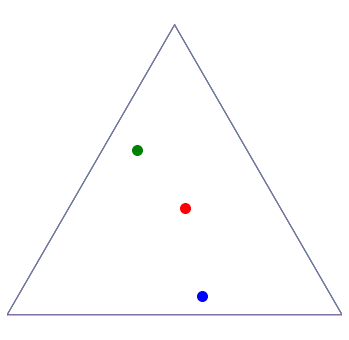} &
        \includegraphics[width=0.07\textwidth]{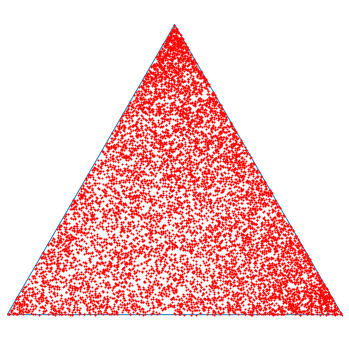} &
        \includegraphics[width=0.07\textwidth]{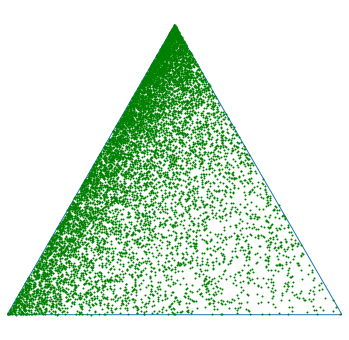} &
        \includegraphics[width=0.07\textwidth]{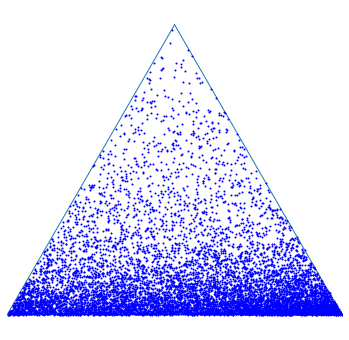} 
    \end{tabular}
    \begin{tabular}{c | c c c}
        \hspace{2pt}\includegraphics[width=0.07\textwidth]{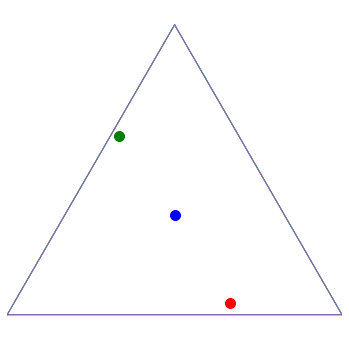} &
        \includegraphics[width=0.07\textwidth]{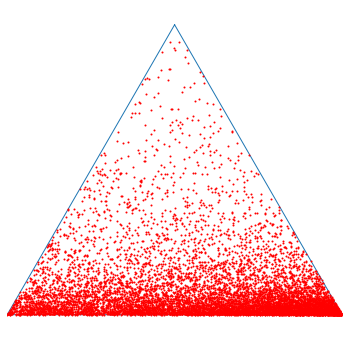} &
        \includegraphics[width=0.07\textwidth]{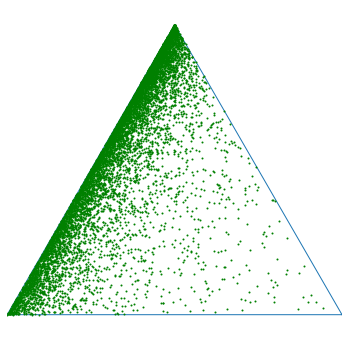} &
        \includegraphics[width=0.07\textwidth]{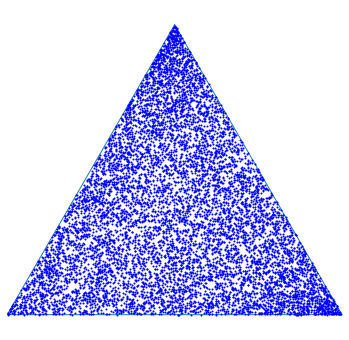} 
    \end{tabular}
    \vspace{3pt}
    \caption{\textbf{(Left.)} The columns of three $T$ matrices plotted on the same simplex: $\mathbf{t}_1,\mathbf{t}_2, \mathbf{t}_3 $ are shown in
    \textcolor{red}{Red}, \textcolor{green}{green} for , \textcolor{blue}{blue}. 
\textbf{(Right.)} Samples from posterior distribution $P(K|T)$. Each column of inferred matrices $K$ are plotted in a separate simplex, indicated by corresponding color.}
    \label{fig:sample_supp}
\end{figure}

\textbf{Visualizing $\text{supp}[P(K|T)]$ through MHMC.} Now we utilize MHMC to visualize supp[$P(K|T)$] and see the effect of cross-ratios. 
Under prior $\mathbf{P}_2$ with $\alpha =1$, Fig.~\ref{fig:sample_supp} illustrates the posterior distributions $P(K|T)$ of three $3\times 3$ couplings. 
% Here couplings $T$ are generated by solving an OT problem with uniform marginals $\mu, \nu$, and random column Dirichlet matrices $K^g$. 
% \pw{what is $K^g$}.

%We simulate $P(K|T)$ using MHMC with the same parameters as in the previous uniform matrix case 
%\pw{fix}. \wt{fixed}
We choose $\alpha=1$ for the Dirichlet prior $P_0(K)$, $\sigma_0=0.5$, $\gamma=3$, and $\delta=1.0$. We run for 10,000 burn-in steps and take 10,000 samples with lags of 100. 
The choices of these numbers are validated in Appendix~\ref{apd:sec:MHMC}. 
Three columns (\textcolor{red}{red}, \textcolor{green}{green}, and \textcolor{blue}{blue} colors, respectively) 
of each $T$ are plotted in the simplex on the left of Fig.~\ref{fig:sample_supp}. 
Locations of these points illustrate the relations between each column encoded by the cross ratios of $T$. 
The right of Fig.~\ref{fig:sample_supp} plots columns of sampled $K$ from $P(K|T)$ with corresponding $T$ to the left. 
We clearly see how the cross-ratio of $T$ affects $P(K|T)$. 
If one column of $T$ is close to one edge, then the posterior distribution of that column will be denser on that edge; on the other hand, if a column of $T$ is located near the center, then the posterior is more evenly distributed on a simplex. 
%Since the relative location of each column should be kept to retain the cross-ratio equivalence for all inferred samples.
E.g. in the top case, the effects on red and blue are exaggerated because the effect depends on relative position (and similarly for the red in the middle panel). 

Visualization of $\text{supp}[P(K|T)]$ with 
incomplete observations is included in Appendix~\ref{apd:sec:missing_value}.

\textbf{Symmetric cost.} In this section, we consider $n\times n$ squared symmetric cost with $C^{g}_{ij}=\left|\frac{(i-j)}{n}\right|^{p}$ for $i,j\in [n]$, $n =$ 10, and $p=$0.5, 1, and 2. 
$T^{g}$ is obtained by EOT with $\lambda=10$ and randomly sampled $\mu$ and $\nu$. 
The relative error $\parallel \hat{C}-C^{g}\parallel/\parallel C^{g}\parallel$ is used to evaluate the quality of estimation $\hat{C}$, where $\parallel \cdot \parallel$ is the $l^2$ norm. MetroMC is used.

%monitor the convergence, where $\hat{C}$ is the median of the inferred cost distribution and $\parallel \cdot \parallel$ is the Frobenius norm.  

%The result of PIOT is comparable to the discrete algorithm proposed in \citep{ma2020learning}. 
We compare PIOT to the algorithm in \citet{ma2020learning} and find comparable performance despite our more general approach. %for symmetric costs with a much softer constraint.
\citet{ma2020learning} projecting the inferred cost to the set of symmetric matrices at every iteration, effectively assuming the ground truth is symmetric. 
Their algorithm reaches an error below $10^{-5}$ in 300 iterations for all three $p$ values. 
The flexibility of PIOT allows us to pick any prior. Here we choose a prior such that $P_0(C)$ follows a Gibbs distribution $\text{exp}(-\beta(\parallel \gamma(C-C^T) \parallel))$.
The results show that after burn-in, the inferred cost distributed is already concentrated around the ground truth cost, therefore, 
with 1000 MC samples the PIOT method achieve an error less than $10^{-5}$ for all the three $p$ cases.
($\hat{C}$ is the median of inferred cost distribution, see simulation details in Appendix~\ref{apd:ma}.) 

\begin{figure}[h!]
    \centering
    \begin{tabular}{c c c}
        \includegraphics[width=0.155\textwidth]{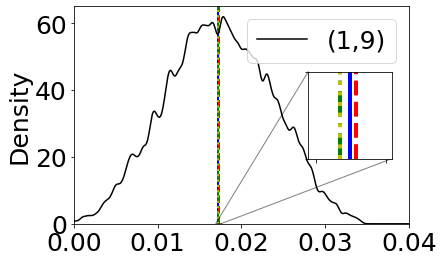}
        \hspace{-3pt}\includegraphics[width=0.155\textwidth]{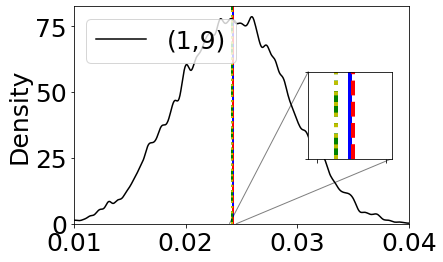}
        \hspace{-3pt}\includegraphics[width=0.155\textwidth]{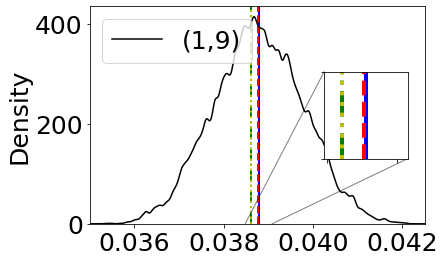}
    \end{tabular}
    \caption{The KDE of the inferred distribution of  $C_{1,9}$. The \textcolor{blue}{blue} solid lines indicate the ground truth cost, the \textcolor{green}{green} dash-dotted lines are results using algorithm from \citep{ma2020learning} at 300-iteration, the \textcolor{yellow}{yellow} dotted lines are Ma's results at 1000-iteration (Overlapped with green lines for all cases.), and the \textcolor{red}{red} dashed lines are the median of PIOT. (Left.) $p=0.5$. (Middle.) $p=1$. (Right.) $p=2$. The insets zoom-in the range of $C^g_{1,9}\pm 10^{-4}$.}
    \label{fig:main_non-sym_cost}
\end{figure}

Next, we test the robustness of both algorithms by adding random perturbations onto the symmetric costs. 
Random matrices with all entries in $[0, 10^{-4})$ are sampled, and added to the symmetric costs.
$C^{g}$ is then obtained by normalizing the combined matrices to 1. 
For PIOT with the uniform prior $\mathbf{P_1}$ described in section 3,
the relative errors fall between $10^{-3}$ to $10^{-4}$ for all $p$ case,
whereas the errors are in the range of $10^{-2}$ to $10^{-3}$ using the algorithm from \citep{ma2020learning}.
Figure~\ref{fig:main_non-sym_cost} shows the results on $C_{1,9}$, other elements behave similarly. (See simulation details in Appendix~\ref{apd:ma}.)
%The PIOT method is approximately an order more accurate. 
%\pw{Probably, this can be combined with previous subsection. 

Thus when observed $T$ did come from a cost aligned with the symmetry assumption, both PIOT and \citep{ma2020learning} achieve comparable results.
However, even with small perturbations, PIOT attains better accuracy. 
%\pw{did not find plot for ma's algo in the appendix?, we may move fig 11 in the appendix to the main text, with ma's algo labeled by an other vertical line.}

\textbf{Bounded noise on observed plan.}
Given an observed noisy $T$, 
we assume $t_{11}$ was perturbed by a bounded noise $\epsilon$ sampled uniformly 
from distribution $[-a, a]$ with $a=0.01$. 
Assuming prior $\mathbf{P}_1$ for $P_0(C)$ with symmetric Dirichlet parameter equals to $1$, 
we use MetroMC to sample $\text{supp}[P(C|T)]$.

% To see the effects of noise on the inferred cost we take five evenly spaced values within the range. A ground truth probability cost matrix $C^g$, whose elements are randomly rolled from a uniform distribution [0, 1] and normalized to $\sum_{i,j}C^g_{i,j}=1$, is generated and a OT problem is solved to obtain a plan $T^g$ with even marginals $\mu, \nu$ and $C^g$. For simplicity we do not add a random noise on $T^g$ to obtain the observed plan $T$, so $T=T^g$. 
%For the PIOT inference, the noise is added to the (1,1)-th element of the  observed plan $T$. 

% In order to make $-log(T)$ lie on the support of the posterior we re-normalize the plans upon performing the MCMC method by a factor $F(T)$.
% \pw{how about swap to }

Notice that initializing $K^{(0)} = T$ may cause the corresponding $C^{(0)}$ lies outside the domain of $P_0(C)$ as 
$\sum_{ij} c^{(0)}_{ij} = \sum_{ij} -\ln(t_{ij})$ is not guaranteed be $1$. This will cause an extremely 
long burning time. In order fix to this issue, we scale $K^{(0)}$ by a constant when necessary.
Scaling does not change the cross-ratios, hence it equivalent to a move within the support of $\Phi^{-1}(T)$. 

%We assume a symmetric Dirichlet prior on $P_0(C)$, where $\sum_{i,j}C_{i,j}=1$ and the Dirichlet parameters = 1 everywhere. 
For each iteration in the MCMC method we generate $m+n-1$ (rather than $m+n$) independent Gaussian random numbers 
with $\sigma = 0.02$ to construct vectors $D^r$ and $D^c$ 
satisfying $\sum_{i,j}\ln(d^r_{i})+\ln(d^c_j)=0$.
%eq.~\ref{eq:condition_bounded_noises}.
The purpose of the above condition is to ensure that the newly generated sample lies on the support of the posterior. Otherwise, the acceptance rate will be extremely low. Because $C$ is a positive matrix, the MetroMC method rejects any sample of $K$ if $\max_{(i,j)}k_{i,j}\ge 1$. 

\begin{figure}[h!]
    \centering
    \begin{tabular}{c c}
    \hspace{-8pt}\includegraphics[width=0.25\textwidth]{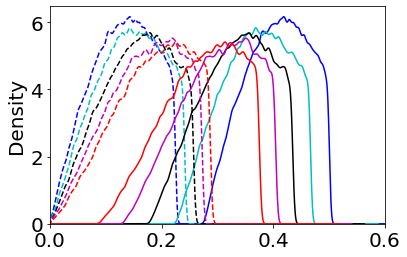}
    \put(-70,-4){$c_{i}+c_{j}$}\put(45,-4){$c_{i}+c_{j}$}
    &
    \hspace{-18pt}\includegraphics[width=0.25\textwidth]{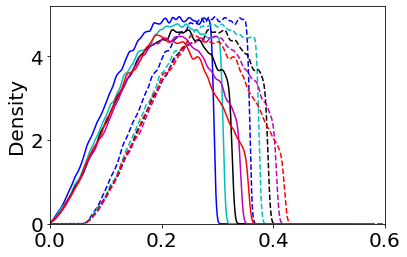}  
    \end{tabular}
    \vspace{5pt}
    \caption{Results on infered cost $C$ given a $T$ with bounded noise $\epsilon$ on $t_{11}$.
    $\epsilon$ varies in $[-0.01, -0.005, 0.0, 0.005, 0.01]$, indicated by [\textcolor{blue}{blue}, \textcolor{cyan}{cyan}, \textcolor{black}{black},  
    \textcolor{magenta}{magenta}, \textcolor{red}{red}] in order.
    %The KDE of the distributions of cross-ratio pairs of pair of elements of $C$ with various noise added to $t_{11}$. 
    %The colors represent the noise, \textcolor{blue}{blue}: -0.01, \textcolor{cyan}{cyan}: -0.005, \textcolor{black}{black}: 0.0,  \textcolor{magenta}{magenta}: 0.005, and \textcolor{red}{red}: 0.01.
    \textbf{(Left)} Solid curves are for ($c_{11}+c_{22}$), and dashed for $(c_{21}+c_{12})$. 
    \textbf{(Right)} The solid curves are for ($c_{12}+c_{23}$), and dashed for $(c_{13}+c_{22})$.
    }
    \label{fig:bounded noise}
\end{figure}

Fig.~\ref{fig:bounded noise} shows the MetroMC results on the inferred cost matrices $C$ for a given $3\times 3$ coupling 
$T$. For each simulation we burn in 10,000 steps and take 100,000 samples with lags of 100. 
The acceptance rates range from 0.51 to 0.58. The MCMC method is validated by checking the autocorrelation function and the running averages (see Appendix~\ref{supsec: alg}.)
Noise level $\epsilon$ is indicated by colors.
The \textbf{Left} panel plots components for a cross-ratio depend on $t_{11}$: $(c_{11}+c_{22})$ by solid curves and $(c_{21}+c_{12})$ by dashed curves.
The \textbf{Right} panel plots components for a cross-ratio dose not depend on $t_{11}$: 
$(c_{12}+c_{23})$ by solid curves and $(c_{13}+c_{22})$ by dashed curves.
All the curves are smoothed using the Gaussian kernel density estimation (KDE) with a bandwidth of 0.05. 
Fig.~\ref{fig:bounded noise} illustrates that curves for $(c_{11} + c_{22})$ and $ (c_{21} + c_{12})$ with the same $\epsilon$ (i.e. curves in the same color) are in the same shape.
The off-set between curves varies as the noisy level changes. 
Whereas off-set between curves for $(c_{12}+c_{23})$ and $(c_{13}+c_{22})$ on the right panel remains a constant for any choice of $\epsilon$.
This is consistent with our theoretical result in Prop.~\ref{prop: bounded_noise} that:
$(c_{11} + c_{22}) - (c_{21} + c_{12}) = - \ln\frac{t'_{11} t_{22}}{t_{21}t_{12}}$ depends on $\epsilon$,
whereas $(c_{12}+c_{23}) - (c_{13}+c_{22}) = - \ln\frac{t_{12} t_{23}}{t_{13}t_{22}}$ is a constant.
%\wt{Dies $t_{12}$ has a prime?}
%The cross-ratio pair including the (1,1)-th element.
% Hence curves for $(c_{11} + c_{22})$ and $ (c_{21} + c_{12})$ with a fixed $\epsilon$, i.e. curves in the same color, should be in the same shape.
% And their off-set varies as the noisy level changes. 
% Whereas curves for $c_{12}+c_{23}$ and $(c_{13}+c_{22})$) should be in the same shape with a fixed off-set.

% so only the difference between cross-ratio pairs of pairs of elements of $C$ which involve the element with noise changes as the value of the noise alters. E.g. the difference between the cross-ratio pair ($c_{11} + c_{22}$) and ($c_{21} + c_{12}$) is controlled by the noise added to $t'_{11}$ while the difference between another pair ($c_{12} + c_{23}$) and ($c_{13} + c_{22}$) is not, see 

% C = [[0.1626, 0.0279, 0.1071],
%       [0.1047, 0.1477, 0.1628],
%       [0.1358, 0.0265, 0.1249]]
% T = [[0.1067, 0.1141, 0.1125],[0.1175, 0.1052, 0.1106],[0.1092, 0.1139, 0.1102]]
% noises = [-0.01, -0.005, 0.0, 0.005, 0.01]

%\subsection{Gaussian noise on observed plan.}
\textbf{Gaussian noise on observed plan.}\label{sec:gaussian_noise} We start with a Toeplitz ground truth cost function $C^g$ and obtain an OT plan, $T^g$, with $C^g$ and uniform $\mu, \nu$. A random Gaussian noise $\epsilon \sim \mathcal{N}(0, 0.004^2)$, about 4\% of the average value of $T$, is added to $t^g_{1,2}$, and the corrupted plan $T=T^g+\epsilon$. We then infer the posterior distribution $P(C|T)$. We put a $\mathbf{P}_1$ prior on $C$ with the hyperparameters follows Toeplitz matrix format. 
%The prior of $P_0(C)$ is assumed to be toeplitz Dirichlet, where the matrix of the Dirichlet parameters is diagonal-constant and $\sum_{i,j}{C_{i,j}}=1$. 
Inference is done by first generating 10 random numbers $\epsilon_i^{*}\sim \mathcal{N}(0, 0.004^2)$ and apply the noise on $t_{12}$ to get $T_i^{*}=T+\epsilon_i^{*}$. Then, we utilize MetroMC on every $T_i^{*}$ to obtain $P(C|T_i^{*}$), and finally $P(C|T)\approx\sum_i P(C|T_i^{*})$. %\wt{Confirm notation with paragraph after remark 3.16.}

For the MetroMC method, we burn in 10,000 steps and then take 10,000 samples with lags of 200 steps for each $P(C|T_i^{*})$. For each step, we use the same sampling procedure as described in the bounded noise case but use a Gaussian standard deviation of 0.003. The acceptance rates are around 0.52-0.70.

\begin{figure}[h!]
    \centering
    \begin{tabular}{c c c}
        \hspace{-8pt}\includegraphics[width=0.15\textwidth]{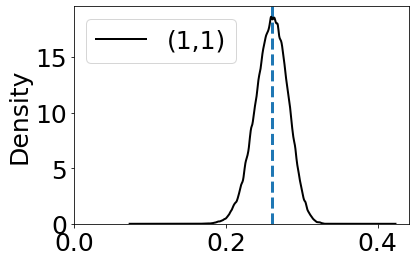} &
        \hspace{-8pt}\includegraphics[width=0.15\textwidth]{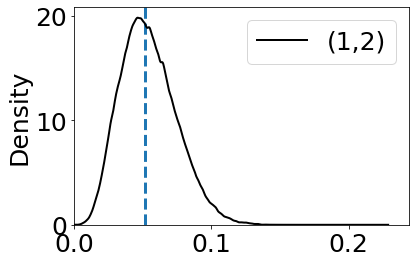} &
        \hspace{-8pt}\includegraphics[width=0.15\textwidth]{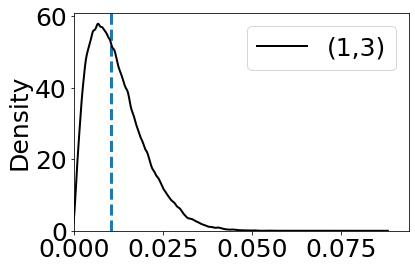} 
    \end{tabular}
    \begin{tabular}{c c c}
        \hspace{-8pt}\includegraphics[width=0.15\textwidth]{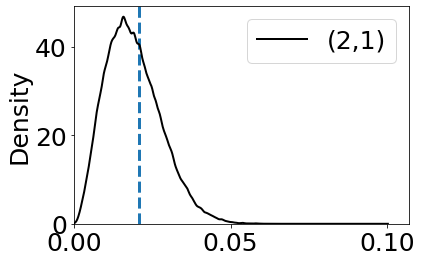} &
        \hspace{-8pt}\includegraphics[width=0.15\textwidth]{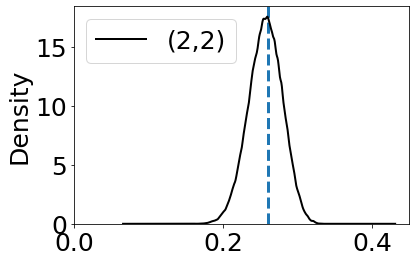} &
        \hspace{-8pt}\includegraphics[width=0.15\textwidth]{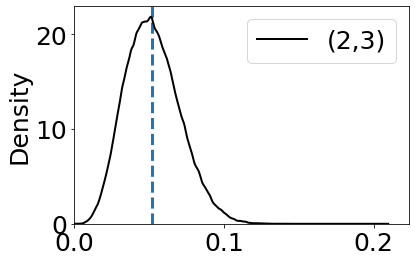} 
    \end{tabular}
    \begin{tabular}{c c c}
        \hspace{-8pt}\includegraphics[width=0.15\textwidth]{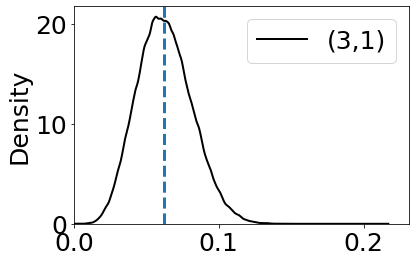} &
        \hspace{-8pt}\includegraphics[width=0.15\textwidth]{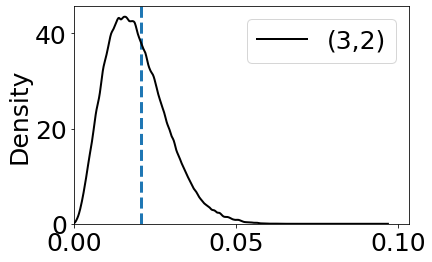} &
        \hspace{-8pt}\includegraphics[width=0.15\textwidth]{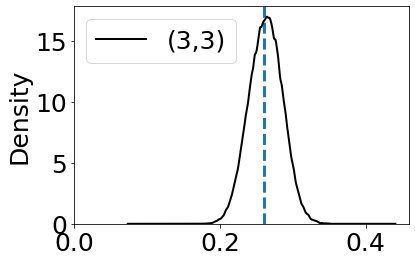} 
    \end{tabular}
    \caption{The KDE of each element of inferred cost $C$ with Gaussian noise on $t_{12}$. The blue dashed vertical lines indicate the ground truth cost $C^g$.
    }
    \label{fig:Gaussian noise}
\end{figure}

The posterior distributions of each element of the inferred $C$ are shown in Fig.~\ref{fig:Gaussian noise}. The inferred distributions are close to the ground truth costs which are indicated by blue dashed lines. This experiment demonstrates that our method successfully make accurate inferences using soft constraints $P_0(C)$, with the results comparable to hard constraints imposed by \citep{stuart2020inverse}. 

% C^g  =[[0.2604, 0.0521, 0.0104],
        % [0.0208, 0.2604, 0.0521],
        % [0.0625, 0.0208, 0.2604]]

% T^g = [[0.0952, 0.1168, 0.1214],
        % [0.1214, 0.0952, 0.1168],
        % [0.1168, 0.1214, 0.0952]]
        
% corruption noise = -0.0011

% noises = [ 0.0004, -0.0008,  0.0029,  0.0002, -0.0056, -0.0005,  0.0006, -0.0023,0.0041,  0.0076]

% d, e, f, g, h = 25., 5., 1.9, 3., 6.
% alpha = torch.tensor([[d, e, f],\
%                       [g, d, e],\
%                       [h, g, d]])

\begin{figure}[h!]
    \centering
    \hspace{-5pt}\includegraphics[width=0.3\textwidth]{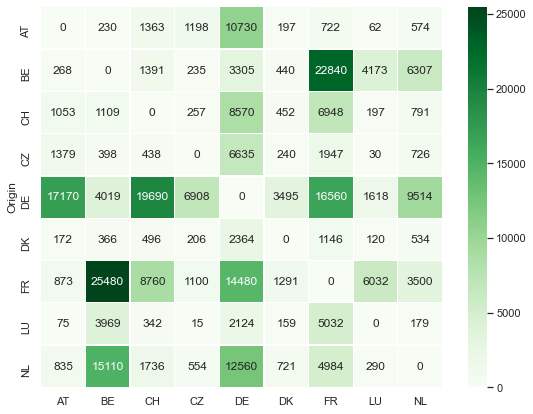}
    \includegraphics[width=0.295\textwidth]{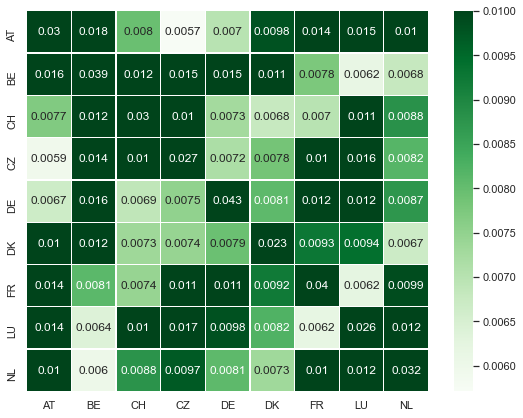}
    \begin{tabular}{cc}
        \hspace{-8pt}\includegraphics[width=0.22\textwidth]{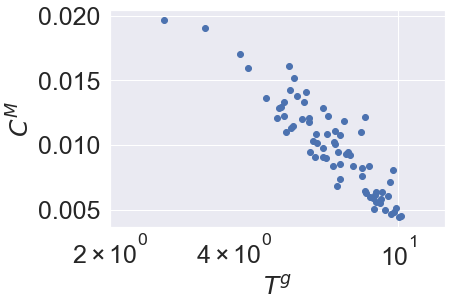}
        \includegraphics[width=0.22\textwidth]{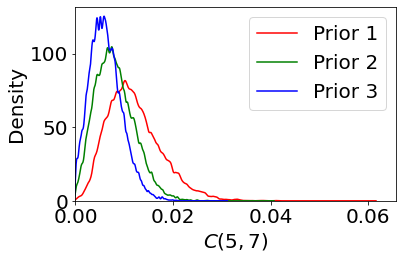}
    \end{tabular}
    \caption{(\textbf{Top.}) European migration data. 
    (\textbf{Middle.}) Mean of the inferred cost using Prior 2.
    (\textbf{Bottom-left.}) Plots for logarithm of each migration data with its corresponding inferred cost mean.
     (\textbf{Bottom-right.}) The inferred cost distributions for DE$\rightarrow$FR under three priors.}
    \label{fig:heatmap_migration_and_cost}
\end{figure}

\textbf{Application on European Migration data.}
We apply PIOT to analyze European migration flow. %as proposed in \citep{stuart2020inverse}.
The observed migration data $T^g$ between 9 European countries for the period 2002- 2007 is shown in Fig.~\ref{fig:heatmap_migration_and_cost} top \citep{raymer2013integrated}. 
% In this section we apply the PIOT method to investigate the cost function of the 9 selected countries from the European migration data.\citep{stuart2020inverse, raymer2013integrated} The observed migration coupling ($T^g$) is shown in the left of Fig.~\ref{fig:heatmap_migration_and_cost}. 
%We first study the effect of prior on the cost function. 
All results in this section are obtained by performing MetroMC with 10,000 burn-in steps, 10,000 samples with lags of 1000,
and the sum of cost matrix is assumed to be a constant, see details in Appendix~\ref{apd:sec:migration}.

First, we demonstrate the ability of PIOT by inferring the latent costs under various priors.
Three priors are considered with increasing strength:
1. Fully uniform prior, Dirichlet with all $\alpha=1$; 
2. Semi-uniform prior, diagonal $\alpha$=25 and 1 elsewhere, as the data does not consider migration within a country itself.
3. Graph-geometrical prior, the diagonals are 25  and $\alpha$=1 if the two countries are directly connected and 1.5 if not, 
as comparison to the hard Graph-geometrical constraint imposed by \citep{stuart2020inverse}.
% (1) Uniform prior: Dirichlet with all $\alpha=1$; (2) Semi-uniform prior, diagonal $\alpha$=25 and 1 elsewhere;
% (3) Graph-geometrical prior, the diagonals are 25 and $\alpha$=1 if the two countries are directly connected and 1.5 if not. 
%The priors are stronger in increasing order. 

%The inverted color between two plots shows that PIOT captures the reciprocal relationship: lower costs 
%lead to higher migration numbers. 
Fig.~\ref{fig:heatmap_migration_and_cost} middle plots the mean of the inferred costs under prior~2. 
Even with a weak prior, PIOT correctly captures the general reciprocal relation between coupling and cost,
lower costs lead to higher migration numbers, shown in Fig.~\ref{fig:heatmap_migration_and_cost} bottom-left. 
Results for the other two priors are similar, see Appendix \ref{supsec: alg}. 
%Exceptions happen due to lots of reasons such as population of the country, immigration policy, and ticket prices, but these practical effects on the cost are out of our scope. 
%Before each simulation we put ones to the diagonal of $T^g$ so that $T^g \in \mathbf{R}_{+}$.  
%The results of the (5,7) matrix element (DE$\rightarrow$FR)
Fig.~\ref{fig:heatmap_migration_and_cost}.bottom-right plots the inferred costs from Germany to France (DE$\rightarrow$FR).
The implications of priors can be observed: the stronger the prior is, the narrower the inferred cost distribution is.
In particular, using prior 3, the mass is concentrated between 0 and 0.01 the same as in \citep{stuart2020inverse}, where hard graphical constraint on the cost is considered. 
%the distributions become more structural as 
% The weakest prior 1 has the widest distribution which is expected. 
% Prior 2 considers a soft constraint on the diagonal following the fact that no migration within each country, and thus the cost should be higher on the diagonal. Such constraint diminishes most of the cost of the other elements. Prior 3 is stronger where the geometrical relations are taken into account, therefore, the cost function becomes more structural. In fact, our result using prior 3 is comparable to the result of \citep{stuart2020inverse}, whereas they considered a much stronger constraint. 
% In Fig.~\ref{fig:heatmap_migration_and_cost} we plot the heat map of the ground truth coupling and the mean of the inferred cost function using prior 2. Our model correctly captures the general reciprocal relation between coupling and cost. Exceptions happen due to lots of reasons such as population of the country, immigration policy, and ticket prices, but these practical effects on the cost are out of our scope. 
% \begin{figure}[h!]
%     \centering
%     \includegraphics[width=0.3\textwidth]{figs/cost_DE_to_FR.png}

%     \caption{The inferred cost distributions for DE$\rightarrow$FR.}
%     \label{fig:cost_DE_to_FR}
% \end{figure}

%An upper limit = 0.012 for the values is set to increase resolution on low cost areas.

%Comparison between the predicted coupling, $T^p$, with the ground truth coupling, $T^g$. $\Delta=(T^p-T^g)/T^g$. \textbf{(Left)} Corresponding element is corrupted by a 4\% Gaussian noise. \textbf{(Right)} Value of the corresponding element is missing. 
\begin{table}[h!]
\caption{Comparison between the predicted coupling, $T^p$, with the ground truth coupling, $T^g$. $\Delta=(T^p-T^g)/\overline{T}^g$. 
Corresponding element is corrupted by a 4\% Gaussian noise on the left column, and missing on the right column. }
\label{table:missing_value}
\vskip 0.1in
\begin{center}
\begin{small}
\begin{sc}
\begin{tabular}{c | ccc | ccc}
\toprule
& & Noise & & \multicolumn{3}{c}{Missing value} \\
%& (4,5)  & (3,6) & (5,7) & (7,3) & (2,6) & (5,3) \\
& (3,6) & (4,5) & (5,7) & (3,6) & (4,5) & (5,7) \\
\midrule
% $T^g$ & 6635 & 452  & 16560 & 8760 & 440  & 19690 \\
% $T^p$ & 6593 & 454  & 16515 & 8993 & 3844 & 14824 \\
$T^g$ & 452 & 6635  & 16560 & 452 & 6635  & 16560 \\
$T^p$ & 454 & 6593  & 16515 & 3699 & 7462 & 12349 \\
\midrule
% $|\Delta|$ & 0.8\% & 0.4\% & 0.3\% &  &  & \\
$|\Delta|$ & .45\% & .87\% & .25\% & 26\% & 6.6\% & 34\%\\
\bottomrule
\end{tabular}
\end{sc}
\end{small}
\end{center}
\vskip -0.3in
\end{table}

% We start with a Toeplitz ground truth cost function $C^g$ and obtain an OT plan, $T^g$, with $C^g$ and uniform $\mu, \nu$. A random Gaussian noise $\epsilon \sim \mathcal{N}(0, 0.004^2)$, about 4\% of the average value of $T$, is added to $t^g_{1,2}$, and the corrupted plan $T=T^g+\epsilon$. We then infer the posterior distribution $P(C|T)$. We put a $\mathbf{P}_1$ prior on $C$ with the hyperparameters follows toeplitz matrix format. 
% %The prior of $P_0(C)$ is assumed to be toeplitz Dirichlet, where the matrix of the Dirichlet parameters is diagonal-constant and $\sum_{i,j}{C_{i,j}}=1$. 
% Inference is done by first generating 10 random numbers $\epsilon_i^{*}\sim \mathcal{N}(0, 0.004^2)$ and apply the noise on $t_{12}$ to get $T_i^{*}=T+\epsilon_i^{*}$. Then, we utilize MetroMC on every $T_i^{*}$ to obtain $P(C|T_i^{*}$), and finally $P(C|T)\approx\sum_i P(C|T_i^{*})$. 

Next, we test PIOT's performance on noisy data. 
%The noisy coupling $T$ is obtained by c
We corrupting an element of $T^g$ with a Gaussian noise to obtain $T$, 
whose standard deviation is set to be 4\% of the value. 
With prior~2, inference on cost distributions is done in the same fashion as in 
Fig.~\ref{fig:Gaussian noise}. 
We then apply EOT on the cost matrix formed by means of inferred distributions with the observed marginals, 
to predict the true migration data.
%We infer the plan by applying EOT to the estimated cost %\hl{E?}OT on 
% $C^M_{(i,j)}$ with the observed marginals to predict the migration number $T^p_{(i,j)}$. 
% %Our predictions are within 1\% to the ground truth values. 
Left of Table~\ref{table:missing_value} shows the results when noise occurs at 
$T^g_{36},T^g_{45}, T^g_{57}$. Our predictions $T^p$ are within 1\% to 
the mean of prior on the ground truth, denoted by $\overline{T}^g$ as shown in the third row. 

% Next, we study the effect of noise on $T^g$. We assume the $(i,j)$-th matrix elements shown in the left of table~\ref{table:missing_value} are corrupted by a Gaussian noise with $\sigma_{i,j}$ equals to 4\% of the value. We generate 10 random numbers $x_{i,j} \in N(0,\sigma_{i,j})$, and each $x_{(i,j)}$ is added to the corrupted element to obtain the corrupted plan. MetroMC is performed to obtain 10,000 inferred cost functions. 
% %We do 10,000 burn-ins and take 10,000 samples with lags of 1,000 for each noise. 
% We take the mean value of all the samples, $C^M_{(i,j)} = \sum_{k}C^(k)_{(i,j)}$, 
%where $k$ numerates each inferred cost function. We infer the plan by applying EOT to the estimated cost %\hl{E?}OT on 
% $C^M_{(i,j)}$ with the observed marginals to predict the migration number $T^p_{(i,j)}$. 
% %Our predictions are within 1\% to the ground truth values. 

Finally, we explore inference of missing data.
%Assuming incomplete observation on $T^g$,
We fill in the missing element with 100 samples ranging uniformly over $[100, 25000)$.
With semi-uniform prior, we estimate the cost and perform EOT as described above.
Right column of Table~\ref{table:missing_value} shows that %the results when $T^g_{36},T^g_{45}, T^g_{57}$ are missing.
PIOT successfully recovered the ground truth with estimation error less than 35\% of the prior mean.

For the European migration data, the true cost function is not simply the Euclidean distance between countries but could be a combination of complex factors such as
ticket prices, international relationships, and languages. To model all the factors is impractical. Strong
priors or constraints may simplify the problem \citep{stuart2020inverse}, but uniform priors and softer constraints provide less biased results.
Our experiments show that one may use PIOT to analyze the underlying geometry, and infer the
migration costs quantitatively. The inferred cost is then an estimate of the consequences of the hidden factors. Table~\ref{table:missing_value} 
%cases where there is
%a noise or a missing value are considered, which are common in real world data. 
%Hence we demonstrate that the practicality of PIOT.
additionally demonstrates inference with noise and missing values.

%Priors used in order: (1) Fully uniform prior. (2) Semi-uniform prior. (3) Graph-geometrical prior.

% \begin{figure}[h!]
%     \centering
%     \begin{tabular}{c c}
%     \includegraphics[width=0.225\textwidth]{figs/cost_pred_4_5.png}
%     \includegraphics[width=0.225\textwidth]{figs/heatmap_noise_mat_diff.png}
%     \end{tabular}

%     \caption{The results where the (4,5)-element of the coupling is corrupted by a 4\% noise.  \textbf{(Left)} The PDF of the predicted cost $C^p(4,5)$ (CZ$\rightarrow$DE). \textbf{(Right)} The difference from the ground truth coupling to the predicted coupling, $T^p$. 
%     }
%     \label{fig:migration_noise}
% \end{figure}

%\pw{what plot for real data?}

%%%%%%%%%%%%%%%%%%%%%%% conclusion %%%%%%%%%%%%%%%%%%%%%%%
\section{Conclusion}
%In conclusion, we presented an analysis of probabilistic Inverse Optimal Transport which infers latent costs from observed couplings. 
%In contrast with prior work on IOT, we consider the geometric foundations inherited from Sinkhorn scaling of entropy-regularized Optimal Transport. We analyze the support of the cost matrices associated with an observed coupling, the implications of prior distributions on costs, and introduce MCMC algorithms for posterior inference. Our models perform well on synthesis data and the results are comparable to previous works. \wt{Do we need to include a short discussion on the limitations of our model?}
We have generalized prior treatments of IOT by defining and studying the underlying support manifold and associated inference problems. 
We provided MCMC methods for inference over general priors on discrete costs and induced kernels. 
Simulations illustrate underlying geometric structure and demonstrate the feasibility of the inference. 
%\pat{FIXME: one last thing?}

\section*{Acknowledgements}
% All acknowledgments go at the end of the paper, including thanks to reviewers who gave useful comments, to colleagues who contributed to the ideas, and to funding agencies and corporate sponsors that provided financial support. 
%We would like to thank reviews for suggesting a previous work and useful comments.
Research was sponsored by the Defense Advanced Research Projects Agency (DARPA) and the Army Research Office (ARO) and was accomplished under Cooperative Agreement Numbers W911NF-20- 2-0001 and HR00112020039 to P.S. The views and conclusions contained in this document are those of the authors and should not be interpreted as representing the official policies, either expressed or implied, of the DARPA or ARO, or the U.S. Government. 
%The U.S. Government is authorized to reproduce and distribute reprints for Government purposes notwithstanding any copyright notation herein.
% To preserve the anonymity, please include acknowledgments \emph{only} in the camera-ready papers.

\bibliography{references}
\bibliographystyle{icml2022}

%%%%%%%%%%%%%%%%%%%%%%%%%%%%%%%%%%%%%%%%%%%%%%%%%%%%%%%%%%%%%%%%%%%%%%%%%%%%%%%
%%%%%%%%%%%%%%%%%%%%%%%%%%%%%%%%%%%%%%%%%%%%%%%%%%%%%%%%%%%%%%%%%%%%%%%%%%%%%%%
% APPENDIX
%%%%%%%%%%%%%%%%%%%%%%%%%%%%%%%%%%%%%%%%%%%%%%%%%%%%%%%%%%%%%%%%%%%%%%%%%%%%%%%
%%%%%%%%%%%%%%%%%%%%%%%%%%%%%%%%%%%%%%%%%%%%%%%%%%%%%%%%%%%%%%%%%%%%%%%%%%%%%%%
\newpage
\appendix
\onecolumn
\section{Examples, Definitions and Proofs} \label{apdx:proofs}

\begin{example}\label{eg: sk}
Let $\mu =\nu=(\frac{3}{8},\frac{5}{8})$, 
$\tiny C = \begin{pmatrix} \ln 1 & \ln 2 \\
 \ln 4 & \ln 1 \end{pmatrix}$. 
For $\lambda=1$, we may obtain $T$ by applying SK scaling on
$\tiny K^{\lambda} = e^{-C} = \begin{pmatrix} 1 & 1/2\\ 1/4 & 1 \end{pmatrix}$:
first, row normalize $K^{\lambda} $ with respect to $\mu$ results:
$\tiny K_1^{\lambda} =  \begin{pmatrix} 1/4 & 1/8\\ 1/8 & 1/2 \end{pmatrix}$. 
Then column normalization of $ K_1^{\lambda}$ with respect to $\nu$ outputs 
$\tiny K_2^{\lambda} = \begin{pmatrix} 1/4 & 1/8\\ 1/8 & 1/2 \end{pmatrix}$. 
As $K_1^{\lambda}= K_2^{\lambda}$, the SK scaling has converged with $T = K_1^{\lambda}$.
In general, multiple iterations may be required to reach the limit. %\pw{may move to apdx}
\end{example}

\begin{repproposition}{prop:piot}
When $T = T^*$, $ P(C|T)$ is supported on the intersection between $\Phi^{-1} (T)$ and the domain of $ P_0(C)$, moreover, we have that
$\displaystyle P(C|T) = \frac{P_0(C)}{\int_{\Phi^{-1} (T)} P_0(C) \text{d} C}$.
\end{repproposition}

\begin{proof}
When $T = T^*$, Eq~\eqref{eq: PIOT} and Eq~\eqref{eq: bayes} imply that:
$P(C|T) = \frac{P(T|C) P_0(C)}{P(T)}$. $P(T) >0$ is the normalizing constant, 
so $\text{Supp}[P(C|T)] = \text{Supp}[P(T|C)] \cap \text{Domain}[P_0(C)] $.
Note that $P(T|C) P_0(C) = P_0(C)$ for $C \in \Phi^{-1}(T)$, otherwise $P(T|C) P_0(C) =0$.
Hence the proposition holds.
\end{proof}

\begin{repproposition}{prop:DAD}
Let $T$ be a non-negative optimal coupling of dimension $m\times n$. $C \in \Phi^{-1} (T)$ if and only if for every $\epsilon > 0$,
there exist two positive diagonal matrices $D^r = \diag\{d^r_1, \dots, d^r_m\}$ and $D^c =\diag \{d^c_1,\dots, d^c_n\}$ such that: 
$\displaystyle |D^r K D^c - T| < \epsilon$, where $K = e^{-C}$ and $|\cdot|$ is the $L^1$ norm. 
In particular, if $T$ is a positive matrix, then $C \in \Phi^{-1} (T)$ 
if and only if there exist positive diagonal matrices $D^r,D^c$ such that $D^r K D^c = T$, i.e.
\begin{equation*}
\Phi^{-1} (T) = \{C| \exists  D^{r} \text{ and } D^{c} \text{ s.t. } K = D^rTD^c\}    
\end{equation*}
\end{repproposition}

\begin{proof}
Let the row and column of marginals of $T$ be $\mu$ and $\nu$.
Then $C \in \Phi^{-1} (T)$ if and only if $(\mu, \nu)$-Sinkhorn scaling of $K$ converges to $T$ in $L^1$ norm 
(for finite matrices, convergence in all $L^k$ norms are equivalent).
Notice that a row (column) normalization step in the Sinkhorn scaling is equivalent to a left(right) matrix multiplication of a positive diagonal matrix. 
Indeed, let $K_0 = {K}$, $\mathbf{r}_0 = K_0 \mathbf{I}_n$ (vector for row sums of $K_0$), $D_0^{r} = \diag(\mu /\mathbf{r}_0)$.
Here $\mu /\mathbf{r}_0$ represents element-wise division. 
Let $K'_0$ be the matrix obtained by row normalization of $K_0$ with respect to $\mu$. Then $K'_0 = D_0^{r} K_0$.
Similarly, let $K_1$ be the matrix obtained by column normalization of $K'_0$ with respect to $\nu$.
Then $K_1 = K'_0 D_0^c$, where $D_0^c = \diag (\nu/ \mathbf{I}_m^T K'_0)$.
Iteratively, we have $K_s = \Pi_{i=1}^{s} D_i^r \cdot K_0 \cdot \Pi_{i=1}^{s} D_i^c$.

The \textbf{only if direction} [${C} \in \Phi^{-1} (T) \Longrightarrow$ existence of $D^r$ and $D^c$ for any $\epsilon$]: 
${C} \in \Phi^{-1} (T) \Longrightarrow $ $(\mu, \nu)$-Sinkhorn scaling of ${K}$ converges to $T$
$\Longrightarrow $, for any $\epsilon >0$, there exists $N>0$ such that for any $s>N$, $|K_s - T| < \epsilon$, 
where $K_s = \Pi_{i=1}^{s} D_i^r \cdot {K} \cdot \Pi_{i=1}^{s} D_i^c$.
Let $D^r = \Pi_{i=1}^{s} D_i^r$, $D^c = \Pi_{i=1}^{s} D_i^c$, the only if direction is complete.

The \textbf{if direction} [existence of $D^r$ and $D^c$ for any $\epsilon$ $\Longrightarrow$ ${C} \in \Phi^{-1} (T)$]: 
Let the limit of $(\mu, \nu)$-Sinkhorn scaling on ${K}$ be $K^{*}$.
Hence $K^{*}$ and $T$ have the same marginals and pattern. 
According to Lemma~A.3 of \citep{pei2019generalizing}, $K^{*}$ and $T$ are diagonally equivalent. 
Further by Proposition~1 of \citep{pretzel1980convergence} $K^{*} = T$.

In particular, when $T$ is a positive matrix, ${K}$ must also be a positive matrix.
Then $T$ and ${K}$ have the same pattern. 
$T = \lim_{s\to \infty} \Pi_{i=1}^{s} D_i^r \cdot {K} \cdot \Pi_{i=1}^{s} D_i^c$
implies that $\lim_{s\to \infty} \Pi_{i=1}^{s} D_i^r$ and $\lim_{s\to \infty} \Pi_{i=1}^{s} D_i^c$ exist. 
Hence the claim. \citep{rothblum1989scalings, idel2016review}

\end{proof}

\begin{replemma}{lemma:cr_dr}
For two positive matrices $A, B$, 
$A \overset{c.r.}{\sim}B$ if and only if 
there exist positive diagonal matrices $D^r$ and $D^c$ such that $A = D^r B D^c$.
\end{replemma}
\begin{proof}
The \textbf{if direction}:
$A = D^r B D^c$  $\Longrightarrow$ 
$\displaystyle \frac{a_{ik}a_{jl}}{a_{il}a_{jk}} = \frac{d^r_ib_{ik}d^{c}_{k} \cdot d^{r}_{j}b_{jl}d^{c}_l}{d^{r}_ib_{il}d^{c}_{l} \cdot d^{r}_jb_{jk}d^{c}_k} 
=  \frac{b_{ik}b_{jl}}{b_{il}b_{jk}}$  $\Longrightarrow$  $A \overset{c.r.}{\sim}B$.

The \textbf{only if direction}: Let the dimension of $A, B$ be $m\times n$, we will prove by induction on $m+n$. 
\textit{Step1.} Assume the dimension of $A, B$ is $2\times 2$.
Let the marginal of $A$ be $\mu_A$ and $\nu_A$, and  $B^* = \Phi (B, \mu_A, \nu_A)$. Then $A$ and $B^*$ 
have the same marginals and cross ratios, which put four same independent constraints on elements of $A$ and $B$,
hence $A =  B^* = D^r B D^c$  for some $D^r, D^c$.
\textit{step2.} Assume the statement holds for $m+n< N$. 

\newpage
Now assume that $m+n = N$. 
Denote the submatrices of $A$ and $B$ consisted by their first $n-1$ columns by $A_{1}$ and $B_{1}$.
$A\overset{c.r.}{\sim} B \Longrightarrow A_{1}\overset{c.r.}{\sim} B_{1}$. By the inductive assumption, there exist
diagonal matrices $D^r_1$ and $D^{c}_1$ such that $A_{1} = D^r_1 B_1 D^c_1$. 
Further $\frac{a_{11}a_{1n}}{a_{i1}}a_{in} = \frac{b_{11}b_{1n}}{b_{i1}}b_{in}$ holds for any $i\in \{1,\dots, m\}$ imply that
there exists an $d>0$ such that $\mathbf{a}_{n} = d \cdot \mathbf{b}_{n}$. Let $D^r = D^r_1$, $D^{c} = \diag{D^c_1, d}$, 
we have $A = D^rB D^c$ holds. Hence the only if direction is completed.
\end{proof}

\begin{reptheorem}{thm:IOT}
Let $T$ be an observed positive optimal coupling of dimension $m\times n$. 
Then $\Phi^{-1} (T)$ is a hyperplane of dimension 
$m+n-1$ embedded in $ (\mathbb{R}^*)^{m\times n}$, which consists all the cost matrices 
that of the form:
\begin{equation}\label{supeq:cr_equiv}
 \Phi^{-1} (T) = \{C \in (\mathbb{R}^*)^{m\times n} | K= e^{-C} \overset{c.r.}{\sim} T\}.    
\end{equation}
\end{reptheorem}

\begin{proof}
Combining Proposition~\ref{prop:DAD} and Lemma~\ref{lemma:cr_dr}, we have Eq.~\eqref{supeq:cr_equiv} holds.
Hence $\Phi^{-1} (T)$ contains all the matrices $C$ satisfying the following set of equations:
$\frac{k_{ik} k_{jl}}{k_{il} k_{jk}} = \frac{t_{ik} t_{jl}}{t_{il} t_{jk}} $ 
for any $i, j \in \{1, \dots, m\}$ and $k, l \in \{1, \dots, n\}$, where $K = e^{-C}$, i.e. $k_{ik} = e^{-c_{ik}}$.
Thus $\Phi^{-1} (T)$ is the solution set of the system of linear equations:
\begin{equation} \label{supeq:linear_sys}
    c_{il}+c_{jk} - c_{ik}-c_{jl} = \ln (t_{ik} t_{jl}) - \ln (t_{il} t_{jk})
\end{equation}
Further Remark~\ref{rmk:basis} shows that there are only $(m-1)(n-1)$ independent equations in the system \eqref{supeq:linear_sys}.
Hence $\Phi^{-1} (T)$ is a hyperplane of dimension $mn - (m-1)(n-1) = m+n-1$.
\end{proof}

\begin{repcorollary}{cor:proj_single_column}
Under prior $\mathbf{P_2}$,
the projection of $\text{supp}[P(K|T)]$ onto each column is a $(m-1)$-dimensional manifold that is homeomorphic to the simplex $\Delta_{m-1}$. 
\end{repcorollary}

\begin{proof}
According to prior $\mathbf{P_2}$, for any $K \in \text{supp}[P(K|T)]$, each column of $K$ sums to $1$. 
Hence the projection of $\text{supp}[P(K|T)]$ onto any $j$-th column is embedded in the simplex $\Delta_{m-1}$. 
To show `homeomorphic ', 
we only need to show that for each $v \in \Delta_{m-1}$, there exists a $K \in \text{supp}[P(K|T)$ such that $\mathbf{k}_{j} =v$.
As discussed in Remark~\ref{rmk: one_free_column}, let $D'^{r} = \diag(v/\mathbf{t}_{j})$, 
$K' = \text{Col}( D'^{r}T)$. It's easy to check that $K'\in \text{supp}[P(K|T)]$, and $\mathbf{k'}_{j} = v$.
Hence the corollary holds.
\end{proof}
% \textit{Cross ratio equivalent} is defined in \citep{pei2019generalizing} as following:
% \begin{definition}
% Let $A = (a_{ij})_{n\times n}$ be a square matrix and $S_n$ be the set of all permutations of length~$n$.
% For any $\sigma \in S_n$, the set of $n$-elements $\{a_{1,\sigma(1)}, a_{2,\sigma(2)}, \dots,a_{t,\sigma(n)}\}$
% is called a \textbf{diagonal} of $A$.
% If every $a_{k, \sigma(k)} > 0$, then the diagonal is said to be \textbf{positive}.
% \end{definition}

% \begin{definition}\label{def:cross ratios}
% Let $A, B$ be two $n\times n$ matrices and $D^A_1=\{a_{1,\sigma(1)},\dots, a_{n, \sigma(n)}\}$ and $D^A_2=\{a_{1,\sigma'(1)},\dots, a_{n, \sigma'(n)}\}$ be two positive diagonals of $A$ determined by permutations $\sigma, \sigma'\in S_n$. 
% Denote the products of elements on $D^A_1$ and $D^A_2$ by $d^A_1=\Pi_{i=1}^{n}a_{i, \sigma(i)}, d^A_2=\Pi_{i=1}^{n}a_{i, \sigma'(i)}$ respectively. Then $\CR(D^A_1, D^A_2)=d^A_1/d^A_2$ is called the \textbf{cross ratio} between $D^A_1$ and $D^A_2$ of $A$. Further, let the diagonals in $B$ determined by the same $\sigma$ and $\sigma'$ be 
% $D^B_1=\{b_{1,\sigma(1)},\dots, b_{n, \sigma(n)}\}$ and $D^B_2=\{b_{1,\sigma'(1)},\dots, b_{n, \sigma'(n)}\}$. We say $A$ is \textbf{cross ratio equivalent} to $B$, denoted by $A\overset{cr}{\sim} B$, if $d_i^A\neq 0 \Longleftrightarrow d_i^B\neq 0$
% and $\CR(D^A_1, D^A_2)=\CR(D^B_1, D^B_2)$ holds for 
% any $D^A_1$ and $D^A_2$.
% \end{definition}

\begin{repproposition}{prop:submanifold}
$C_s \in \text{supp}[P(C_s|T)]$ if and only if there exists positive diagonal matrices $D_s^c, D_s^r$ such that $K_s = D_s^c T_s D_s^r$ and
the system of equations shown below have a positive solution for $\{x_1, \dots, x_{m-s_1}\}$.
\begin{equation}\label{supeq: submainfold}
  (x_1, \dots, x_{m-s_1}) T_{m-s}  = (1/d^c_1, \dots, 1/d^c_{s_2}) - \mathbf{1}_s D_s^rT_s  
\end{equation}

\end{repproposition}

\begin{proof}
The \textbf{if direction}: Let $(x_1, \dots, x_{m-s_1})$ be a positive solution of Eq.~\eqref{supeq: submainfold}, 
let $D'^{r} = \diag (D^{r}_s, x_1, \dots, x_{m-s_1})$ be an extension of $D^r_s$.
Denote the column sum of $D'^{r} T$ by $\nu'$. Then Eq.~\eqref{supeq: submainfold} implies that $\nu'_i = d^c_i$ for 
$i \in \{1, \dots, s_2\}$. Let $D'^c = \diag(D^c_s, \nu'_{s_2+1}, \dots,\nu'_{n})$. 
It is clear that $K' = D'^r T D'^c \in \text{supp}[P(K|T)]$, and $K_s$ is $K'$'s submatrix corresponding to $X_s\times Y_s$.

The \textbf{only if direction}: for $C_s \in \text{supp}[P(C_s|T)]$, let $K_s = e^{-C_s}$ be the submatrix of $K'\in \text{supp}[P(K|T)$.
Then there exists $D'^r, D'^c$ such that $K' = D'^r T D'^c $. Let the corresponding submatrices of $D'^r, D'^c$ be $D^r_s, D^c_s$. 
We have $K_s = D_s^c T_s D_s^r$ hold. Further it is easy to verify that $(d'^r_{s_1+1}, \dots d'^r_{n})$ is a positive solution for 
Eq.~\eqref{supeq: submainfold}. Hence the proof is completed.
\end{proof}

\begin{repcorollary}{cor:missing_element}
Under prior $\mathbf{P_2}$, 
$\mathcal{K}_1$ is a line segment in $\Delta_{m-1}$ that can be parameterized as:
$\mathcal{K}_1 = \{ (d_1t_{11}, \dots, d_m t_{m1})/\sum_{i = 1}^{m} d_it_{i1}| t_{m1} \in (0,\infty) \}$.
\end{repcorollary}

\begin{proof}
Corollary~\ref{cor:proj_single_column} implies that $\mathcal{K}_1 \subset \Delta_{m-1}$. 
Further Remark~\ref{rmk: one_free_column} implies that for each choice of $t_{m1} >1$, 
a known $\mathbf{k}_l$ uniquely determines a point in form of $(d_1t_{11}, \dots, d_m t_{m1})/\sum_{i = 1}^{m} d_it_{i1}$ 
in $\mathcal{K}_1$. Hence the corollary holds.
\end{proof}

\begin{repcorollary}{cor:parallel}
Let $T_1, T_2$ be two positive matrices of dimension $m \times n$.
The hyperplanes $\Phi^{-1}(T_1)$ and $\Phi^{-1}(T_2)$, have the same normal direction.
In particular, if $T_1 \overset{c.r.}{\sim}T_2$ then $\Phi^{-1}(T_1) = \Phi^{-1}(T_2)$.
Otherwise $\Phi^{-1}(T_1)$ is parallel to $\Phi^{-1}(T_2) $.
\end{repcorollary}

\begin{proof}
According to the proof of Theorem~\ref{thm:IOT} above, 
both $\Phi^{-1}(T_1)$ and $\Phi^{-1}(T_2)$ are defined by system of equations in form of
Eq.~\eqref{supeq:linear_sys}. In particular, they have the same coefficients, 
only the constants on the right side of the equations are different.
Hence, $\Phi^{-1}(T_1)$ and $\Phi^{-1}(T_2)$, have the same normal direction.
\end{proof}

\begin{repproposition} {prop: bounded_noise}
For a coupling $T$,  assume uniform observation noise on $t_{11}$ with bounded size $a$, 
\begin{equation}\label{supeq: bounded_noise}
 \text{supp}[P(C|T)] = \cup_{T'\in \mathbb{B}_{a}(T)} \Phi^{-1}(T'),   
\end{equation}

where $\mathbb{B}_{a}(T)$ is the set of matrices $T'$ of the same dimension as $T$ with the property that:
$t'_{11}>0$, $|t'_{11} - t_{11}| \leq a$ and $t'_{ij} = t_{ij}$ for other $i, j$.
Moreover, $\Phi^{-1}(T')$ can be expressed as intersection of two hyperplanes (may be in different dimensions):
one with equation: $c_{11} + c_{22} - c_{21} - c_{12} = - \ln\frac{t'_{11} t_{22}}{t_{21}t_{12}}$, 
and the other equation does not depend on the value of $t'_{11}$. Assume the angle between these two hyperplanes is $\theta$.
Then $d(\Phi^{-1}(T_1'), \Phi^{-1}(T_2')) \leq \ln \frac{t_{11} + a}{t_{11} - a} /\sin \theta $ for $T_1', T_2' \in \mathbb{B}_{a}(T)$.
\end{repproposition}

\begin{proof}
With bounded noise on $t_{11}$, the domain for $t^*_{11}$ is $[t_{11}-a, t_{11}+a] \cap \{t^*_{11} > 0\}$.
Hence Eq.~\eqref{supeq: bounded_noise} holds. $\Phi^{-1}(T')$ is defined by the cross ratios of $T$ as shown in 
Eq.~\ref{supeq:linear_sys}. Using cross-ratio basis say $\mathcal{B} =\{r_{mjnk}(T')| j = 1, \dots, m-1, k=1,\dots, n-1 \}$, Eq.~\ref{supeq:linear_sys} 
can be simplified into a system with only independent $(m-1)(n-1)$, where only the equation defined by $r_{m1n1}(T')$ involves $t'_{11}$,
the other equations can be combined into one equation by substitutions.
Each of these two equations determines a hyperplane,
hence $\Phi^{-1}(T')$ can be expressed as the intersection of two hyperplanes.
Moreover, let the defining hyperplanes for $T^1, T^2 \in \mathbf{B}_a(T)$ be $\{H^1_1, H^1_2\}$ and  $\{H^2_1, H^2_2\}$ respectively,
where $H^i_1$ is the one defined by $ c_{11}+c_{mn} - c_{m1}-c_{1n} = \ln (t_{m1} t_{1n}) - \ln (t^i_{11} t_{mn})$, where $i =1,2$.
Then the Euclidean distance between $H^1_1, H^1_1$ is $d(H^1_1, H^1_1) = |\ln t^1_{1,1} - \ln t^2_{1,1}| \leq |\ln (t_11 +a) - \ln (t_11 -a)|$.
Notice that $H^1_2 = H^2_2$. So the distance between $\Phi^{-1}(T^1), \Phi^{-1}(T^2)$ is bounded by $(\ln (t_{11} +a) - \ln (t_{11} -a))/\sin\theta$.
\end{proof}

\begin{repproposition}{prop: gaussian_noise}
Let $T$ be an observed coupling of dimension $m\times n$ with Gaussian noise on $t_{11}$.
Further, let $\mathcal{B}$ be a basis for cross ratios of $m\times n$ matrices, that contains only one cross ratio depending on $t_{11}$.
Eliminate the cross ratio depending on $t_{11}$ in $\mathcal{B}$, denote the new set by $ \mathcal{B}^{-}$. Then: 

\centerline{$\text{supp}[P(C|T)] = \{C|r(K) = r(T) \text{ for } r \in \mathcal{B}^{-}\}$}
In particular, $P(C|T)$ is supported on a hyperplane that is one dimensional higher than $P(C|T^*)$.
\end{repproposition}

\begin{proof}
Since the domain for $\epsilon \sim \mathcal{N}(0,\sigma^2)$ is $(-\infty, \infty)$, 
hence the domain for possible $t^*_{11}$ is $(0, \infty)$. 
Thus the observation $T$ essentially put no constraint on the cross ratio of $T^*$ depends on $t^*_{11}$.
Hence $\text{supp}[P(C|T)] = \{C|r(K) = r(T) \text{ for } r \in \mathcal{B}^{-}\}$. 
As only one equation was eliminated, so $P(C|T)$ is supported on a hyperplane that is one dimensional higher than $P(C|T^*)$.
\end{proof}
\section{Additional details of experiments} \label{supsec: alg}

\subsection{Auto-correlation function.} \label{apd:sec:autocorrelation}

To monitor the efficiency of our MC methods, we compute the auto-correlation function during the burn-in phase and choose the lags accordingly.

The auto-correlation function is defined as
\begin{equation}
R(t)=\frac{1}{(N-t)\sigma^2}\sum_{l=1}^{N-t}\sum_{i,j}(K_{i,j}^{(l)}-\bar{K}_{i,j})\cdot(K^{(l+t)}_{i,j}-\bar{K}_{i,j}),
\label{eq:autocorrelation}
\end{equation}
where $N$ is the total number of samples generated, (i,j) runs over all indices of the matrices $K$, $\bar{K}$ is the mean averaging over all the samples, and $\sigma_K^2=\sum_{i,j}\sigma_{K,i,j}^2$ is the variance. 

\subsection{Re-normalization of $T$ for MetroMC over prior $\mathbf{P}_1$.}

For using this method, not every random $T$ satisfies the condition such that $C^{(0)}=-\text{ln}(K^{(0)})=-\text{ln}(T)$ lies on the support of the posterior, therefore, we need to re-normalize $T$ so that $C^{(0)}-\text{ln}[T/F(T,\lambda)]$ is on the support. $F(T,\lambda)$ is defined as
\begin{equation}\label{eq:renormal_T}
    F(T, \lambda) = \text{exp}\left({\frac{\lambda+\sum_{i=1}^{m}\sum_{j=1}^{n}\text{ln}(T_{i,j})}{m*n}}\right).
\end{equation}

\begin{figure}
\centering
  \includegraphics[width=0.3\textwidth]{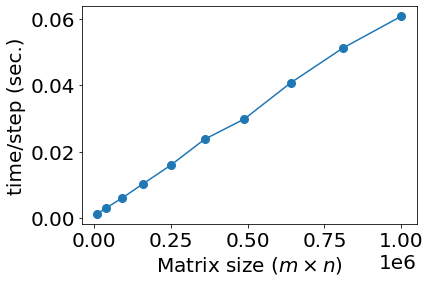}
  \caption{PIOT running time per MC step v.s. matrix size.}
  \label{fig:run_time}
\end{figure}

\subsection{Running time of PIOT method.} 
The PIOT inference has a linear computational complexity of $\mathcal{O}(mn)$ ($m$ and $n$ are the dimensions of the matrix.) as shown in Fig.~\ref{fig:run_time}. The slope can be further reduced by optimizing the implementation. Prediction uses the entropy-regularized OT (EOT). Note that our algorithm is highly parallelizable.

\subsection{MHMC on uniform matrix.} \label{apd:sec:MHMC}
In this section, we validate our MHMC algorithm by apply it to an uniform $T$, i.e $T_{m,n}=1/m$ 
under prior $\mathbf{P}_2$.
%on uniform matrices, e.g. $T_{m,n}=1/m$ for the column-wise Dirichlet prior case. 
The cross-ratios of $T$ are all 1 and stay the same for all inferred sample $K^{(i)}$. 
Thus, each $K^{(i)}$ has same copies of $n$ columns. 
We compare posterior sampled by MHMC to a 
$m$-dimensional uniform symmetric Dirichlet distribution generated by the SciPy package \citep{2020SciPy-NMeth}.

% Fig.~\ref{fig:uniform_3x3} shows the results the 3x3 case.
% are the same for every element. We choose $\alpha=1$ for the Dirichlet prior, 
% $\sigma_0=0.5$, $\gamma=3$, and $\delta=1.0$. 
% We run for 10,000 burn-in steps and take 10,000 samples 
% with lags of 100. The result is shown in Fig.~\ref{fig:uniform_3x3}. 
% We choose the lag by observing the auto-correlation function in the burn in, where the chain becomes uncorrelated after $\sim$ 100 steps.
% Also, the running averages of the row sums are stable at taking 10,000 samples, 
% which means 10,000 samples can effectively represent the posterior distribution. 

\begin{figure}[h!]
    \centering
    \begin{tabular}{c c}
        \hspace{-4pt}\includegraphics[width=0.33\textwidth]{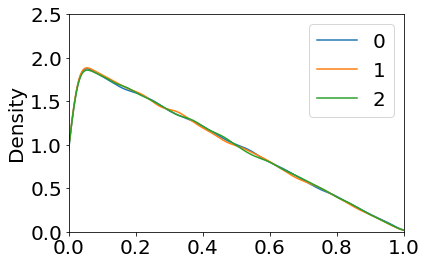}
        \hspace{-6pt}\includegraphics[width=0.33\textwidth]{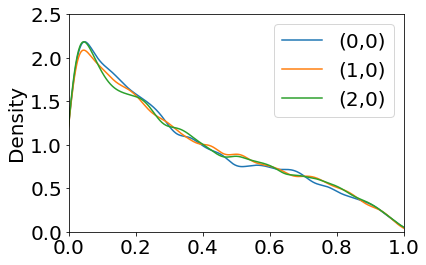}
    \end{tabular}
    \begin{tabular}{c c}
        \hspace{-4pt}\includegraphics[width=0.33\textwidth]{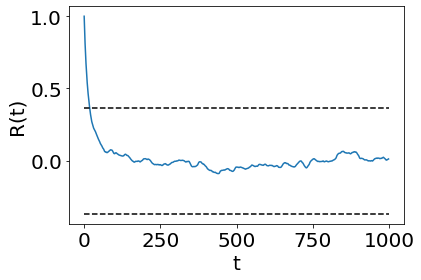}
        \hspace{-6pt}\includegraphics[width=0.33\textwidth]{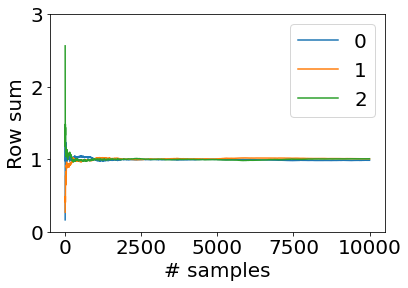}
    \end{tabular}
    
    \caption{\textbf{(Top left.)} The distributions of each component of three-dimensional Dirichlet vectors with symmetric $\alpha=1$ generated using SciPy package. \textbf{(Top right.)} The distributions of each component of the first column of $P(K|T)$ sampled by the MHMC method. \textbf{(Lower left.)} The auto-correlation function of the MHMC simulation. The black dashed lines indicate the range [$-1/e$, $1/e$]. \textbf{(Lower right.)} The running average of the row sums of the MHMC simulation.
    }
    \label{fig:uniform_3x3}
\end{figure}

We perform our MHMC algorithm on a $3\times 3$ uniform matrix with prior $P_2$ and the Dirichlet concentration parameter $\alpha$ are the same for every element. We choose $\alpha=1$ for the Dirichlet parameter, $\sigma_0=0.5$, $\gamma=3$, and $\delta=1.0$. We run for 10,000 burn-in steps and take 10,000 samples with lags of 100. The result is shown in Fig.~\ref{fig:uniform_3x3}. We choose the lag by observing the auto-correlation function in the burn in, where the chain becomes uncorrelated after $\sim$ 100 steps. Also, the running averages of the row sums are stable at taking 10,000 samples, which means 10,000 samples can effectively represent the posterior distribution. 

The posterior distribution (Shown in top right of Fig.~\ref{fig:uniform_3x3}.) is compared to the distribution $P(K|T)$ of three-dimensional vectors generated by Dirichlet vector sampler \citep{2020SciPy-NMeth} with symmetric $\alpha=1$ in top left of  Fig.~\ref{fig:uniform_3x3}. The results are comparable, which suggests that our MHMC method is able to generate distributions which well represent the support of the posterior distributions.

\subsection{MetroMC on uniform matrix.}
\begin{figure}[h!]
    \centering
    \includegraphics[width=0.33\textwidth]{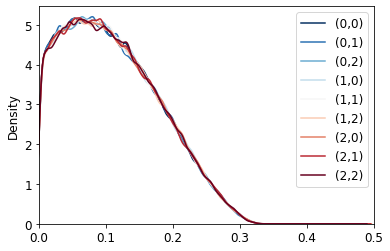}
    \begin{tabular}{c c}
        \includegraphics[width=0.33\textwidth]{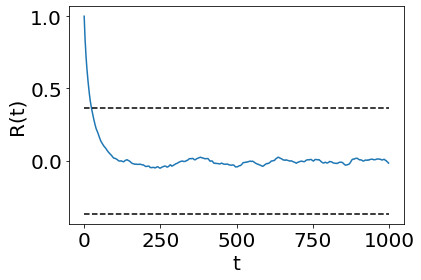}
        \includegraphics[width=0.33\textwidth]{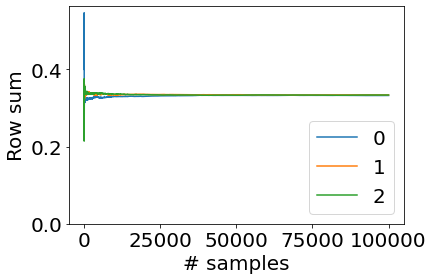}
    \end{tabular}
    \caption{\textbf{(Top.)} The distribution of each matrix element of $P(C|T)$ with a uniform $T$ and a prior $\mathbf{P}_1$ over $C$ matrix. \textbf{(Bottom left.)} The autocorrelation function of the simulation. \textbf{(Bottom right.)} The running averages of the row sums.}
    \label{fig:uniform_3x3_2}
\end{figure}

We test MetroMC on 3x3 uniform $T$, where $T_{i,j}=1/(m*n)$, and assume the prior $\mathbf{P}_1$ is put on $C$. We take $K^{(0)}=T/F(T, \lambda=1)$.
For each MC iteration we sample $m+n-1$ Gaussian random numbers with standard deviations of 0.02 for the diagonal matrices $D^r$ and $D^c$ satisfying the condition 
\begin{equation}
\sum_{i,j}\text{ln}(d^r_{i})+\text{ln}(d^c_j)=0. 
\label{eq:condition_bounded_noises}
\end{equation}
We burn in 10,000 steps and take 100,000 samples with 100 lags in between. The Gaussian Kernel density estimation (KDE) of each component of $P(C|T)$ is demonstrated in Fig.~\ref{fig:uniform_3x3_2}. We use bandwidths of 0.05 for all the Gaussian KDE in this article. There are a few notable features. First, we assume a Dirichlet prior on the whole $C$ matrix, the distributions will be smaller at large values because each large value will suppress the value of the other elements. Second, since we have cross-ratio = 1 everywhere in $T$ and prior $\mathbf{P}_1$ on $C$, there will not be a single element with value close to one. Otherwise, the cross-ratio involving that element could never be 1 anymore. Also, for the same reason all distributions should be similar.

\subsection{Visualizing $\text{supp}[P(K|T)]$ with incomplete observations.}\label{apd:sec:missing_value}
\begin{figure}[h!]
    \centering
    \includegraphics[width=0.5\textwidth]{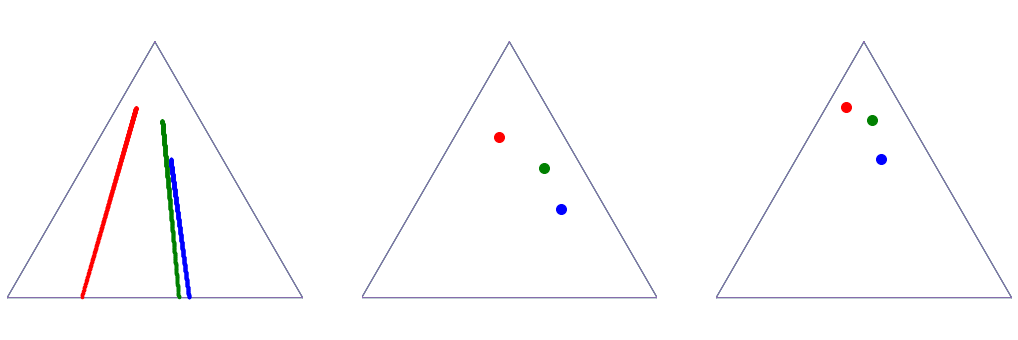}
    \caption{Visualization of $\text{supp}[P(K|T)]$ for a $3\times 3$ observed $T$ with $t_{31}$ missing.
    Three sampled $K \in \text{supp}[P(K|T)]$ are shown here with 
    each column is plotted in a copy of $\Delta_2$ as shown.
    Colored dots in the middle $\Delta_2$ represents uniformly sampled $\mathbf{k}_2$'s. 
    The uniquely determined $\mathbf{k}_3$'s are shown on the right, 
    the set $\mathcal{K}_1$ for the corresponding $\mathbf{k}_1$'s is shown on the left.}
    \label{fig:missing_element_T}
\end{figure}

Fig.~\ref{fig:missing_element_T} illustrates the 
$\text{supp}[P(K|T)]$ for a $3\times 3$ observation $T$ with $t_{31}$ missing under prior $\mathbf{P}_2$. % \hl{FIXME}
%$T = [[5.84, 4.32, 4.10],[5.90, 9.00, 4.29],[t_{31}, 5.49, 7.18]]$ is constructed by sampling each element uniformly from $[0,10]$.
Columns of three sampled $K \in \text{supp}[P(K|T)]$ are plotted in three colored copies $\Delta_2$ as shown.
Here, three $\mathbf{k}_2$ are sampled uniformly from the middle $\Delta_2$. The uniquely determined $\mathbf{k}_3$ are shown in the right $\Delta_2$.
The corresponding set for $\mathbf{k}_1$ ($\mathcal{K}_1$) is plotted in the left $\Delta_2$.
$\mathcal{K}_1$ is obtained by uniformly sample a thousand $t_{31}$ from $[0,10]$. 
Each $t_{31}$ uniquely determines a $\mathbf{k}_1$.
As explained in Corollary~\ref{cor:missing_element}, 
each $\mathcal{K}_1$ forms a line segments. 
The slope of each line is determined by $T$'s cross ratios. 

\subsection{Monte Carlo results for synthetic symmetric costs.}\label{apd:ma}

\begin{figure}[h!]
    \centering
    \includegraphics[width=0.33\textwidth]{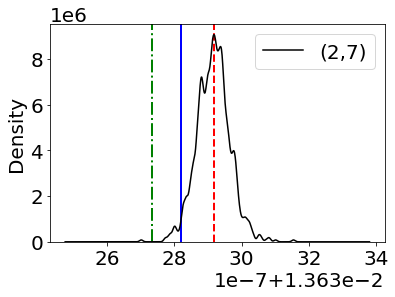}
    \includegraphics[width=0.33\textwidth]{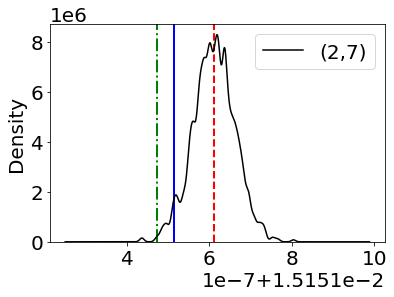}
    \includegraphics[width=0.33\textwidth]{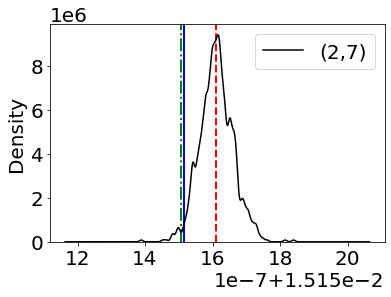}
    \caption{The KDE of the inferred posterior cost function of the matrix element $C_{2,7}$. The blue solid lines indicate the ground truth cost, the green dash-dotted lines are results using algorithm from \citep{ma2020learning} at 300-iteration, and the red dashed lines are the median of each inferred posterior distribution of the cost function. (Left.) $p=0.5$. (Middle.) $p=1$. (Right.) $p=2$. At 1000 iteration, their algorithm converge to the ground truth cost.}
    \label{fig:sym_cost}
\end{figure}

For the symmetric costs we choose a constrained prior: $P_0(C)$ follows a Gibbs distribution $\text{exp}(-\beta(\parallel \gamma(C-C^T) \parallel)$. We take $\beta=10$ and $\gamma=10^6$. We use $\lambda=10$ in this example. Before the MCMC simulation we first pre-process $K^{(0)}=\text{diag}(D^{r_0})F(\frac{T^{(0)}}{\mu\nu^T},\lambda=10)\text{diag}(D^{c_0})$, where $\text{diag}(D^{r_0})$ and $\text{diag}(D^{c_0})$ are diagonal matrices such that the diagonals of $K^{(0)}$ are 1 and $F(\cdot)$ is the re-normalization function as described in \ref{eq:renormal_T}. Then, during MetroMC we sample $D^{r} =$ exp$( N(0, \sigma * \mathbb{I}_{n}) )$ and set $D^{c} = 1/D^{r}$. We use $\sigma=10^{-6}$ for the burn-in steps and $\sigma=10^{-7}$. We burn-in 100,000 steps and take 1,000 samples with lags of 300 steps. The acceptance ratios are 0.6-0.7. The KDE of the sampled posterior distribution $C_{2,7}$ are shown in Fig.~\ref{fig:sym_cost} for $p=$ 0.5 (left), 1 (middle), and 2 (right). The blue vertical lines indicate value of the the ground truth cost and the red dashed lines are the median of the distributions. The averaged inferred $C_{2,7}$ at 300 iteration over 20 instances using algorithm from \citep{ma2020learning} is shown in green dashed lines. At 1000 iteration, their algorithm converge to the ground truth cost. The other elements show similar behaviors. Note that the scale in x-axis is $10^{-7}$, so the differences between the ground truth and the inferred costs are small. The PIOT achieves achieve accurate result with relative error falls below $10^{-5}$ in all cases.  

\begin{figure}[h!]
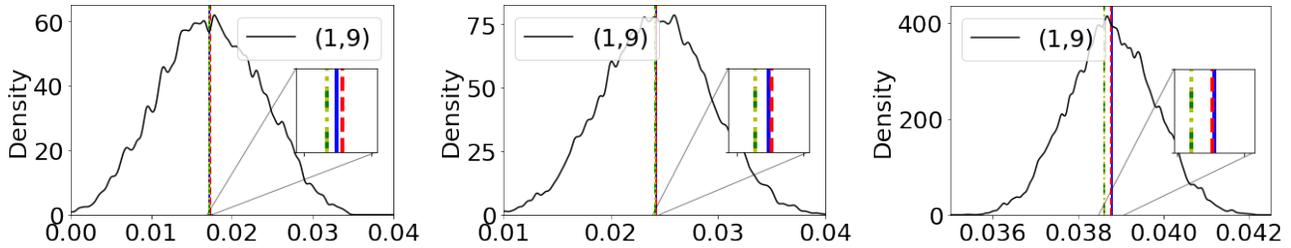

    \centering
    \includegraphics[width=0.33\textwidth]{figs/non-sym_cost_p0.5.png}
    \includegraphics[width=0.33\textwidth]{figs/non-sym_cost_p1.png}
    \includegraphics[width=0.33\textwidth]{figs/non-sym_cost_p2.png}
    \caption{The KDE of the inferred posterior cost function of the matrix element $C_{1,9}$. The blue solid lines indicate the ground truth cost, the green dash-dotted lines are results using algorithm from \citep{ma2020learning} at 300-iteration, the yellow dotted lines are Ma's results at 1000-iteration (Overlapped with green lines for all cases.), and the red dashed lines are the median of each inferred posterior distribution of the cost function. (Left.) $p=0.5$. (Middle.) $p=1$. (Right.) $p=2$. The insets have range of $C^g_{1,9}\pm 10^{-4}$.}
    \label{fig:non-sym_cost}
\end{figure}

Next, to test the robustness we add $n\times n$ random matrices with all entries in $[0, 10^{-4})$ to the symmetric costs and then normalize the combined matrices to 1 as the cost functions. We use the prior $\mathbf{P_1}$ as described in section 3. The hyper-parameter $\alpha$ is set to 1 for all entries. We pre-process $K^{(0)}=F(\text{diag}(D^{r_0})\frac{T^{(0)}}{\mu\nu^T}\text{diag}(D^{c_0}),\lambda=10)$, where \text{diag}($D^{r_0})$ and $\text{diag}(D^{c_0})$ are diagonal matrices such that the diagonals of $\frac{T^{(0)}}{\mu\nu^T}$ are 1 and $F(\cdot)$ is the re-normalization function as described in \ref{eq:renormal_T}. In MCMC we sample $D^{r} =$ exp$( N(0, \sigma * \mathbb{I}_{n}) )$ and set $D^{c} = 1/D^{r}$. We choose $\sigma=$ $5*10^{-3}$, $5*10^{-3}$, and $1*10^{-3}$ for $p=$ 0.5, 1, and 2, respectively. We burn-in 100,000 steps and take 10,000 samples with lags of 500 steps between samples. The inferred posterior distribution of $C_{1,9}$ is shown in Fig.~\ref{fig:non-sym_cost}. The blue solid lines indicate the values of the ground truth, and the red dashed lines are the medians of each distribution, and the green dash-dot lines and yellow dotted lines are averaged results over 20 instances using algorithm from \citep{ma2020learning} at 300 and 1000 iterations , respectively. The green and yellow lines overlap, meaning their algorithm converges at around 300 iteration. The inset shows a zoomed-in region of $C^g_{1,9}\pm 10^{-4}$. Overall, the PIOT accurately infers the cost function with smaller relative errors ( $\sim 10^{-3}$ to $10^{-4}$) than \citep{ma2020learning} ($\sim 10^{-2}$ to $10^{-3}$) for all $p$. 
%\pw{plot for ma's algo?, we may include fig 11 in the main text.}
%Note that since we are taking logarithm of $K$ during MCMC, for convenience we add a small constant $10^{-6}$ on the diagonals to avoid numerical issue. The error thus has a lower bound of $10^{-6}$. 

\subsection{Supporting plots for the MetroMC method on noisy T.}

In this section, we provide data for diagnostics of the MetroMC simulations on the noise models in section 5 of the main text.

\begin{figure}[h!]
    \centering
    \begin{tabular}{c c}
        \begin{picture}(140,140)(80,0)% width and height of the picture
            \includegraphics[width=0.44\textwidth]{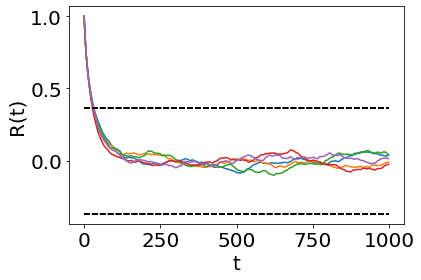}
            \put(-30,125){\textbf{a.}}
        \end{picture}
        \begin{picture}(140,140)(0,0)% width and height of the picture
            \includegraphics[width=0.45\textwidth]{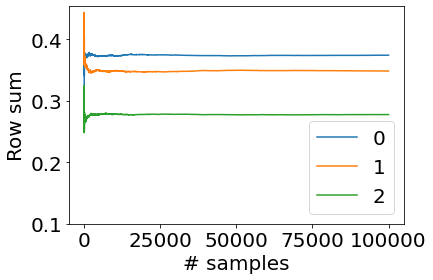}
            \put(-30,125){\textbf{b.}}
        \end{picture}

    \end{tabular}
    \begin{tabular}{c c}
        \begin{picture}(140,140)(80,0)% width and height of the picture
            \includegraphics[width=0.45\textwidth]{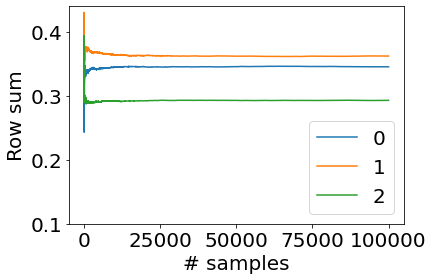}
            \put(-30,125){\textbf{c.}}
        \end{picture}
        \begin{picture}(140,140)(0,0)% width and height of the picture
            \includegraphics[width=0.45\textwidth]{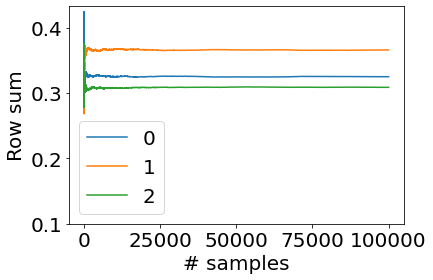}
            \put(-30,125){\textbf{d.}}
        \end{picture}
    \end{tabular}
    \begin{tabular}{c c}
        \begin{picture}(140,140)(80,0)% width and height of the picture
            \includegraphics[width=0.45\textwidth]{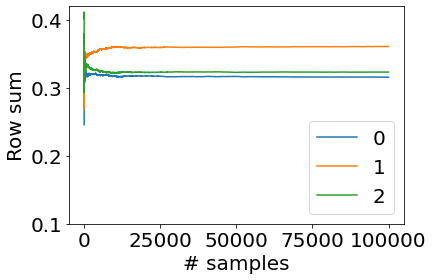}
            \put(-30,125){\textbf{e.}}
        \end{picture}
        \begin{picture}(140,140)(0,0)% width and height of the picture
            \includegraphics[width=0.45\textwidth]{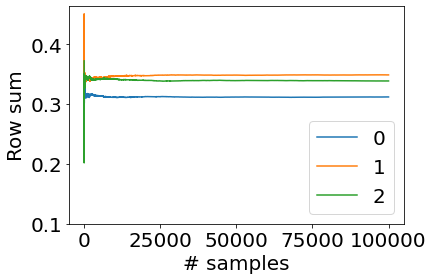}
            \put(-30,125){\textbf{f.}}
        \end{picture}
    \end{tabular}
    \caption{MC diagnostics for the Bounded noise example. \textbf{(a.)} The autocorrelation funciton for all noises. \textbf{(b.)-(f.)} The running averages of row sums of the samples for noises -0.01, -0.005, 0.0, 0.005, and 0.01 added to $t_{11}$, respectively.
    }
    \label{fig:MC_details_bounded_noise}
\end{figure}

\textbf{Bounded noise on observed plan.} The autocorrelation function and the running averages of row sums of the samples for each noise are plotted in Fig.~\ref{fig:MC_details_bounded_noise}. In the simulation we set 100 for the lags by observing that the chains become uncorrelated after $\sim$ 100 steps for all cases. We take 100,000 samples in total for all cases since the running averages are stable at (or before) 100,000 MC steps, which means the samples reaches a stationary distribution.

\begin{figure}[h!]
    \centering
    
    \begin{tabular}{c c c}
        \begin{picture}(140,100)(20,0)% width and height of the picture
            \includegraphics[width=0.31\textwidth]{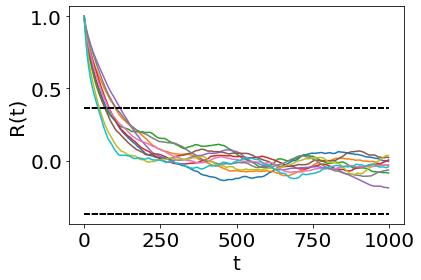}
            \put(-30,80){\textbf{a.}}
        \end{picture}
        \begin{picture}(140,100)(10,0)% width and height of the picture
            \includegraphics[width=0.32\textwidth]{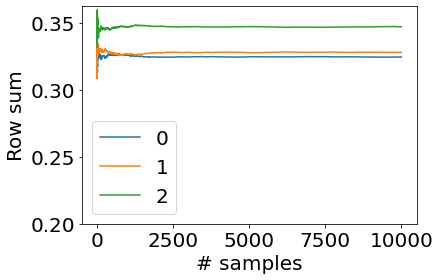}
            \put(-30,30){\textbf{b.}}
        \end{picture}
        \begin{picture}(140,100)(0,0)% width and height of the picture
            \includegraphics[width=0.32\textwidth]{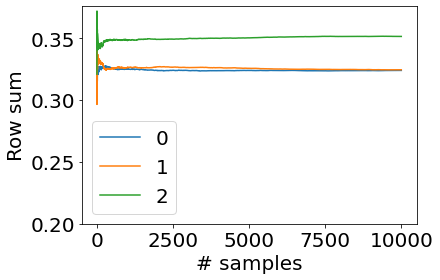}
            \put(-30,30){\textbf{c.}}
        \end{picture}
    \end{tabular}
    \begin{tabular}{c c c}   
        \begin{picture}(140,100)(20,0)% width and height of the picture
            \includegraphics[width=0.32\textwidth]{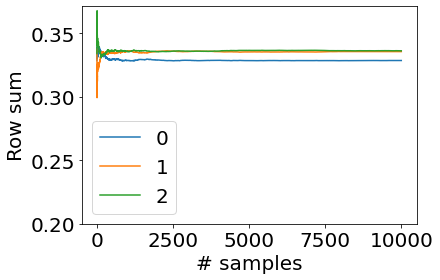}
            \put(-30,30){\textbf{d.}}
        \end{picture}
        \begin{picture}(140,100)(10,0)% width and height of the picture
            \includegraphics[width=0.32\textwidth]{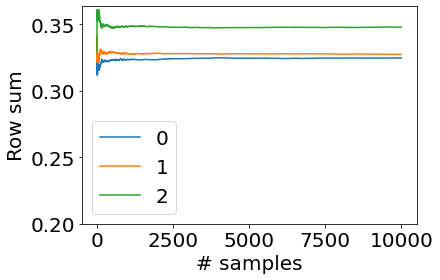}
            \put(-30,30){\textbf{e.}}
        \end{picture}
        \begin{picture}(140,100)(0,0)% width and height of the picture
            \includegraphics[width=0.32\textwidth]{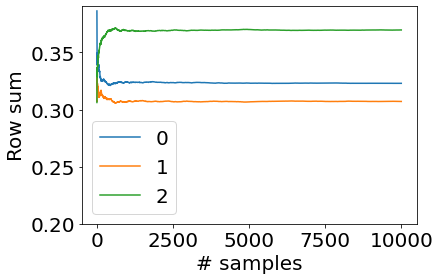}
        \put(-30,30){\textbf{f.}}
        \end{picture}
    \end{tabular}
    \begin{tabular}{c c c}  
        \begin{picture}(140,100)(20,0)% width and height of the picture
            \includegraphics[width=0.32\textwidth]{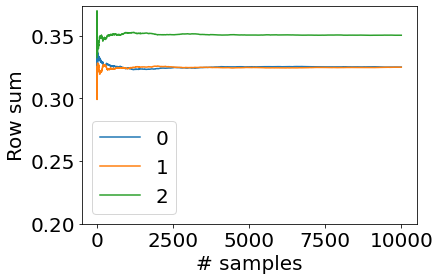}
            \put(-30,30){\textbf{g.}}
        \end{picture}
        \begin{picture}(140,100)(10,0)% width and height of the picture
            \includegraphics[width=0.32\textwidth]{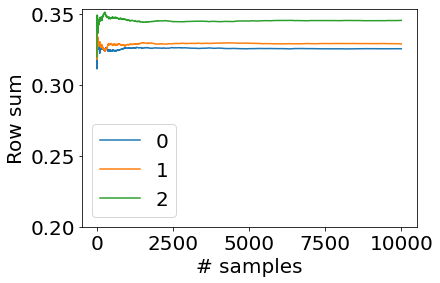}
            \put(-30,30){\textbf{h.}}
        \end{picture}
        \begin{picture}(140,100)(0,0)% width and height of the picture
            \includegraphics[width=0.32\textwidth]{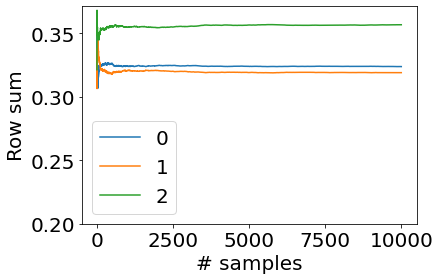} 
            \put(-30,30){\textbf{i.}}
        \end{picture}
    \end{tabular}
    \begin{tabular}{c c c}  
        \begin{picture}(140,100)(20,0)% width and height of the picture
            \includegraphics[width=0.32\textwidth]{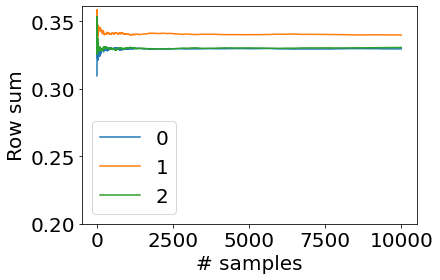}
        \put(-30,30){\textbf{j.}}
        \end{picture}
        \begin{picture}(140,100)(0,0)% width and height of the picture
            \includegraphics[width=0.32\textwidth]{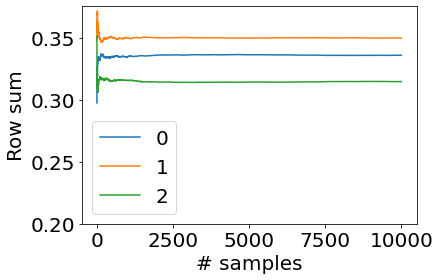}      
            \put(-30,30){\textbf{k.}}
        \end{picture}
    \end{tabular}

    \caption{MC diagnostics for the Bounded noise example. \textbf{(a.)} The autocorrelation function for all noises. \textbf{(b.)-(k.)} The running averages of row sums of the samples for the 10 random Gaussian noises added to $t_{12}$.
    }
    \label{fig:MC_details_running_average_gaussian_noise}
\end{figure}

\textbf{Gaussian noise on observed plan.}
The autocorrelation function and the running averages of row sums of the samples for the 10 Gaussian noises are plotted in Fig.~\ref{fig:MC_details_running_average_gaussian_noise}. We set 200 for the lags since the chains become uncorrelated after $\sim$ 180 steps for all cases. 10,000 samples are taken for all cases since the running averages of the row sums are stable at 10,000 MC steps.

\newpage

\subsection{Matrices.}\label{apd:sec:matrices}

We list the matrices used in section 5: \textbf{Simulations on synthetic data} below.

The three $T$ matrices in Fig.~\ref{fig:2x2_subspace} are

\begin{align*}
T_a = 
\begin{bmatrix}
0.3096 & 0.3785 & 0.0544 & 0.2575 \\
0.2522 & 0.3203 & 0.1860 & 0.2415 \\
0.4318 & 0.1433 & 0.4196 & 0.0053 \\
0.0064 & 0.1579 & 0.3400 & 0.4957 \\
\end{bmatrix},
\end{align*}

\begin{align*}
T_b = 
\begin{bmatrix}
0.2532 & 0.4143 & 0.2894 & 0.0431\\
0.1925 & 0.0548 & 0.0958 & 0.6569\\
0.4459 & 0.0905 & 0.3480 & 0.1156\\
0.1083 & 0.4404 & 0.2669 & 0.1844
\end{bmatrix},
\end{align*}

\begin{align*}
T_c = 
\begin{bmatrix}
0.4790 & 0.0994 & 0.0838 & 0.3378\\
0.1343 & 0.1514 & 0.1920 & 0.5224\\
0.1678 & 0.6182 & 0.1963 & 0.0177\\
0.2189 & 0.1310 & 0.5279 & 0.1222
\end{bmatrix}.
\end{align*}

The matrix $T$ in Fig.~\ref{fig:missing_element_T} is
\begin{align*}
T = 
\begin{bmatrix}
0.4583 & 0.2297 & 0.2633\\
0.4631 & 0.4785 & 0.2755\\
0.0785 & 0.2919 & 0.4611
\end{bmatrix}.
\end{align*}

The matrices $T$ in Fig.~\ref{fig:sample_supp} are

\begin{align*}
T_1 = 
\begin{bmatrix}
0.1104 & 0.0684 & 0.1545 \\
0.0505 & 0.2401 & 0.0428 \\
0.1725 & 0.0249 & 0.1360
\end{bmatrix},
\end{align*}

\begin{align*}
T_2 = 
\begin{bmatrix}
0.0950 & 0.1100 & 0.1283\\
0.1155 & 0.0343 & 0.1835\\
0.1228 & 0.1890 & 0.0215
\end{bmatrix},
\end{align*}

\begin{align*}
T_3 =
\begin{bmatrix}
0.1053 & 0.1193 & 0.1088\\
0.2148 & 0.0090 & 0.1096\\
0.0133 & 0.2051 & 0.1150
\end{bmatrix}.
\end{align*}

The matrix $T$ for the bounded noise case is

\begin{align*}
T =
\begin{bmatrix}
0.1067 & 0.1141 & 0.1125\\
0.1175 & 0.1052 & 0.1106\\
0.1092 & 0.1139 & 0.1102
\end{bmatrix}.
\end{align*}

The ground truth $C^g$ in the Gaussian noise example is
\begin{align*}
C^g = 
\begin{bmatrix}
0.2604 & 0.0521 & 0.0104\\
0.0208 & 0.2604 & 0.0521\\
0.0625 & 0.0208 & 0.2604
\end{bmatrix},
\end{align*}
and the hyperparameter matrix for the prior is
\begin{align*}
\alpha =
\begin{bmatrix}
25.0  & 5.0     & 1.9\\
3.0   & 25.0    & 5.0\\
6.0   & 3.0     & 25.0
\end{bmatrix}.
\end{align*}

\subsection{Details of the MCMC simulations on European migration data.}\label{apd:sec:migration}

\begin{figure}[h!]
    \centering
    \includegraphics[width=0.7\textwidth]{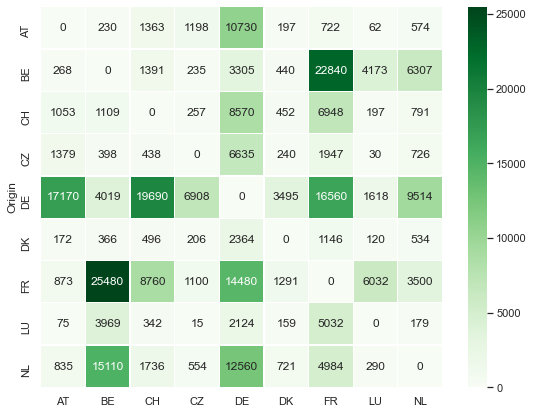}
    \caption{European migration flow matrix between the selected 9 countries. }
    \label{fig:heatmap_migration_only}
\end{figure}

For this part of simulations we consider the migration flows\citep{raymer2013integrated} between the following 9 countries: Austria (AT), Belgium (BE), Switzerland (CH), Czech Republic (CZ), Germany (DE), Denmark (DK), France (FR), Luxembourg (LU) and Netherlands (NL). The migration flow matrix is shown in Fig.~\ref{fig:heatmap_migration_only}.

\begin{figure}[h!]
    \centering
    
    \begin{tabular}{c c c}
        \begin{picture}(140,100)(20,0)% width and height of the picture
            \includegraphics[width=0.31\textwidth]{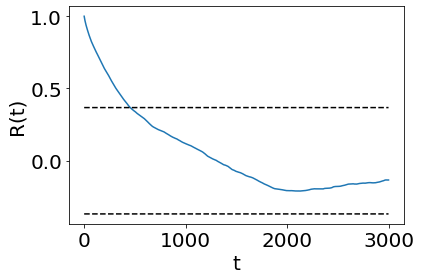}
            \put(-30,80){\textbf{a.}}
        \end{picture}
        \begin{picture}(140,100)(10,0)% width and height of the picture
            \includegraphics[width=0.31\textwidth]{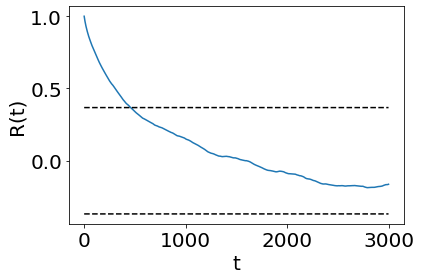}
            \put(-30,80){\textbf{b.}}
        \end{picture}
        \begin{picture}(140,100)(0,0)% width and height of the picture
            \includegraphics[width=0.31\textwidth]{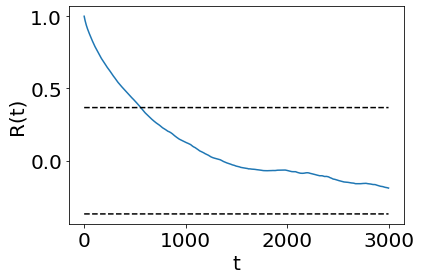}
            \put(-30,80){\textbf{c.}}
        \end{picture}
    \end{tabular}
    \begin{tabular}{c c c}   
        \begin{picture}(140,100)(20,0)% width and height of the picture
            \includegraphics[width=0.31\textwidth]{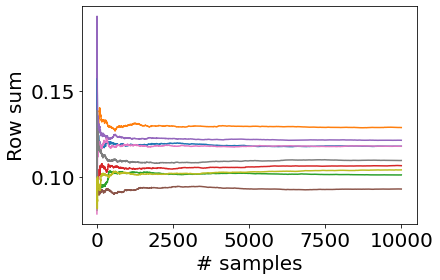}
            \put(-30,80){\textbf{d.}}
        \end{picture}
        \begin{picture}(140,100)(10,0)% width and height of the picture
            \includegraphics[width=0.31\textwidth]{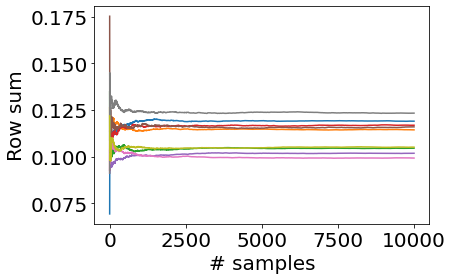}
            \put(-30,80){\textbf{e.}}
        \end{picture}
        \begin{picture}(140,100)(0,0)% width and height of the picture
            \includegraphics[width=0.31\textwidth]{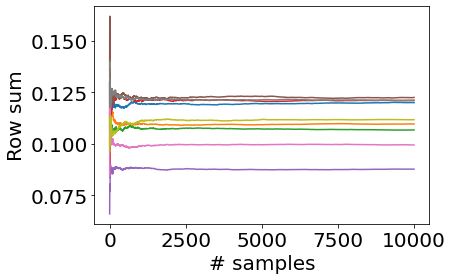}
        \put(-30,80){\textbf{f.}}
        \end{picture}
    \end{tabular}

    \caption{MC diagnostics for the European migration flow data. \textbf{(a.)-(c.)} are the autocorrelation function for using prior 1, 2, and 3, respectively. \textbf{(d.)-(f.)} are the running averages of row sums of the samples for using prior 1, 2, and 3.
    }
    \label{fig:MC_details_EU_migration}
\end{figure}

\begin{figure}[h!]
    \centering
    
    \begin{tabular}{c c c c c c c c c}
        \includegraphics[width=0.105\textwidth]{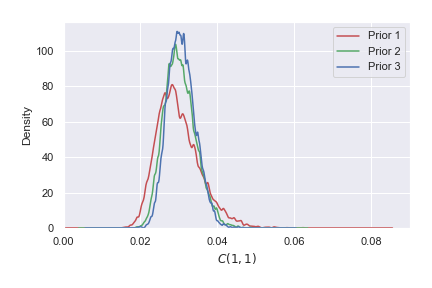}
        \includegraphics[width=0.105\textwidth]{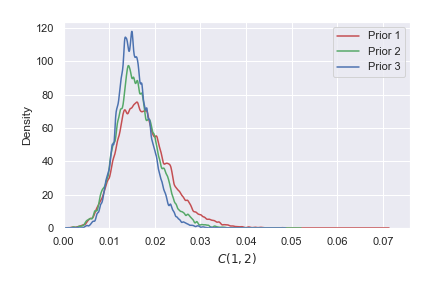}
        \includegraphics[width=0.105\textwidth]{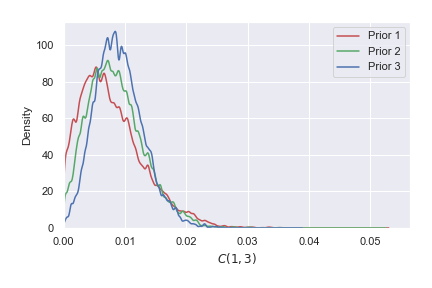}
        \includegraphics[width=0.105\textwidth]{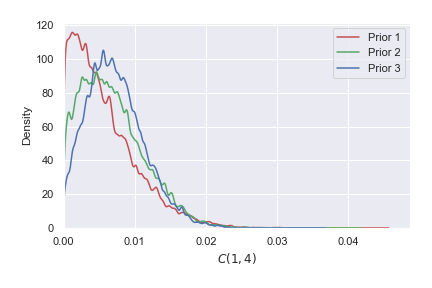}
        \includegraphics[width=0.105\textwidth]{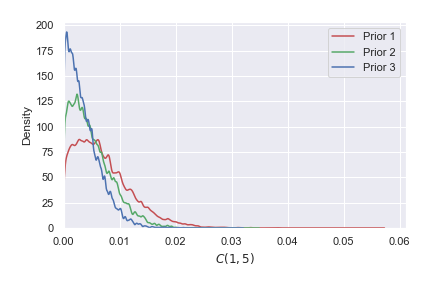}
        \includegraphics[width=0.105\textwidth]{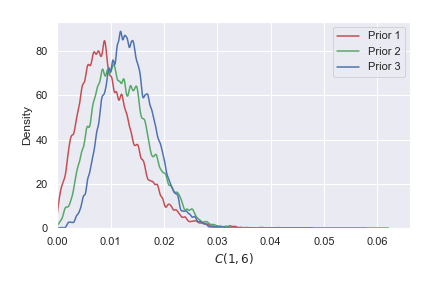}
        \includegraphics[width=0.105\textwidth]{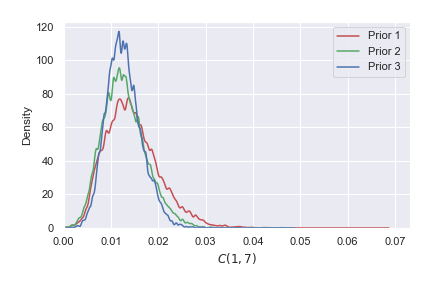}
        \includegraphics[width=0.105\textwidth]{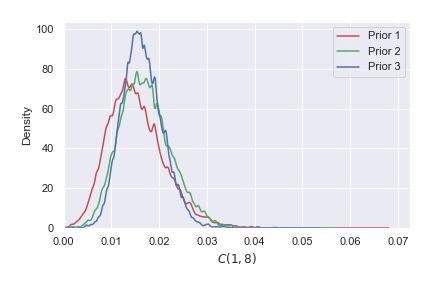}
        \includegraphics[width=0.105\textwidth]{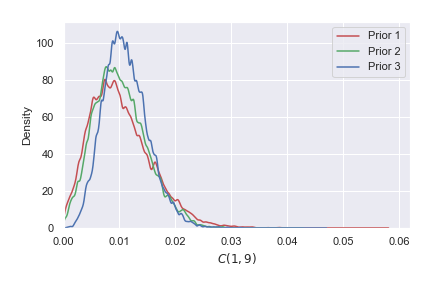}
    \end{tabular}
    \begin{tabular}{c c c c c c c c c}
        \includegraphics[width=0.105\textwidth]{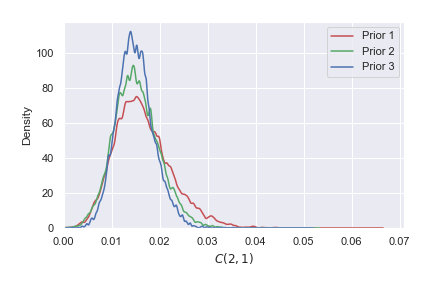}
        \includegraphics[width=0.105\textwidth]{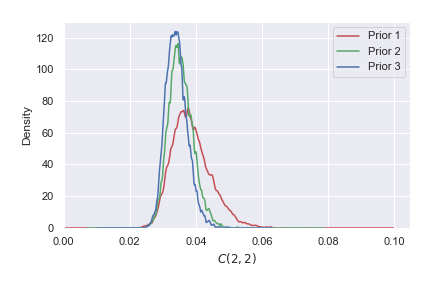}
        \includegraphics[width=0.105\textwidth]{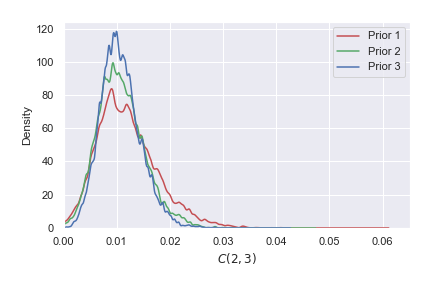}
        \includegraphics[width=0.105\textwidth]{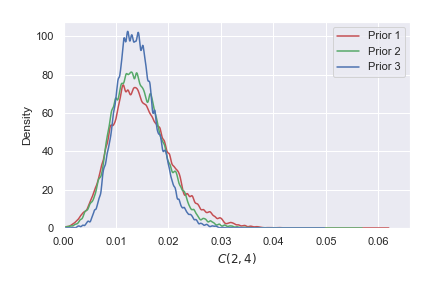}
        \includegraphics[width=0.105\textwidth]{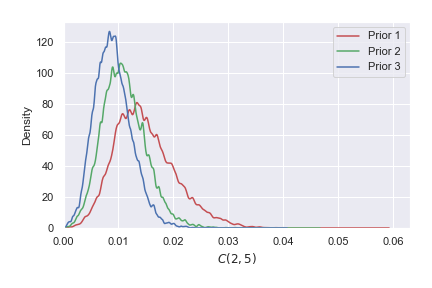}
        \includegraphics[width=0.105\textwidth]{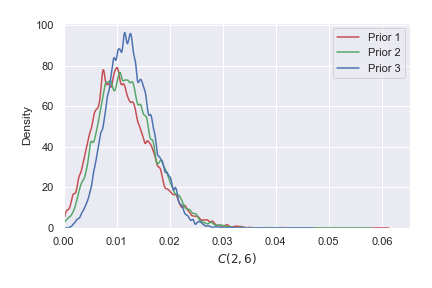}
        \includegraphics[width=0.105\textwidth]{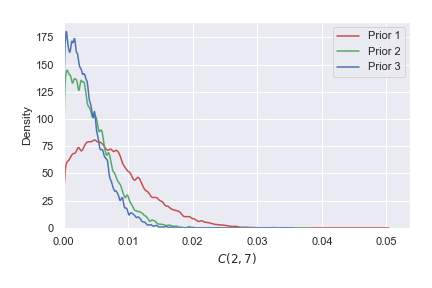}
        \includegraphics[width=0.105\textwidth]{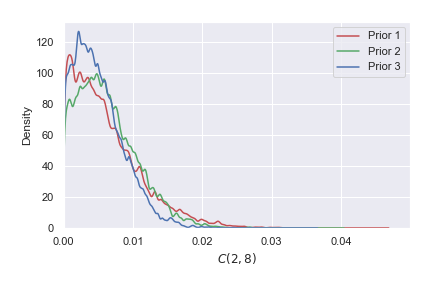}
        \includegraphics[width=0.105\textwidth]{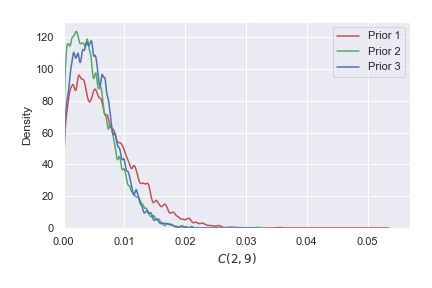}
    \end{tabular}    
    \begin{tabular}{c c c c c c c c c}
        \includegraphics[width=0.105\textwidth]{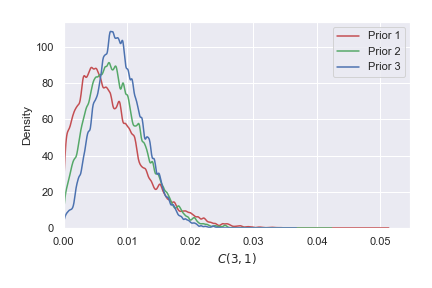}
        \includegraphics[width=0.105\textwidth]{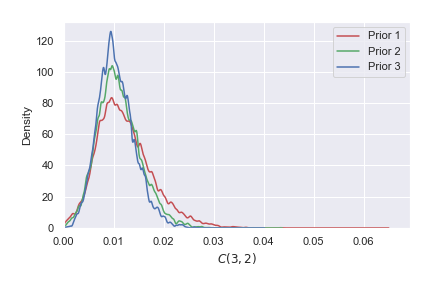}
        \includegraphics[width=0.105\textwidth]{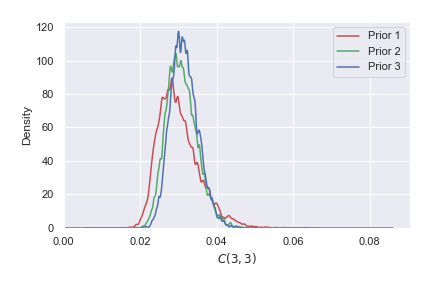}
        \includegraphics[width=0.105\textwidth]{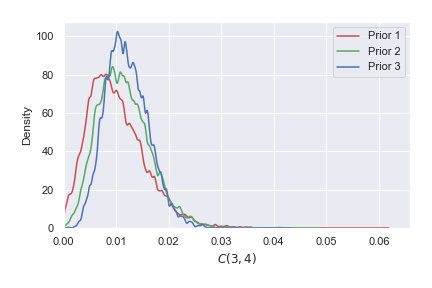}
        \includegraphics[width=0.105\textwidth]{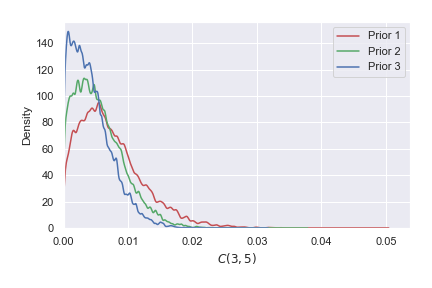}
        \includegraphics[width=0.105\textwidth]{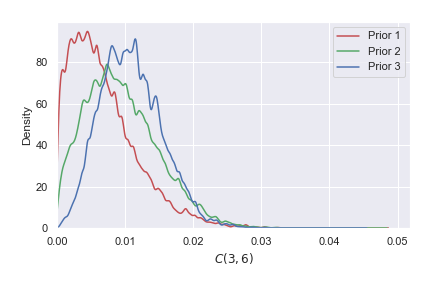}
        \includegraphics[width=0.105\textwidth]{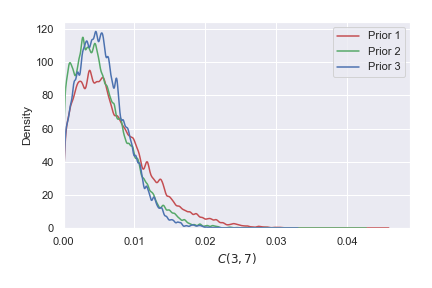}
        \includegraphics[width=0.105\textwidth]{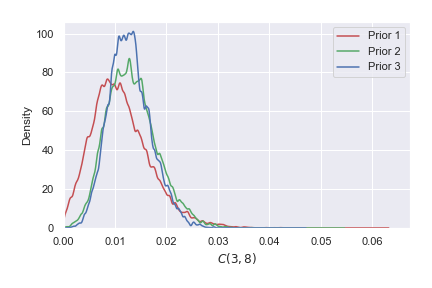}
        \includegraphics[width=0.105\textwidth]{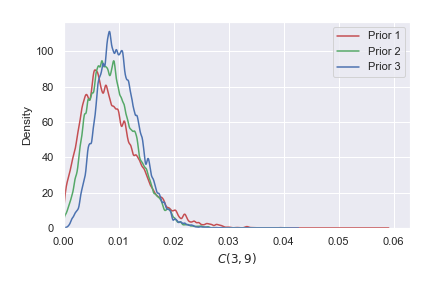}
    \end{tabular}
    \begin{tabular}{c c c c c c c c c}
        \includegraphics[width=0.105\textwidth]{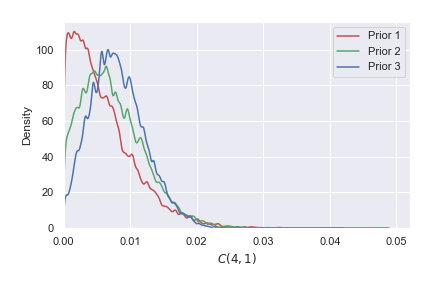}
        \includegraphics[width=0.105\textwidth]{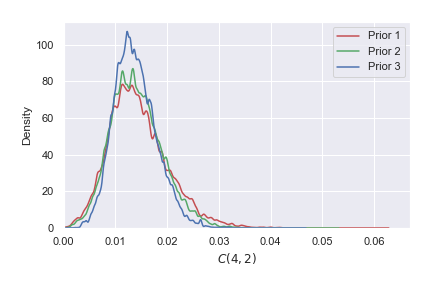}
        \includegraphics[width=0.105\textwidth]{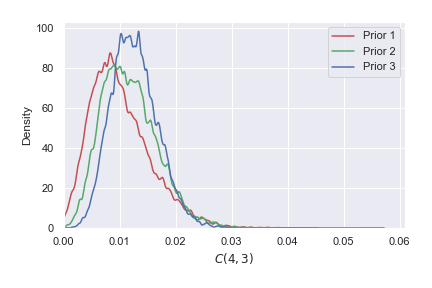}
        \includegraphics[width=0.105\textwidth]{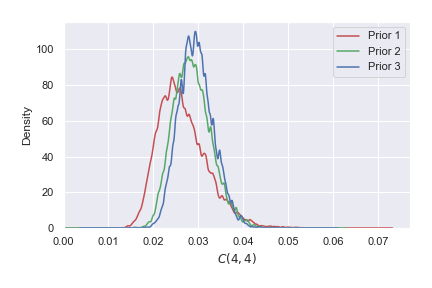}
        \includegraphics[width=0.105\textwidth]{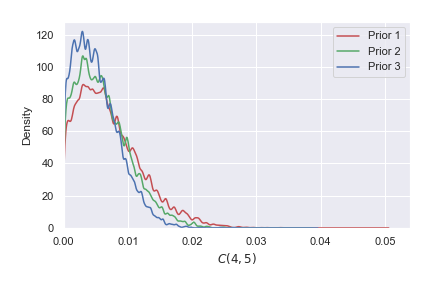}
        \includegraphics[width=0.105\textwidth]{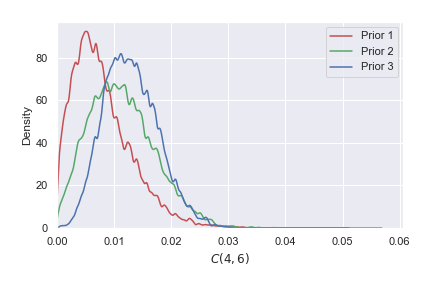}
        \includegraphics[width=0.105\textwidth]{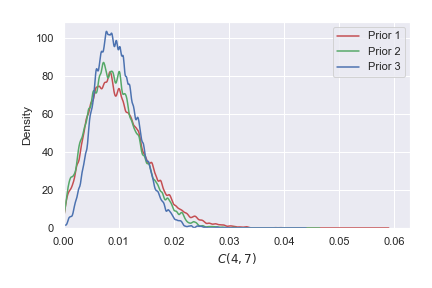}
        \includegraphics[width=0.105\textwidth]{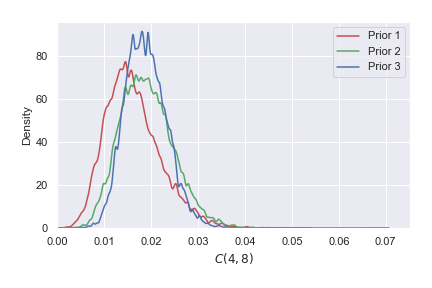}
        \includegraphics[width=0.105\textwidth]{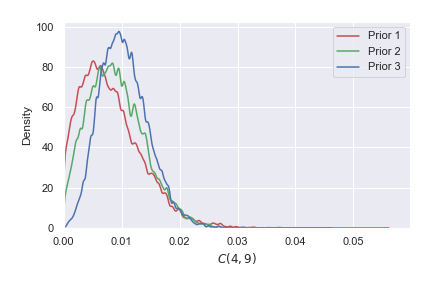}
    \end{tabular}
    \begin{tabular}{c c c c c c c c c}
        \includegraphics[width=0.105\textwidth]{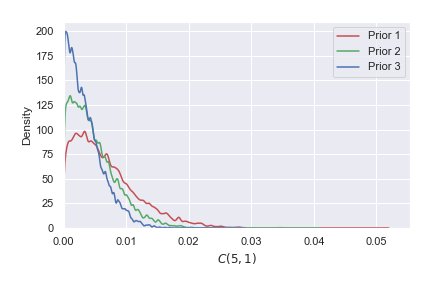}
        \includegraphics[width=0.105\textwidth]{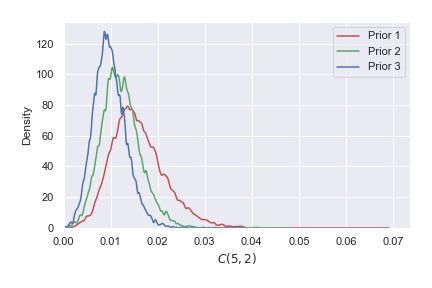}
        \includegraphics[width=0.105\textwidth]{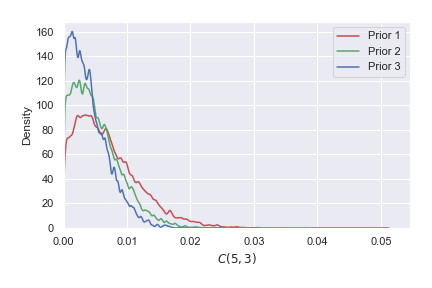}
        \includegraphics[width=0.105\textwidth]{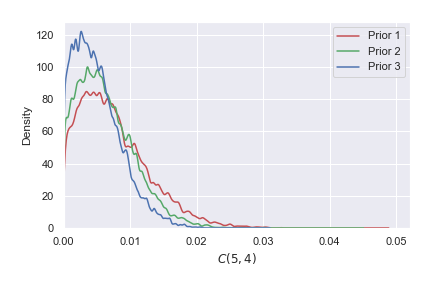}
        \includegraphics[width=0.105\textwidth]{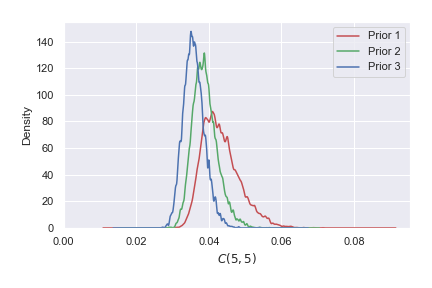}
        \includegraphics[width=0.105\textwidth]{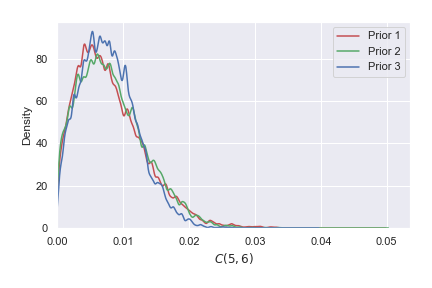}
        \includegraphics[width=0.105\textwidth]{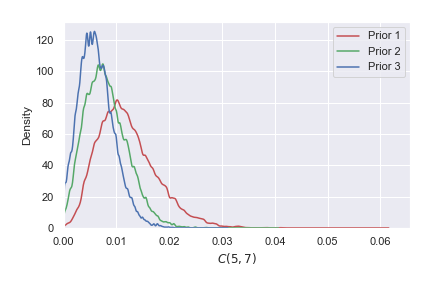}
        \includegraphics[width=0.105\textwidth]{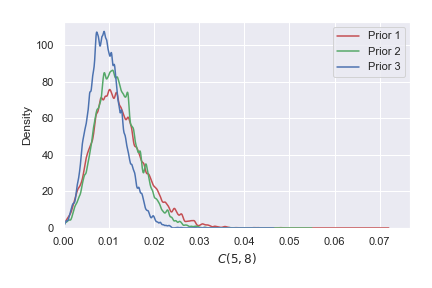}
        \includegraphics[width=0.105\textwidth]{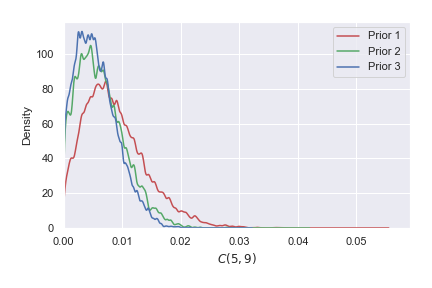}
    \end{tabular}
    \begin{tabular}{c c c c c c c c c}
        \includegraphics[width=0.105\textwidth]{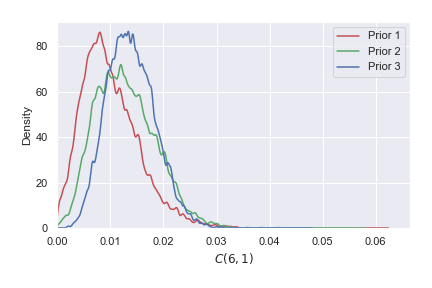}
        \includegraphics[width=0.105\textwidth]{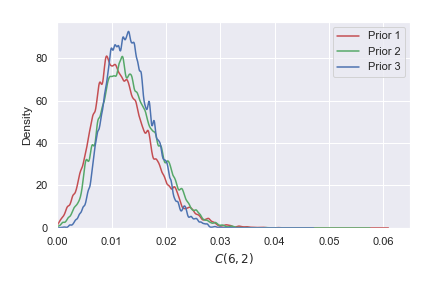}
        \includegraphics[width=0.105\textwidth]{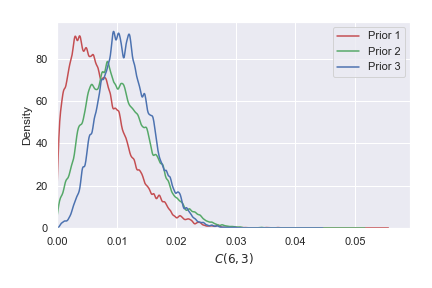}
        \includegraphics[width=0.105\textwidth]{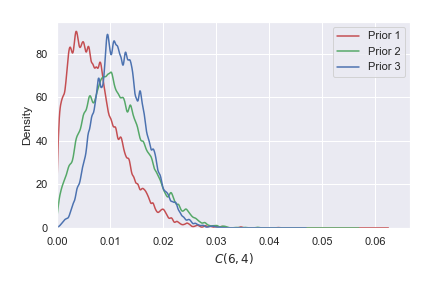}
        \includegraphics[width=0.105\textwidth]{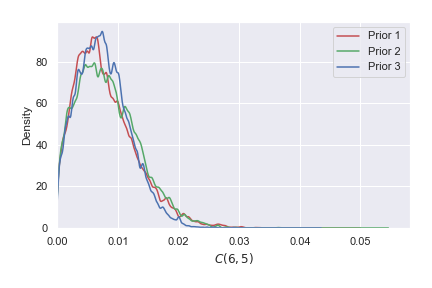}
        \includegraphics[width=0.105\textwidth]{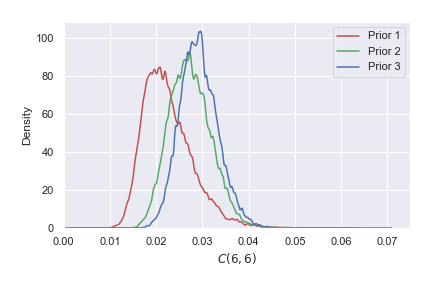}
        \includegraphics[width=0.105\textwidth]{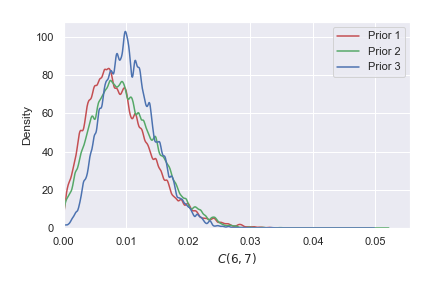}
        \includegraphics[width=0.105\textwidth]{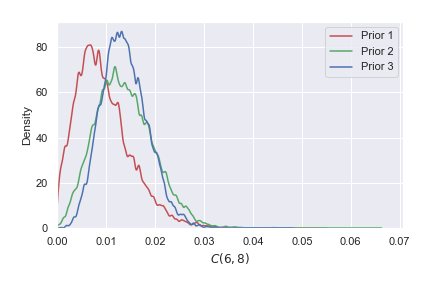}
        \includegraphics[width=0.105\textwidth]{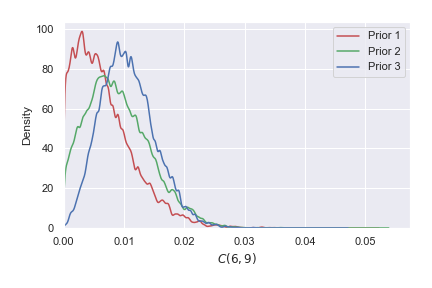}
    \end{tabular}
    \begin{tabular}{c c c c c c c c c}
        \includegraphics[width=0.105\textwidth]{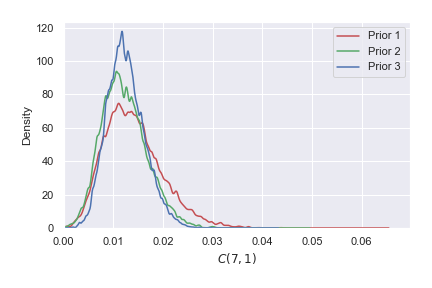}
        \includegraphics[width=0.105\textwidth]{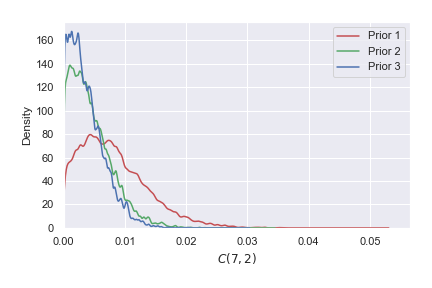}
        \includegraphics[width=0.105\textwidth]{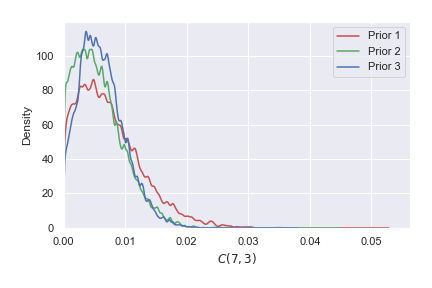}
        \includegraphics[width=0.105\textwidth]{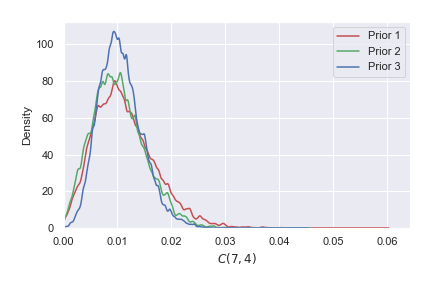}
        \includegraphics[width=0.105\textwidth]{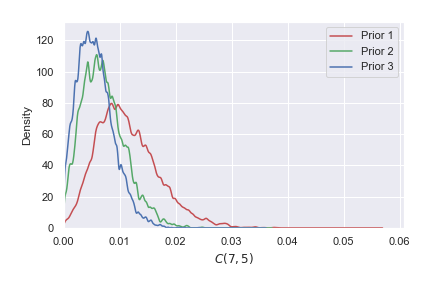}
        \includegraphics[width=0.105\textwidth]{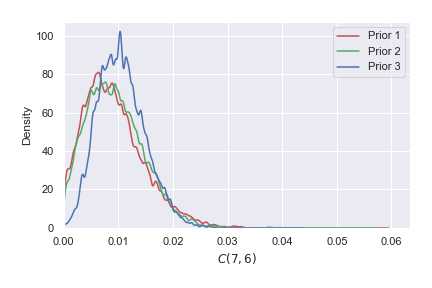}
        \includegraphics[width=0.105\textwidth]{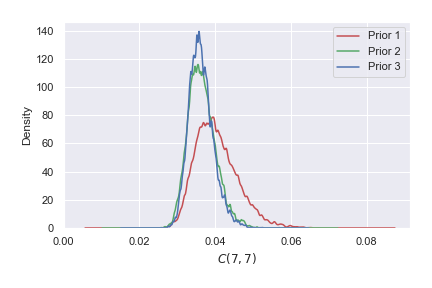}
        \includegraphics[width=0.105\textwidth]{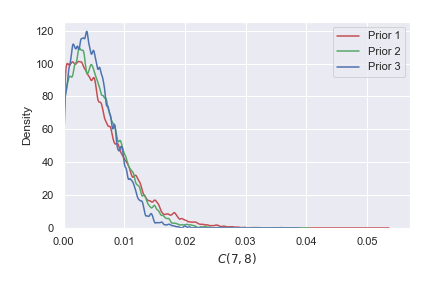}
        \includegraphics[width=0.105\textwidth]{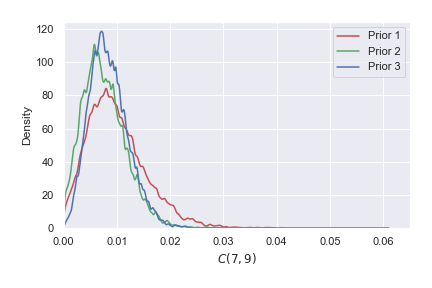}
    \end{tabular}
    \begin{tabular}{c c c c c c c c c}
        \includegraphics[width=0.105\textwidth]{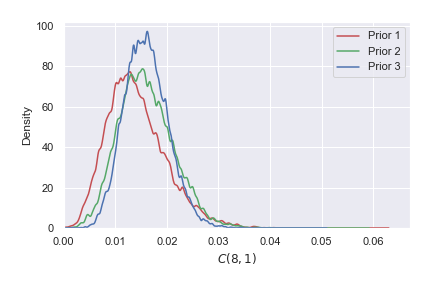}
        \includegraphics[width=0.105\textwidth]{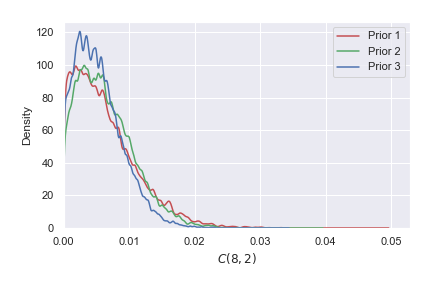}
        \includegraphics[width=0.105\textwidth]{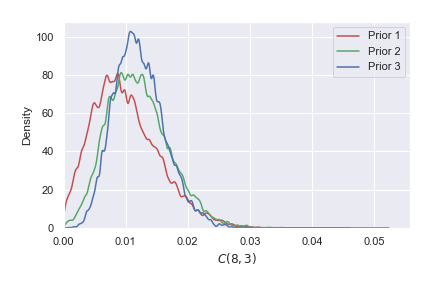}
        \includegraphics[width=0.105\textwidth]{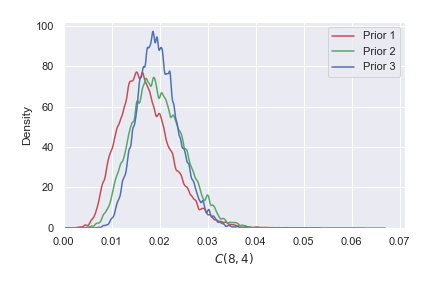}
        \includegraphics[width=0.105\textwidth]{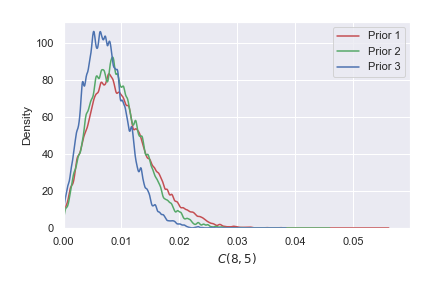}
        \includegraphics[width=0.105\textwidth]{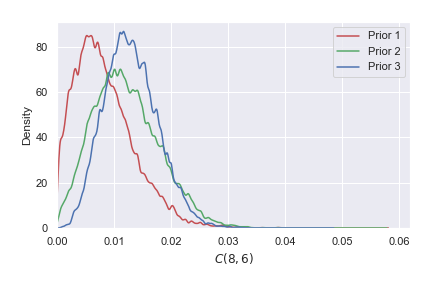}
        \includegraphics[width=0.105\textwidth]{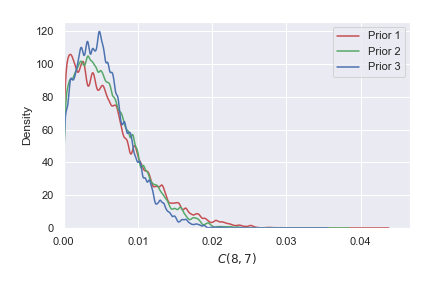}
        \includegraphics[width=0.105\textwidth]{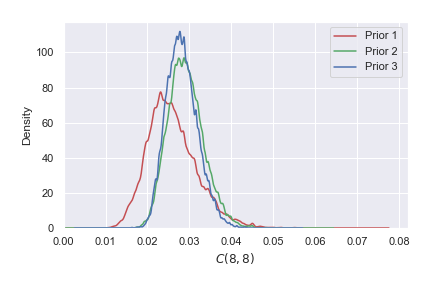}
        \includegraphics[width=0.105\textwidth]{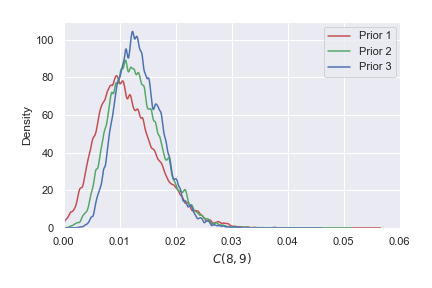}
    \end{tabular}
    \begin{tabular}{c c c c c c c c c}
        \includegraphics[width=0.105\textwidth]{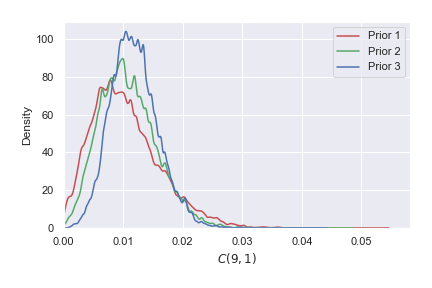}
        \includegraphics[width=0.105\textwidth]{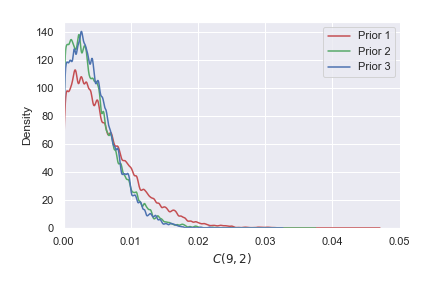}
        \includegraphics[width=0.105\textwidth]{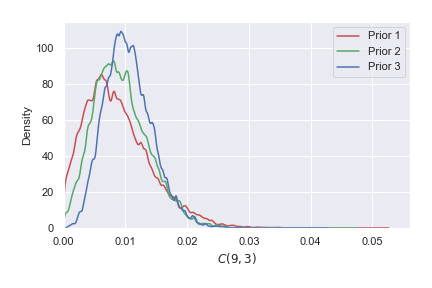}
        \includegraphics[width=0.105\textwidth]{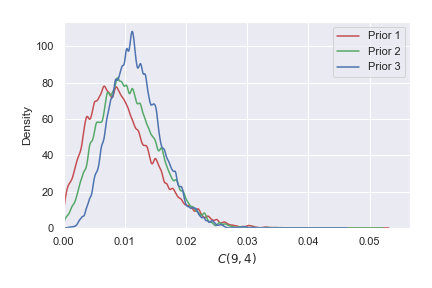}
        \includegraphics[width=0.105\textwidth]{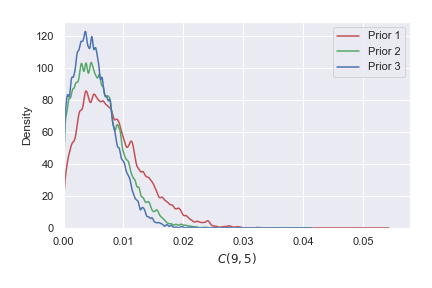}
        \includegraphics[width=0.105\textwidth]{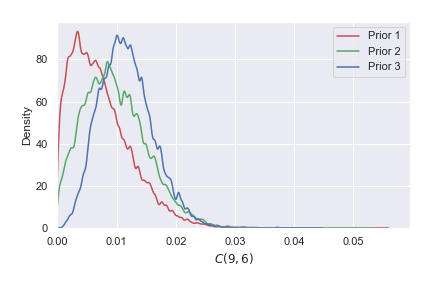}
        \includegraphics[width=0.105\textwidth]{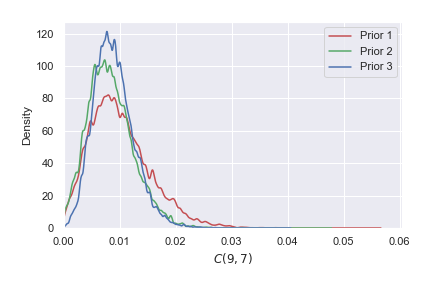}
        \includegraphics[width=0.105\textwidth]{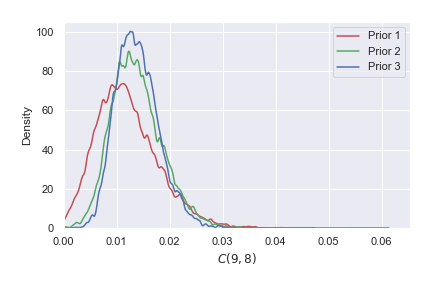}
        \includegraphics[width=0.105\textwidth]{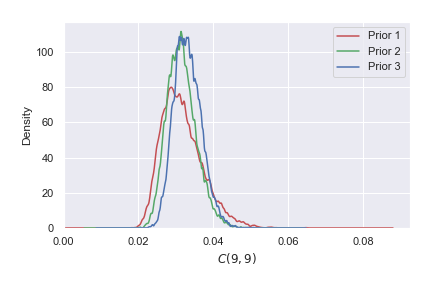}
    \end{tabular}
    \caption{The KDE of the inferred cost functions from the EU migration flow coupling. All $9\times 9 $ matrix elements are shown in order. The red curves are for prior (1), green for (2), and blue for (3).
    }
    \label{fig:EU_migration_cost}
\end{figure}

\begin{figure}[h!]
    \centering
    
    \begin{tabular}{c c c}
        \includegraphics[width=0.3\textwidth]{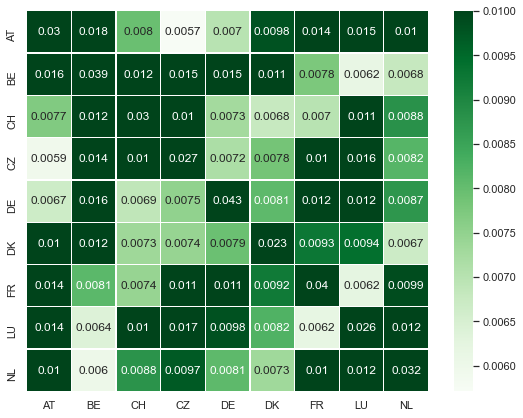}
        \includegraphics[width=0.3\textwidth]{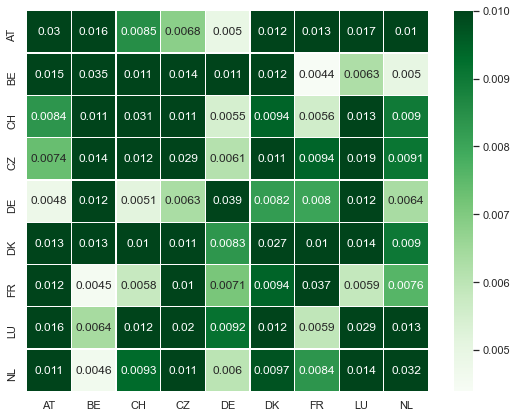}
        \includegraphics[width=0.3\textwidth]{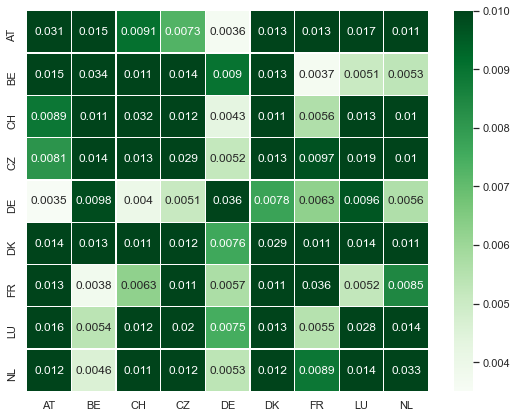}

    \end{tabular}
    
    \caption{The heatmaps of the mean of the inferred cost functions using \textbf{(left)} prior 1,  \textbf{(middle)} prior 2, and \textbf{(right)} prior 3.
    }
    \label{fig:heatmap_mean_cost}
\end{figure}

We utilize MetroMC method to infer the cost functions. In this section we assume that the sum of cost function is a constant. We use $\sum_{i,j}C_{i,j}=\sum_{i,j}\lambda C_{i,j}'$=320 and put Dirichlet priors on $C'$ so that there exists cost functions following the Dirichlet priors can be inferred from the ground truth plan. For each MCMC simulation using the three different Dirichlet priors we burn-in 10,000 steps and then take 10,000 samples with lags of 1,000 between. The standard deviations of the independent Gaussian random numbers (Centered at 0.) in each MCMC iteration are 0.1, 0.1, and 0.07 for prior (1), (2), and (3), respectively. The acceptance rate for the MCMC simulations are 0.58-0.68. The autocorrelation function and the running average of each simulation are shown in Fig.~\ref{fig:MC_details_EU_migration}. 1,000 steps are enough to de-correlate samples and 10,000 samples are able to reach convergence. The kernel density estimations (KDE) of each matrix element of the inferred cost functions are shown in Fig.~\ref{fig:EU_migration_cost}. The mean of each element of the cost for the three priors are shown in Fig.~\ref{fig:heatmap_mean_cost}.

For the noisy simulations on the EU migration flow we first generate a noise ($\epsilon$) which is 4\% of the value of $T^g_{(i,j)}$ and add it to $T^g_{(i,j)}$. Then, we sample ten Gaussian random numbers centered at 0 having a standard deviation equal to 4\% of $T^g_{(i,j)}$. The generated numbers are shown in table \ref{table:Gaussian_noises}. The MetroMC method is used for cost function inferences. In each Monte Carlo step we sample random variables $x \in \mathcal{N}(0,0.1^2)$ for the diagonal matrices $D^r$ and $D^c$. For each MCMC simulation we burn-in 10,000 steps and take 10,000 samples with a lag of 1,000 steps in between samples. The acceptance rates are 0.5-0.7. Once we infer the cost functions we compute the mean of them, $C^M$. Afterwards, forward OT simulation on $C^M$ using the observed marginals is performed to obtain the predicted coupling, $T^p$. The results are shown in table~\ref{table:missing_value}.

We follow the same procedure and use the same parameters for the MCMC simulations of predicting the missing values, only instead of ten random Gaussian samples we take 100 samples uniformly in range of [100, 25000) to fill the missing value.

\begin{table}[h!]
\caption{Sampled Gaussian noises. }
\label{table:Gaussian_noises}
\vskip 0.1in
\begin{center}
\begin{small}
\begin{sc}
\begin{tabular}{c | c | c | cccccccccc}
\toprule
(i,j) & $T^g_{(i,j)}$ & $\epsilon$ & \multicolumn{10}{c}{Noises} \\
\midrule
(3,6) & 452 & 3 & 7 & -2 & 5 & 22 & -22 & -6 & 3 & 6 & -24 & 5 \\
(4,5) & 6635 & -131 & -404 & 763 & -118 & -315 & 7 & 422 & -16 & 66 &-145 & 79 \\
(5,7) & 16560 & -243 & 785 & 205 & 269 & 525 & 678 & -164 & 217 & -180 & -379 & -459 \\
\bottomrule
\end{tabular}
\end{sc}
\end{small}
\end{center}
\vskip -0.3in
\end{table}

% You can have as much text here as you want. The main body must be at most $8$ pages long.
% For the final version, one more page can be added.
% If you want, you can use an appendix like this one, even using the one-column format.
%%%%%%%%%%%%%%%%%%%%%%%%%%%%%%%%%%%%%%%%%%%%%%%%%%%%%%%%%%%%%%%%%%%%%%%%%%%%%%%
%%%%%%%%%%%%%%%%%%%%%%%%%%%%%%%%%%%%%%%%%%%%%%%%%%%%%%%%%%%%%%%%%%%%%%%%%%%%%%%

\end{document}

\maketitle

\section{Proofs and Definitions for Section~\ref{sec:piot}}

\begin{repprop}{prop:DAD}
Let $T^{\lambda}$ be an observed optimal coupling, $\widehat{C} \in \Phi^{-1} (T^{\lambda})$ if and only if for every $\epsilon > 0$, 
there exist two positive diagonal matrices $D^r = \diag\{d^r_1, \dots, d^r_m\}$ and $D^c =\diag \{d^c_1,\dots, d^c_n\}$ such that: 
$\displaystyle |D^r \widehat{K}^{\lambda} D^c - T^{\lambda}| < \epsilon$, where $\widehat{K}^{\lambda} = e^{-\lambda \widehat{C}}$ and $|\cdot|$ is the $L^1$ norm. 
In particular, assume $T^{\lambda}$ is a positive matrix. Then $\widehat{C} \in \Phi^{-1} (T^{\lambda})$ 
if and only if there exists positive diagonal matrices $D^r,D^c$ such that $D^r \widehat{K}^{\lambda} D^c = T^{\lambda}$.
\end{repprop}

\begin{proof}
Let the row and column of marginals of $T^{\lambda}$ be $\mu$ and $\nu$.
Then $\widehat{C} \in \Phi^{-1} (T^{\lambda})$ if and only if $(\mu, \nu)$-Sinkhorn scaling of $\widehat{K}^{\lambda}$ converges to $T^{\lambda}$ in $L^1$ norm 
(for finite matrices, convergence in all $L^k$ norms are equivalent).
Notice that a row (column) normalization step in the Sinkhorn scaling is equivalent to a left(right) matrix multiplication of a positive diagonal matrix. 
Indeed, let $K_0 = \widehat{K}^{\lambda}$, $\mathbf{r}_0 = K_0 \mathbf{I}_n$ (vector for row sums of $K_0$), $D_0^{r} = \diag(\mu /\mathbf{r}_0)$.
Here $\mu /\mathbf{r}_0$ represents element-wise division. 
Let $K'_0$ be the matrix obtained by row normalization of $K_0$ with respect to $\mu$. Then $K'_0 = D_0^{r} K_0$.
Similarly, let $K_1$ be the matrix obtained by column normalization of $K'_0$ with respect to $\nu$.
Then $K_1 = K'_0 D_0^c$, where $D_0^c = \diag (\nu/ \mathbf{I}_m^T K'_0)$.
Iteratively, we have $K_s = \Pi_{i=1}^{s} D_i^r \cdot K_0 \cdot \Pi_{i=1}^{s} D_i^c$.

The \textbf{only if direction} [$\widehat{C} \in \Phi^{-1} (T^{\lambda}) \Longrightarrow$ existence of $D^r$ and $D^c$ for any $\epsilon$]: 
$\widehat{C} \in \Phi^{-1} (T^{\lambda}) \Longrightarrow $ $(\mu, \nu)$-Sinkhorn scaling of $\widehat{K}^{\lambda}$ converges to $T^{\lambda}$
$\Longrightarrow $, for any $\epsilon >0$, there exists $N>0$ such that for any $s>N$, $|K_s - T^{\lambda}| < \epsilon$, 
where $K_s = \Pi_{i=1}^{s} D_i^r \cdot \widehat{K}^{\lambda} \cdot \Pi_{i=1}^{s} D_i^c$.
Let $D^r = \Pi_{i=1}^{s} D_i^r$, $D^c = \Pi_{i=1}^{s} D_i^c$, the only if direction is complete.

The \textbf{if direction} [existence of $D^r$ and $D^c$ for any $\epsilon$ $\Longrightarrow$ $\widehat{C} \in \Phi^{-1} (T^{\lambda})$]: 
Let the limit of $(\mu, \nu)$-Sinkhorn scaling on $\widehat{K}^{\lambda}$ be $K^{*}$.
Hence $K^{*}$ and $T^{\lambda}$ have the same marginals and pattern. 
According to Lemma~A.3 of \cite{pei2019generalizing}, $K^{*}$ and $T^{\lambda}$ are diagonally equivalent. 
Further by Proposition~1 of \cite{pretzel1980convergence} $K^{*} = T^{\lambda}$.

In particular, when $T^{\lambda}$ is a positive matrix, $\widehat{K}^{\lambda}$ must also be a positive matrix.
Then $T^{\lambda}$ and $\widehat{K}^{\lambda}$ have the same pattern. 
$T^{\lambda} = \lim_{s\to \infty} \Pi_{i=1}^{s} D_i^r \cdot \widehat{K}^{\lambda} \cdot \Pi_{i=1}^{s} D_i^c$
implies that $\lim_{s\to \infty} \Pi_{i=1}^{s} D_i^r$ and $\lim_{s\to \infty} \Pi_{i=1}^{s} D_i^c$ exist. 
Hence the claim. \citep{rothblum1989scalings, idel2016review}

\end{proof}

\textit{Cross ratio equivalent} is defined in \citep{pei2019generalizing} as following:

\begin{definition}
Let $A = (a_{ij})_{n\times n}$ be a square matrix and $S_n$ be the set of all permutations of length~$n$.
For any $\sigma \in S_n$, the set of $n$-elements $\{a_{1,\sigma(1)}, a_{2,\sigma(2)}, \dots,a_{t,\sigma(n)}\}$
is called a \textbf{diagonal} of $A$.
If every $a_{k, \sigma(k)} > 0$, then the diagonal is said to be \textbf{positive}.
\end{definition}

%\pw{fix notation below}
\begin{definition}\label{def:cross ratios}
Let $A, B$ be two $n\times n$ matrices and $D^A_1=\{a_{1,\sigma(1)},\dots, a_{n, \sigma(n)}\}$ and $D^A_2=\{a_{1,\sigma'(1)},\dots, a_{n, \sigma'(n)}\}$ be two positive diagonals of $A$ determined by permutations $\sigma, \sigma'\in S_n$. 
Denote the products of elements on $D^A_1$ and $D^A_2$ by $d^A_1=\Pi_{i=1}^{n}a_{i, \sigma(i)}, d^A_2=\Pi_{i=1}^{n}a_{i, \sigma'(i)}$ respectively. Then $\CR(D^A_1, D^A_2)=d^A_1/d^A_2$ is called the \textbf{cross ratio} between $D^A_1$ and $D^A_2$ of $A$. Further, let the diagonals in $B$ determined by the same $\sigma$ and $\sigma'$ be 
$D^B_1=\{b_{1,\sigma(1)},\dots, b_{n, \sigma(n)}\}$ and $D^B_2=\{b_{1,\sigma'(1)},\dots, b_{n, \sigma'(n)}\}$. We say $A$ is \textbf{cross ratio equivalent} to $B$, denoted by $A\overset{cr}{\sim} B$, if $d_i^A\neq 0 \Longleftrightarrow d_i^B\neq 0$
and $\CR(D^A_1, D^A_2)=\CR(D^B_1, D^B_2)$ holds for 
any $D^A_1$ and $D^A_2$.
\end{definition}
\section{Additional details of experiments}

\subsection{Details of Monte Carlo method}

In Algorithm~\ref{alg:MCMC} we propose a general approach for the MCMC method. Here we describe the method for this work in details. For the initial step of the MCMC we assume the observed coupling $T$ came from some unknown cost and take $K^{(0)}=T$ as the starting point of the inverse OT. We assume a symmetric Dirichlet prior with concentration parameter $\alpha$, and samples are drawn from the posterior distribution. In the $i$-th MC step the acceptance probability is calculated by $r = \frac{P_{Dir}(K^{i+1}| \alpha)}{P_{Dir}(K^{i}| \alpha)}$, where $P_{Dir}(M| \alpha) = \prod_j{\text{Dir}(m_j| \alpha)}$, $m_j$ is the $j$-th column of $M$, and $\text{Dir}(m_j| \alpha)$ is the Dirichlet probability density function with the same concentration $\alpha$ for all the elements in $m_j$. Because a new state in the Markov chain is obtained by multiplying the current matrix by two diagonal matrices from the left and right and then normalizing the columns of the new matrix and the two diagonal matrices are generated by taking the exponential of zero-mean Gaussian distribution, the reversibility of the Markov chain is assured by the symmetry of the Gaussian. 

\subsection{Hyperparameters}

To enable standard checks of convergence and mixing, we employ multiple chains based on random restarts. We take samples from 50 Markov chains, and each chain is run for 11,000 total steps, with 1,000 burn-in, and samples drawn with a lag of 1,000. Thus, we draw 10 total samples from each chain and a total of 500 samples in one simulation. We consider $\sigma$ = 0.005 for the Gaussian variance to generate the two diagonal matrices in all simulations and $\alpha$ = 0.5, 1.0, and 2.0 for the Dirichlet prior in different runs. 

\section{Additional results}

Fig.~\ref{fig:KDE_transfer_supp} plots the same quantities as in Fig.~\ref{fig:KDE_transfer} from the main text. In addition to the results simulated using $\alpha=1.0$ (middle row of Fig.~\ref{fig:KDE_transfer_supp} and main text) we show the cases for $\alpha=0.5$ (top row) and $\alpha=2.0$ for comparison. One significant difference lies in the orange curves, where the KDEs of the $\alpha=0.5$ case has the largest mean and width among all three $\alpha$ cases, while the $\alpha=2.0$ has the smallest mean and width. The reason is that when using $\alpha=0.5$ for the Dirichlet prior the distribution of the values in the inferred cost have larger variance. When we shuffle the values in the matrix, the variance stays the same and leads to wider widths in the KDE of $D(\tilde{T}^{shfl},T)$ and $D(\text{log}(CR(\tilde{K}^{shfl})),\text{log}(CR(T)))$. The next difference is that the blue curves for both quantities with $\alpha=0.5$ has longer tail toward large $D$. This suggests that the distribution of the values in the observed coupling $T$ has smaller variance than the $\alpha=0.5$ case, so the difference between the log of the cross-ratio matrices of $K$ and $T$ has higher chance to obtain a large value. While in the $\alpha=1.0$ and $\alpha=2.0$ cases the matrix value variance is closer to $T$, therefore, the tail in large D is less obvious. 
%Nevertheless, the above differences do not affect our main conclusion where the non-shuffled results have much smaller mean value and shallower width, which suggesting that our predictions on the coupling and cross-ratio of the cost are accurate.
These results support the main conclusion that our predictions regarding the coupling and cross-ratio are remarkably accurate. 

\begin{figure}[h!]
    \centering
    \begin{tabular}{c c c c}
        \includegraphics[width=0.47\textwidth]{figs/KDE_transfer_prediction_alpha0p5.png}
        &
        \includegraphics[width=0.47\textwidth]{figs/KDE_transfer_CR_alpha0p5.png}
    \end{tabular}
    \begin{tabular}{c c c c}
        \includegraphics[width=0.47\textwidth]{figs/KDE_transfer_prediction.png}
        &
        \includegraphics[width=0.47\textwidth]{figs/KDE_transfer_CR.png}
    \end{tabular}
    \begin{tabular}{c c c c}
        \includegraphics[width=0.47\textwidth]{figs/KDE_transfer_prediction_alpha2.png}
        &
        \includegraphics[width=0.47\textwidth]{figs/KDE_transfer_CR_alpha2.png}
    \end{tabular}
    \caption{Predicting player performance based on the previous season's inferred costs using different Dirichlet parameter $\alpha$. \textbf{(top)} $\alpha=0.5$. \textbf{(middle)} $\alpha=1.0$ \textbf{(bottom)} $\alpha=2.0$. The $\alpha=1.0$ plots are the same figures as in the main text. \textbf{(left)} The blue curve plots the KDE of the $L_2$ distance between the predicted couplings for the assembled team, $\tilde{T}$ and the observed coupling, $T$. The orange curve plots the KDE of the $L_2$ distance but we shuffle the elements in $\tilde{K}$ and then compute the EOT plan, $\tilde{T}^{shfl}$. \textbf{(right)} The blue curve represents the KDE of the $L_2$ distance between the log of the cross-ratio matrices of the negative log of the cost for the assembled team and the log of the cross-ratio matrix of the observed coupling. The orange curve plots the same KDE of the $L_2$ distance but the elements in $\tilde{K}$ are shuffled. The bandwidths of the KDE are 0.2.  }
    \label{fig:KDE_transfer_supp}
\end{figure}

\bibliographystyle{plainnat}
\bibliography{AISTATS-2022/references}